\documentclass{article} %
\usepackage{iclr2027_conference,times}

\usepackage{amsmath,amsfonts,bm}

\def\1{\bm{1}}

\def\vr{{\bm{r}}}

\def\vv{{\bm{v}}}

\def\vx{{\bm{x}}}

\DeclareMathAlphabet{\mathsfit}{\encodingdefault}{\sfdefault}{m}{sl}
\SetMathAlphabet{\mathsfit}{bold}{\encodingdefault}{\sfdefault}{bx}{n}

\newcommand{\E}{\mathbb{E}}

\DeclareMathOperator*{\argmax}{arg\,max}

\usepackage{amsmath,amssymb,amsthm,mathtools}
\usepackage{booktabs}
\usepackage{graphicx}
\usepackage{subcaption}
\usepackage[font=small]{caption}
\usepackage[colorlinks=true,linkcolor=blue,citecolor=blue]{hyperref}
\usepackage{cleveref}
\usepackage{algorithm}
\usepackage{wrapfig}
\usepackage[noend]{algpseudocode}
\usepackage{multirow}
\newtheorem{theorem}{Theorem}

\newtheorem{proposition}{Proposition}

\newtheorem{remark}{Remark}

\newcommand{\methodfull}{DUal-readout of diffusion models for OOD detection}
\newcommand{\method}{DUO}

\title{DUO: Dual Readouts of Diffusion Models for OOD Detection}

\author{Alireza Javid\textsuperscript{1}, Shirin Shoushtari\textsuperscript{2}, Nuria Gonz{\'a}lez-Prelcic\textsuperscript{1}\\
\textsuperscript{1}University of California, San Diego\\
\textsuperscript{2}Washington University in St.~Louis\\
\texttt{sajavid@ucsd.edu, s.shirin@wustl.edu, ngprelcic@ucsd.edu}}

\iclrfinalcopy %

\begin{document}

\maketitle
\makeatletter
\lhead{}\renewcommand{\headrulewidth}{0pt}
\newcommand{\ftype@noticebox}{8}
\enlargethispage{2\baselineskip}
\@float{noticebox}[b]\footnotesize Preprint.\end@float
\makeatother

\begin{abstract}

Diffusion models are powerful generative methods for learning and sampling from complex data distributions. Safely deploying these models requires detecting inputs that fall outside their training distribution. Because diffusion models recover clean samples from noisy inputs, reconstruction error is often used for out-of-distribution (OOD) detection. This approach assumes that in-distribution (ID) inputs are reconstructed more accurately than OOD inputs. However, OOD inputs may be reconstructed equally well or even better, making this score unreliable. We therefore propose the leading \emph{sensitivity spectrum} of intermediate denoising features as a complementary OOD signal and demonstrate that it remains informative when reconstruction error is misleading. We introduce \methodfull\ (\method), a fully unsupervised, training-free OOD detector that combines reconstruction error and feature sensitivity. \method\ improves average AUROC by approximately 6\% points over the strongest baselines while substantially reducing inference cost, and achieves SOTA on near-OOD and high-resolution benchmarks. 

\end{abstract}

\section{Introduction}

Reliable deployment of machine-learning systems requires detecting when test inputs fall outside the training distribution. Such OOD inputs can compromise predictions in safety-critical applications, including medical imaging and autonomous driving~\citep{hendrycks2016baseline,yang2024generalized}. We consider a fully unsupervised setting, where only unlabeled ID data are available. Diffusion models are well-suited to this setting because they learn the ID data through denoising, and their response to test inputs can provide useful OOD signals~\citep{graham2023denoising}.

The most direct way to use generative models for OOD detection is through likelihood. However, deep generative models have been shown to assign higher likelihood to OOD inputs than to ID data~\citep{nalisnick2019deep,choi2019waic,ren2019likelihood}. For diffusion models, denoising or reconstruction error is a natural alternative~\citep{graham2023denoising,liu2023unsupervised,yang2024diffusion}, based on the assumption that ID inputs are reconstructed more accurately than OOD inputs. Yet reconstruction quality reflects how easily an input can be denoised, not necessarily whether it belongs to the training distribution. Consequently, an ID-trained diffusion model may reconstruct OOD inputs as well as ID inputs, since the learned denoiser adapts to the geometry of the input~\citep{kadkhodaie2024generalization,kamb2024creativity}, producing overlapping or even inverted reconstruction scores.

Recent diffusion-based OOD detection methods have explored signals beyond reconstruction, including diffusion-path geometry~\citep{heng2024diffpath,abdi2025zero}, equivariance consistency~\citep{rouzoumka2026gepc}, score-curvature ratios~\citep{barkley2025scoped}, and posterior-covariance spectra~\citep{shoushtari2026eigenscore}. These methods demonstrate that diffusion models contain useful OOD information beyond reconstruction error. However, they primarily characterize the denoiser through its output or diffusion trajectory, leaving the local response of intermediate denoising features largely unexplored. These internal responses may preserve distributional information that is lost in the final reconstruction. This leaves open whether intermediate-feature sensitivity can provide an informative OOD signal and how it can be identified and calibrated using only ID data.

\begin{figure}[t]
    \centering
     \includegraphics[width=.9\linewidth]{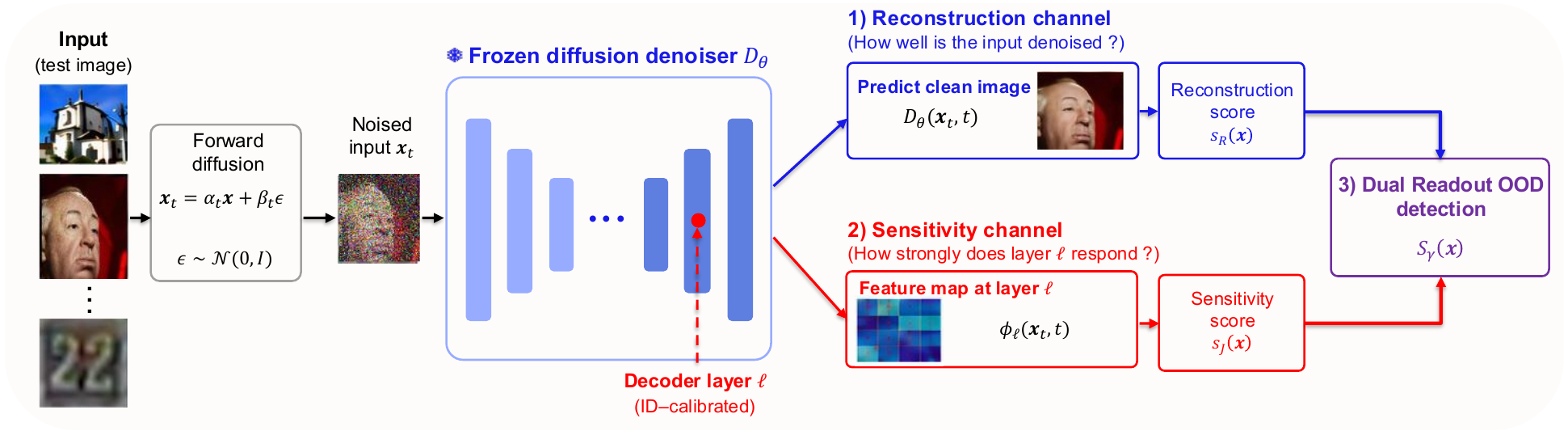}
     \vspace{-3mm}
    \caption{
    \textbf{Overview of \method.}
    A test image $\vx$ is noised and passed through a diffusion denoiser $D_\theta$. The reconstruction channel measures the denoising residual, giving $s_R(\vx)$; the sensitivity channel measures the local response of an intermediate feature map $\phi_\ell(\vx_t,t)$ to input perturbations, giving $s_J(\vx)$. The two scores are combined into the \method\ score $S_\gamma(\vx)$, with higher values indicating OOD (Section~\ref{sec:method}).
    }
    \label{fig:method-overview}
\end{figure}

Our key idea is that reconstruction and intermediate-feature sensitivity capture complementary information from the denoiser. Reconstruction measures how well the model recovers a clean sample from a noisy input, whereas sensitivity measures how intermediate features respond to small input perturbations. We quantify this response using the leading spectrum of the feature Jacobian. By probing intermediate features, this signal may capture local responses that are weakened or lost in the final reconstruction. This distinction is supported by Tweedie’s and Miyasawa’s identities, which connect the denoised output and its Jacobian to the smoothed score and posterior covariance, respectively~\citep{miyasawa1961empirical,vincent2011connection,manor2024posterior}. We therefore introduce \textbf{\method}, a fully unsupervised, training-free OOD detector that combines these signals. Using only ID data, \method\ selects an informative feature layer and noise levels, calibrates both signals, and determines their fusion. The resulting one-sided score assigns higher values to stronger OOD evidence. Figure~\ref{fig:method-overview} summarizes the complete pipeline.

\method\ improves average AUROC on the main benchmark by approximately 6\% points over the strongest baseline. It also achieves state-of-the-art performance on near-OOD and high-resolution benchmarks, with particularly large gains in the high-resolution setting. Our theoretical and empirical results show that intermediate-feature sensitivity remains informative when reconstruction is unreliable, providing complementary evidence for separating ID from OOD inputs. DUO achieves these improvements while running more than $30\times$ faster than the strongest baseline~\citep{shoushtari2026eigenscore}.
Our main contributions are:

\begin{itemize}

\item We propose \method, a fully unsupervised, training-free framework that uses the leading sensitivity spectrum of intermediate denoising features as a complementary signal to reconstruction for OOD detection.

\item We provide theoretical insight into why reconstruction scores can be unreliable, including a regime in which their ID--OOD ordering inverts. 

\item We show that intermediate-feature sensitivity can retain OOD-relevant information weakened or lost at the denoiser output, and characterize when an ID-calibrated combination of reconstruction and sensitivity improves ID--OOD separation.

\item We demonstrate SOTA average AUROC on the main 12-pair, near-OOD, and high-resolution benchmarks, while achieving over $30\times$ faster inference than EigenScore, the strongest baseline on the main benchmark.

\end{itemize}

\section{Background and problem setup}

{\bf Diffusion models.} Let $X_1 \in \mathbb{R}^d$ denote a data sample drawn from the ID data distribution $p$, and let $X_0 \sim \mathcal{N}(0,I_d)$ be an independent Gaussian noise sample. For $t \in [0,1]$, we define the forward noising process as \begin{equation} X_t = \alpha_t X_1 + \beta_t X_0, \qquad X_0 \perp X_1, \label{eq:vp-forward} \end{equation} where $\alpha_t,\beta_t \geq 0$ are noise-schedule coefficients satisfying $\alpha_0=0$, $\alpha_1=1$, $\beta_0=1$, and $\beta_1=0$. We assume that $\alpha_t$ is nondecreasing and $\beta_t$ is nonincreasing, so that $X_t$ interpolates from pure Gaussian noise at $t=0$ to clean data at $t=1$. The reverse process aims to recover the clean data sample, or equivalently, the injected noise, from the noisy observation $X_t$. This is typically done by training a denoising network using a mean-squared error objective. Depending on the chosen parameterization, the model may be trained to predict either the clean sample $X_1$ or the noise $X_0$: 
\begin{equation} \mathcal{L}_{\mathrm{data}}(\theta) = \mathbb{E}_{X_1,X_0,t} \left[ \left\| D_\theta(X_t,t) - X_1 \right\|_2^2 \right], \quad \mathcal{L}_{\mathrm{noise}}(\theta) = \mathbb{E}_{X_1,X_0,t} \left[ \left\| \epsilon_\theta(X_t,t) - X_0 \right\|_2^2 \right]. \label{eq:noise-pred-loss} \end{equation} 
These objectives encourage the denoiser to learn the conditional structure of the data distribution across noise levels, which enables sampling by progressively transforming Gaussian noise into data-like samples. Throughout, bold lowercase $\vx,\epsilon,\vx_t$ denote realizations of the random variables $X_1,X_0,X_t$.

\begin{figure}[t]  
    \centering

    \begin{subfigure}[t]{.75\linewidth}
        \centering
        \includegraphics[width=\linewidth]{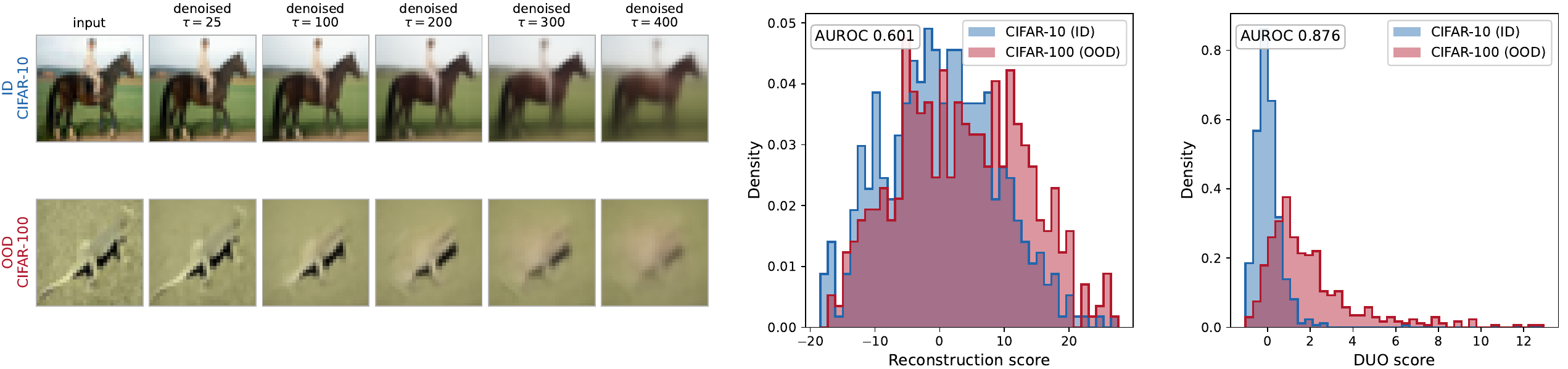}
        \caption{CIFAR-10 as ID and CIFAR-100 as OOD.}
        \label{fig:mse_failure_cifar}
    \end{subfigure}

    \begin{subfigure}[t]{.75\linewidth}
        \centering
        \includegraphics[width=\linewidth]{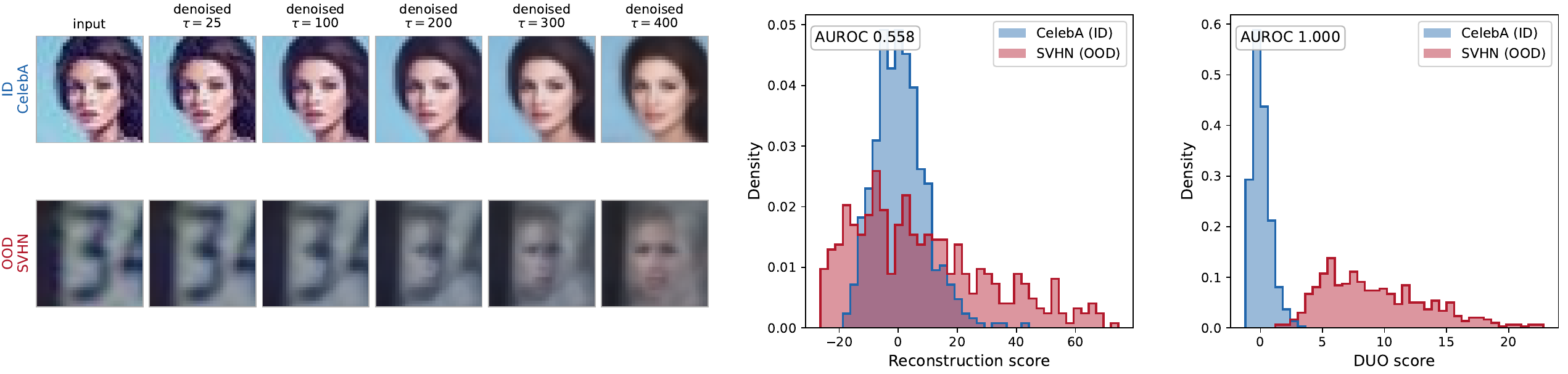}
        \caption{CelebA as ID and SVHN as OOD.}
        \label{fig:mse_failure_celeba}
    \end{subfigure}
\vspace{-2.5mm}
    \caption{
    \textbf{Reconstruction-only scoring fails where the dual readout succeeds.}
    For each ID--OOD pair we show, from left to right, representative inputs and their denoised estimates at several noise levels using an ID diffusion model, the distribution of the ID-calibrated reconstruction score, and the distribution of fused \method\ score. Although trained only on ID data, the denoiser reconstructs OOD inputs plausibly, producing substantial overlap between ID and OOD reconstruction scores. By incorporating intermediate-feature sensitivity, \method\ produces better-separated score distributions and higher AUROC.
    }
    \label{fig:mse_failure}
\end{figure}

{\bf Reconstruction MSE as an OOD score.} The diffusion training objective in Eq.~\eqref{eq:noise-pred-loss} suggests a natural OOD score, the expected reconstruction error of a test input 
\begin{equation} s_{\mathrm{MSE}}(\vx) = \mathbb{E}_{t,\epsilon} \left[ \left\| D_\theta(\alpha_t \vx+\beta_t\epsilon,t)-\vx \right\|_2^2 \right], \qquad \epsilon\sim\mathcal{N}(0,I). 
\label{eq:mse-ood-score} \end{equation} 
Since the model is trained only on ID data, one might expect lower reconstruction error for ID inputs than for OOD inputs. However, this assumption does not always hold. As shown in the left and middle blocks of Figure~\ref{fig:mse_failure}, an ID denoiser can produce plausible reconstructions for OOD inputs across multiple noise levels, particularly when the two distributions share low-level statistics, texture, or semantic structure. Consequently, their reconstruction scores can overlap substantially, limiting reliable threshold-based detection. More fundamentally, reconstruction error summarizes the denoising response only through residual magnitude, discarding information about how the model responds locally to the input.

\section{\method: Dual Readouts of Diffusion Models}
\label{sec:method}
In this section, we first explain why reconstruction MSE can be unreliable, then define an intermediate-feature sensitivity score and its calibrated combination with reconstruction, and finally describe how \method\ selects the feature layer and noise levels using only ID data.

\subsection{Why MSE is insufficient}
\label{subsec:mse-insufficient}

For ID data sampled from distribution $p$, the MSE-optimal denoiser at noise level $t$ is the posterior mean
\[
    D_p(\vx_t,t)
    =
    \mathbb{E}_p[X_1\mid X_t=\vx_t].
\]
Low reconstruction error therefore indicates that an input is easy to denoise under the learned posterior mean, not necessarily that it belongs to the ID. Under the forward process in
Eq.~\eqref{eq:vp-forward}, Tweedie's formula gives, for $\alpha_t>0$ 
\begin{equation}
    D_p(\vx_t,t)
    =
    \frac{1}{\alpha_t}
    \left(
        \vx_t+\beta_t^2\nabla\log p_t(\vx_t)
    \right),
    \label{eq:tweedie}
\end{equation}
where $p_t$ denotes the density of $X_t$ when $X_1\sim p$, and $q_t$ the
density when $X_1\sim q$, the OOD data. 
To characterize when reconstruction reverses the desired ID--OOD ordering, we consider the fixed-noise score at level $t$ corresponding to Eq.~\eqref{eq:mse-ood-score}
\begin{equation}
    s_{\mathrm{MSE}}(\vx;t)
    =
    \mathbb{E}_{\epsilon}
    \left[
        \left\|
        \vx-D_p(\alpha_t\vx+\beta_t\epsilon,t)
        \right\|_2^2
    \right],
    \qquad
    \epsilon\sim\mathcal{N}(0,I_d).
    \label{eq:mse-fixed-t}
\end{equation}

The following proposition analyzes this Bayes-optimal score, separating an intrinsic limitation of reconstruction from approximation error in the trained denoiser $D_\theta$. It decomposes the ID--OOD score gap into a relative denoising-difficulty term (denoisability gap), which can favor either distribution, and a nonnegative distribution-mismatch term. If OOD inputs are sufficiently easier to denoise than ID inputs, their denoising advantage can dominate the mismatch and give them lower reconstruction scores.

\begin{proposition}[MSE gap decomposition]
\label{prop:mse}
Let
$D_p=\mathbb{E}_p[X_1\mid X_t]$ and
$D_q=\mathbb{E}_q[X_1\mid X_t]$
be the Bayes denoisers under the ID law $p$ and an OOD law $q$. Let $s_{\mathrm{MSE}}(\cdot\,;t)$ be the reconstruction score induced by ID Bayes denoiser, as defined in Eq.~\eqref{eq:mse-fixed-t}, and define 
\[
    \Delta_{\mathrm{MSE}}(t)
    =
    \mathbb{E}_{X_1\sim q}
    \big[
        s_{\mathrm{MSE}}(X_1;t)
    \big]
    -
    \mathbb{E}_{X_1\sim p}
    \big[
        s_{\mathrm{MSE}}(X_1;t)
    \big].
\]
Thus, $\Delta_{\mathrm{MSE}}(t)<0$ indicates that the reconstruction score is inverted, assigning lower values to OOD inputs on average. Then
\begingroup
\footnotesize
\begin{align}
    \Delta_{\mathrm{MSE}}(t)
    =
    \underbrace{
    \mathbb{E}_{q_t}
    \left[
        \operatorname{tr}
        \operatorname{Cov}_q[X_1\mid X_t]
    \right]
    -
    \mathbb{E}_{p_t}
    \left[
        \operatorname{tr}
        \operatorname{Cov}_p[X_1\mid X_t]
    \right]
    }_{B(t)\ \text{denoisability gap}}
    +
    \underbrace{
    \frac{\beta_t^4}{\alpha_t^2}
    \mathbb{E}_{q_t}
    \left[
        \left\|
        \nabla\log p_t-\nabla\log q_t
        \right\|_2^2
    \right]
    }_{\text{score-mismatch energy}\ge 0}.
    \label{eq:mse-decomp}
\end{align}
\endgroup
\end{proposition}

Proposition~\ref{prop:mse} decomposes the reconstruction-score gap into a denoisability gap $B(t)$ and a nonnegative score-mismatch term. The mismatch term vanishes when the smoothed ID and OOD score fields agree on $q_t$, whereas $B(t)$ compares their average posterior uncertainty. When OOD inputs are easier to denoise, $B(t)$ can be negative and outweigh the mismatch term, giving OOD inputs lower reconstruction errors and reversing the desired ID--OOD ordering. Appendix~\ref{app:proof-mse} provides the proof and the precise inversion condition, while Appendix~\ref{app:prop2-validation} supports this mechanism on the two most strongly inverted benchmark pairs.

Because reconstruction MSE uses only the residual magnitude, it does not capture the local geometry of the denoising map. Miyasawa’s posterior-covariance identity provides access to this complementary information:
\begin{equation}
    \operatorname{Cov}_p[X_1\mid X_t=\vx_t]
    =
    \frac{\beta_t^2}{\alpha_t}
    \partial_{\vx_t}D_p(\vx_t,t)
    =
    \frac{\beta_t^2}{\alpha_t^2}
    \left(
        I+\beta_t^2\nabla^2\log p_t(\vx_t)
    \right).
    \label{eq:miyasawa}
\end{equation}
Equation~\eqref{eq:miyasawa} links the denoiser Jacobian to posterior covariance, so inputs with similar reconstructions can still exhibit different local sensitivities. \method\ captures this complementary information through the leading sensitivity spectrum of intermediate denoising features. At intermediate layers, informative response directions may remain visible before being attenuated or canceled by downstream mappings.

\subsection{Feature sensitivity spectrum} \label{subsec:sensitivity}

We measure feature sensitivity by perturbing an intermediate decoder representation of a noisy input. Let $\phi_\ell(\vx_t,t)\in\mathbb{R}^{m_\ell}$ be the activation at decoder tap $\ell$, evaluated at $\vx_t=\alpha_t\vx+\beta_t\epsilon$ with $\epsilon\sim\mathcal{N}(0,I_d)$. For $M$ isotropic unit directions $\vv_1,\ldots,\vv_M$ and probe scale $\delta>0$, define the central-difference responses
\[\vr_j=\frac{\phi_\ell(\vx_t+\delta\vv_j,t)-\phi_\ell(\vx_t-\delta\vv_j,t)}{2\delta},\qquad j=1,\ldots,M.\]
Let $R_\ell(\vx_t,t)\in\mathbb{R}^{M\times m_\ell}$ contain $\vr_j^\top$ as its rows, and define the empirical sensitivity covariance
\[C_\ell(\vx_t,t)=\frac{1}{M}R_\ell(\vx_t,t)^\top R_\ell(\vx_t,t).\]
Its leading eigenvalues capture the dominant directions in which the intermediate representation responds to input perturbations. We define the sensitivity score as the noise-averaged top-$K_{\mathrm{spec}}$ spectral mass:
\begin{equation} 
s_J(\vx) = \mathbb{E}_{t,\epsilon} \left[ \sum_{k=1}^{K_{\mathrm{spec}}} \lambda_k \big( C_\ell(\alpha_t\vx+\beta_t\epsilon,t) \big) \right], 
\label{eq:sJ} 
\end{equation}
where $\lambda_k(\cdot)$ is the $k$-th largest eigenvalue. 
Let $J_\ell=\partial\phi_\ell/\partial\vx_t$ denote the feature Jacobian. 
As $\delta\to0$ and the number of probes $M$ increases, $C_\ell$ estimates $J_\ell J_\ell^\top$ up to a probe normalization constant. Because $\operatorname{rank}(C_\ell)\leq M$, its leading eigenvalues provide a low-rank estimate of the dominant sensitivity directions.
At the output layer, this Jacobian is linked to posterior-covariance geometry through Eq.~\eqref{eq:miyasawa}. At intermediate layers, informative response directions may remain visible before being attenuated or canceled by downstream mappings; see Appendix~\ref{app:tap-readout-two-sheet} for an analysis in the two-sheet model and Appendix~\ref{app:tap-output-check} for empirical evidence. Appendix~\ref{app:feature-selection} provides the exact estimator and its efficient computation. This provides a signal complementary to reconstruction: $s_R(\vx):=s_{\mathrm{MSE}}(\vx)$ measures residual magnitude, whereas $s_J(\vx)$ measures local feature sensitivity.

\subsection{Feature selection and score construction}
\label{subsec:feature-selection}

Constructing the \method\ score requires choosing where to measure feature sensitivity and how to aggregate both signals across diffusion timesteps. We denote discrete DDPM timesteps by $\tau$, with larger $\tau$ corresponding to higher noise. For sensitivity, we use \emph{only} a held-out ID calibration set to select one decoder feature tap and the timesteps at which its response is most stable. The same calibration set is then used to standardize, aggregate, and combine the scores.

{\bf Feature-tap selection.}
We select the decoder tap using proxy shifts generated from the ID calibration data. The goal is to identify a tap whose sensitivity responds to changes in ID image structure, without using any target OOD samples. We generate these proxy shifts using the geometric transformations
$\mathcal{G}=\{\mathrm{rot}_{90},\mathrm{rot}_{180},\mathrm{rot}_{270},\mathrm{vflip}\}$ and  treat the transformed images only as a self-supervised stress test.
Let $\mathcal{P}$ denote the candidate sensitivity timesteps and $e_{\ell,\tau}(\vx)$ the sensitivity score at tap $\ell$ and timestep $\tau$. For each tap $\ell$, transformation $g\in\mathcal{G}$, and timestep $\tau\in\mathcal{P}$, we compute the AUROC for separating the scores of the original images, $e_{\ell,\tau}(\vx)$, from those of the transformed images, $e_{\ell,\tau}(g(\vx))$, treating the transformed images as the positive class. We average the resulting AUROC across transformations and timesteps and select the tap $\ell^\star$ with the highest average score.

{\bf Timestep selection.}
After selecting $\ell^\star$, we choose the sensitivity timesteps using an ID-stability criterion. For each candidate timestep $\tau\in\mathcal{P}$, we compute the sensitivity scores of the untransformed ID calibration samples and measure their coefficient of variation, defined as the standard deviation divided by the absolute mean. We retain the $K_{\mathrm{ens}}$ timesteps with the smallest coefficient of variation, denoted by $\mathcal{T}^\star$, and aggregate their sensitivity scores using normalized inverse-CV weights $w_\tau$. This assigns greater weight to more stable timesteps. Following the ID-only selection principle of GEPC~\citep{rouzoumka2026gepc}, proxy shifts are used to select the tap, while timestep selection and final calibration use untransformed ID data. We restrict the candidate pool \(\mathcal{P}\) to low-to-intermediate noise levels, where enough input-specific structure remains for sensitivity to be informative; at high noise, this structure is largely washed out. This behavior is analyzed in Proposition~\ref{prop:high-noise-collapse} and evaluated empirically in Appendix~\ref{app:feature-selection}.

{\bf Timestep aggregation and ID calibration.}
Because score scales can vary across timesteps, we first standardize each
per-timestep score using the ID calibration set. We write $\operatorname{std}(u;\mu,\sigma):=(u-\mu)/(\sigma+\eta)$ 
where $\eta>0$ is a small numerical constant. Let
$(\mu_\tau^J,\varsigma_\tau^J)$ and
$(\mu_\tau^R,\varsigma_\tau^R)$ denote the ID means and standard deviations
of the sensitivity score $e_{\ell^\star,\tau}$ and reconstruction score
$e_\tau^{\mathrm{rec}}$, respectively. We aggregate the standardized scores as
\begin{equation}
\widetilde{s}_J(\vx)
=
\sum_{\tau\in\mathcal{T}^\star}
w_\tau
\operatorname{std}
\left(
e_{\ell^\star,\tau}(\vx);
\mu_\tau^J,\varsigma_\tau^J
\right),
\qquad
\widetilde{s}_R(\vx)
=
\sum_{\tau\in\mathcal{R}}
\operatorname{std}
\left(
e_\tau^{\mathrm{rec}}(\vx);
\mu_\tau^R,\varsigma_\tau^R
\right),
\label{eq:aggregate-channels}
\end{equation}
where $\mathcal{R}$ is the fixed reconstruction timestep set. We then standardize the aggregated scores using their means and standard
deviations on the ID calibration set:
\begin{equation}
z_R(\vx)
=
\frac{\widetilde{s}_R(\vx)-\mu_R}{\varsigma_R+\eta},
\qquad
z_J(\vx)
=
\frac{\widetilde{s}_J(\vx)-\mu_J}{\varsigma_J+\eta}.
\label{eq:zchannels}
\end{equation}
This calibration places the aggregated reconstruction and sensitivity
scores on comparable scales.

{\bf Reconstruction--sensitivity fusion.}
The final \method\ score combines the calibrated reconstruction and
sensitivity scores
\begin{equation}
S_\gamma(\vx)
=
\gamma z_R(\vx)
+
(1-\gamma)z_J(\vx),
\qquad
\gamma\in[0,1].
\label{eq:fuse}
\end{equation}
Larger values indicate stronger OOD evidence. The parameter $\gamma$ controls the relative contribution of the two signals:
$\gamma=1$ uses only reconstruction, $\gamma=0$ uses only sensitivity, and
intermediate values combine both. We select $\gamma$ using only ID calibration
data, as described in Appendix~\ref{app:id-gamma}. Algorithms~\ref{alg:calibrate} and~\ref{alg:detect} summarize the complete calibration and scoring
procedure, respectively. Algorithm~\ref{alg:selection} in
Appendix~\ref{app:feature-selection} gives the ID-only tap and timestep
selection procedure.

\begin{figure}[t]
\centering

\begin{minipage}[t]{0.49\linewidth}
\vspace{0pt}
\renewcommand{\baselinestretch}{1.025}\selectfont
\begin{algorithm}[H]
\caption{\textsc{\method}: ID-calibration}
\label{alg:calibrate}
\small
\begin{algorithmic}[1]
\Require ID denoiser $D_\theta$;
calibration set $\mathcal{X}_{\mathrm{cal}}$;
selected tap $\ell^\star$;
sensitivity timesteps $\mathcal{T}^\star$ and weights $\{w_\tau\}$;
reconstruction timesteps $\mathcal{R}$
\Ensure Calibration statistics $\mathcal{C}$

\State Estimate $(\mu_\tau^J,\varsigma_\tau^J)$ for every
$\tau\in\mathcal{T}^\star$ from $\mathcal{X}_{\mathrm{cal}}$

\State Estimate $(\mu_\tau^R,\varsigma_\tau^R)$ for every
$\tau\in\mathcal{R}$ from $\mathcal{X}_{\mathrm{cal}}$

\For{$\vx\in\mathcal{X}_{\mathrm{cal}}$}
    \State Compute $\widetilde{s}_J(\vx)$ and
    $\widetilde{s}_R(\vx)$ using
    Eq.~\eqref{eq:aggregate-channels}
\EndFor

\State $(\mu_J,\varsigma_J)$ and
$(\mu_R,\varsigma_R)$ from
$\{\widetilde{s}_J(\vx)\}$ and
$\{\widetilde{s}_R(\vx)\}$

\State Store all estimated statistics in $\mathcal{C}$
\State \Return $\mathcal{C}$
\end{algorithmic}
\end{algorithm}
\end{minipage}
\hfill
\begin{minipage}[t]{0.49\linewidth}
\vspace{0pt}
\begin{algorithm}[H]
\caption{\textsc{\method}: Test-time OOD scoring}
\label{alg:detect}
\small
\begin{algorithmic}[1]
\Require ID denoiser $D_\theta$;
test set $\mathcal{X}_{\mathrm{te}}$;
selected tap $\ell^\star$;
sensitivity timesteps $\mathcal{T}^\star$ and weights
$\{w_\tau\}_{\tau\in\mathcal{T}^\star}$;
reconstruction timesteps $\mathcal{R}$;
fusion weight $\gamma$;
calibration statistics $\mathcal{C}$
\Ensure OOD scores
$\{S_\gamma(\vx)\}_{\vx\in\mathcal{X}_{\mathrm{te}}}$

\For{$\vx\in\mathcal{X}_{\mathrm{te}}$}
    \State Compute
$\{e_\tau^{\mathrm{rec}}(\vx)\}_{\tau\in\mathcal{R}}$
and
$\{e_{\ell^\star,\tau}(\vx)\}_{\tau\in\mathcal{T}^\star}$
using the per-timestep forms of
Eqs.~\eqref{eq:mse-ood-score} and~\eqref{eq:sJ}

    \State Compute $\widetilde{s}_R(\vx)$ and
    $\widetilde{s}_J(\vx)$ using
    Eq.~\eqref{eq:aggregate-channels}

    \State Compute $z_R(\vx)$ and $z_J(\vx)$ using
    Eq.~\eqref{eq:zchannels}

    \State Compute $S_\gamma(\vx)$ using
    Eq.~\eqref{eq:fuse}
\EndFor

\State \Return
$\{S_\gamma(\vx)\}_{\vx\in\mathcal{X}_{\mathrm{te}}}$
\end{algorithmic}
\end{algorithm}
\end{minipage}

\end{figure}

\section{Analysis of the Dual-Readout Design}
\label{sec:theory}
We analyze three design choices behind \method. First, a simple data model shows that Bayes denoiser sensitivity can remain informative when reconstruction is unreliable. Second, we explain how intermediate features can contain sensitivity that downstream mixing suppresses. Third, we characterize when combining the calibrated reconstruction and sensitivity signals improves ID--OOD separation.

\subsection{Branch ambiguity: what remains when reconstruction inverts}
\label{subsec:theory-molrg}

\begin{wrapfigure}[13]{r}{0.51\linewidth}
\vspace{-5mm}
    \centering
    \includegraphics[width=0.75\linewidth]{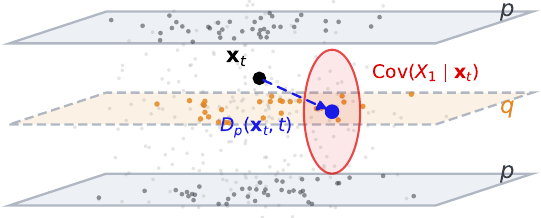}
    \caption{
\textbf{Why reconstruction fails while posterior uncertainty remains informative.} ID samples from $p$ (dark) lie on two sheets and OOD samples from $q$ (orange) on their midplane. Under heavy noise, the denoiser predicts $D_p(\vx_t,t)$ near the midplane, so OOD samples reconstruct well. The posterior covariance $\operatorname{Cov}(X_1\mid\vx_t)$ (red) still captures the direction between the sheets.
    }
    \label{fig:branch-intuition}
\end{wrapfigure}

Figure~\ref{fig:branch-intuition} illustrates a simple reconstruction failure mode. The ID consists of two parallel sheets, while the OOD distribution lies on their midplane. When noise makes the originating ID sheet difficult to identify, the posterior-mean denoiser averages across the two branches and moves samples toward the midplane. As a result, OOD samples on the midplane can be reconstructed more accurately than ID samples, reversing the desired reconstruction-score ordering. However, the unresolved ambiguity between the ID branches remains visible in the posterior covariance as spectral mass along the direction separating them. 
The following theorem formalizes this contrast between reconstruction error and the posterior-covariance spectrum.

\begin{theorem}[Branch dichotomy]
\label{thm:branch-dichotomy}
Let the ID $p$ be an equal mixture of two Gaussians supported on parallel
$r$-dimensional affine subspaces separated by $\omega$. Let the OOD $q$
be the corresponding Gaussian distribution on their midplane, with the
same within-subspace covariance. Thus, $p$ and $q$ differ in branch
structure but have the same intrinsic dimension and within-subspace
geometry.

Define $\zeta_t:=\omega\alpha_t/\beta_t$, which measures the separation
between the ID sheets relative to the noise level. Let
$\Delta_{\mathrm{spec}}(t)$ denote the OOD--ID difference in the
expected posterior-covariance spectral score; thus,
$\Delta_{\mathrm{spec}}(t)>0$ indicates higher scores for OOD samples.
There exist thresholds $0<\zeta_-<\zeta_+<\infty$ such that:
\begin{enumerate}
    \item[(i)] \textbf{Reconstruction.}
    When $\zeta_t\le\zeta_-$, the expected reconstruction error is lower
    for OOD than for ID, so $\Delta_{\mathrm{MSE}}(t)<0$.
    When $\zeta_t\ge\zeta_+$, the ordering is restored and
    $\Delta_{\mathrm{MSE}}(t)>0$.

    \item[(ii)] \textbf{Covariance spectrum.}
    For every interior noise level $t$, the spectral gap satisfies
    $\Delta_{\mathrm{spec}}(t)>0$ when computed using the covariance
    trace, or any top-$K_{\mathrm{spec}}$ spectral mass satisfying
    $\lambda_{K_{\mathrm{spec}}}(C_t)<\omega^2/4$,
    where $C_t$ is the within-sheet posterior covariance and its
    eigenvalues are ordered decreasingly.
\end{enumerate}
\end{theorem}

\noindent 
When the noise is large relative to the distance between the two ID sheets, a noisy input may have originated from either sheet. The denoiser therefore averages the two possible clean samples and predicts near the OOD midplane, allowing OOD samples to have lower reconstruction error than ID samples. With less noise, the correct sheet becomes easier to identify and the reconstruction ordering is restored. Even when reconstruction fails, the posterior covariance retains this ambiguity as variation along the direction separating the sheets, which the covariance spectral score captures.
This mechanism applies more generally to clustered ID distributions. When a noisy input is compatible with several ID components, the posterior covariance captures both uncertainty within each component and uncertainty about which component generated the input. Through Miyasawa's identity~\eqref{eq:miyasawa}, this uncertainty also appears in the spectrum of the Bayes denoiser's output Jacobian, motivating sensitivity as an OOD signal. Appendix~\ref{app:molrg-diagnostics} gives the proof and the posterior-covariance decomposition for Gaussian mixtures.

\subsection{From posterior ambiguity to intermediate feature sensitivity}
\label{subsec:theory-feature-sensitivity}

Theorem~\ref{thm:branch-dichotomy} shows that posterior covariance can remain informative when reconstruction fails. Miyasawa's identity~\eqref{eq:miyasawa} connects this covariance to the Jacobian of the Bayes denoiser. \method\ measures sensitivity at an intermediate feature because subsequent layers may weaken informative responses before they reach the output.
A downstream readout can discard feature sensitivity: in the two-sheet model, a near-optimal denoiser can reverse the output's sensitivity ordering while the feature's is unchanged (Appendix~\ref{app:tap-readout-two-sheet}), so we measure it before the output. As the analysis cannot identify which feature a trained network computes, we select the tap empirically and compare it with the output in Appendix~\ref{app:tap-output-check}.
The argument does not imply that every intermediate feature is informative.
\method\ therefore selects the tap using only ID calibration data, as described in Section~\ref{sec:method}.

\subsection{Interpreting the calibrated fusion}
\label{subsec:theory-fusion}
We combine the standardized reconstruction and sensitivity scores using Eq.~\eqref{eq:fuse}. To understand when this combination helps, let $Z_R=z_R(X_1)$ and $Z_J=z_J(X_1)$ for a clean test input $X_1$ drawn from $p$ or $q$. Both scores are standardized to have mean zero and variance one on ID data.
Let
$\Delta_R = \mathbb{E}_q[Z_R]-\mathbb{E}_p[Z_R]$ and $\Delta_J = \mathbb{E}_q[Z_J]-\mathbb{E}_p[Z_J]$
measure how much each score changes from ID to OOD. Positive values mean that the corresponding score is higher for OOD samples. We also define $\rho=\operatorname{Corr}_p(Z_R,Z_J)$, which measures how similarly the two scores behave on ID data. To compare different fusion weights on a common scale, we measure how far the
OOD mean moves from the ID mean in units of the ID standard deviation:
$d(S)=\bigl(\mathbb{E}_q[S]-\mathbb{E}_p[S]\bigr)/\sqrt{\operatorname{Var}_p(S)}$.
We refer to $d(S)$ as the standardized mean gap. Larger values indicate better ID--OOD separation. For the fused score $S_\gamma = \gamma Z_R + (1-\gamma)Z_J$, this becomes
\begin{equation}
d(S_\gamma)
=
\frac{
\gamma\Delta_R+(1-\gamma)\Delta_J
}{
\sqrt{
\gamma^2+(1-\gamma)^2+
2\gamma(1-\gamma)\rho
}
}.
\label{eq:convex-deflection}
\end{equation}
The numerator measures the combined OOD signal, while the denominator measures its variation on ID data. Fusion is therefore most useful when both channels respond to OOD inputs but do not behave identically on ID data. If one channel provides no additional information, the best score relies mainly on the other.
Theorem~\ref{thm:two-channel-fusion} formalizes this result, and
Appendix~\ref{app:fusion-diagnostic} compares this separation measure with empirical AUROC.

\section{Experiments}
\label{sec:experiments}
We benchmark \method\ against state-of-the-art unsupervised OOD detectors across diverse ID--OOD pairs. Experiments demonstrate improved average detection performance, compare its computational cost with existing diffusion-based detectors, and assess robustness to hyperparameter choices.
\label{subsec:exp-setup}

\textbf{Datasets.} Our first benchmark includes CIFAR-10 and CIFAR-100~\citep{krizhevsky2009learning}, SVHN~\citep{netzer2011reading}, and CelebA~\citep{liu2015deep}. Each dataset serves as ID data against the other three, giving 12 ID--OOD pairs. We additionally evaluate near-OOD detection involving TinyImageNet and the CIFAR-10/CIFAR-100 pair (Appendix~\ref{app:near-ood}), high-resolution detection with FFHQ-256~\citep{karras2019stylegan} as ID against AFHQ~\citep{choi2020stargan} and Microscopy~\citep{chen2023chammi} (Appendix~\ref{app:highres}), and ImageNet-64 as ID against SVHN and Textures (Appendix~\ref{app:imagenet64}). Further experiments consider LSUN, iSUN, and Textures as OOD datasets for CIFAR-10 and CIFAR-100 (Appendix~\ref{app:additional-ood}). We also evaluate trajectory-level OOD detection on D4RL using frozen Diffuser models (Appendix~\ref{app:rl-planning}).

\textbf{Baselines.}
We compare \method{} with the likelihood-based methods
DoS~\citep{morningstar2021density},
TT~\citep{nalisnick2019detecting},
WAIC~\citep{choi2019waic}, NLL (Negative Log-Likelihood), and IC~\citep{serra2020input}, as well as the
diffusion-based detectors DDPM-OOD~\citep{graham2023denoising},
LMD~\citep{liu2023unsupervised},
DiffPath~\citep{heng2024diffpath},
DiffPathV2~\citep{abdi2025zero},
GEPC~\citep{rouzoumka2026gepc},
SCOPED~\citep{barkley2025scoped}, and
EigenScore~\citep{shoushtari2026eigenscore}. Implementation details,
hyperparameters, and protocols are provided in
Appendix~\ref{app:implementation}.

\textbf{Implementation.}
Unless otherwise stated, all experiments use $K_{\mathrm{spec}}=3$ leading eigenvalues, $M=32$ finite-difference probe directions, $N_\epsilon=1$ sensitivity noise draw per image and noise level, probe scale $\delta=0.1$, and $K_{\mathrm{ens}}=3$ selected sensitivity noise levels. These settings also apply to the experiments in Appendix~\ref{app:additional-experiments}. For each ID dataset, the fusion weight is computed using only ID calibration data and the plug-in rule described in Appendix~\ref{app:id-gamma}; the resulting weights are reported in Table~\ref{tab:id-gamma}. No hyperparameters are tuned using OOD data, and all remaining settings are fixed.

\begin{table}[t]
\caption{
Main OOD detection results measured by AUROC across 12 ID--OOD pairs. We compare \method\ with likelihood- and diffusion-based generative OOD detectors. The best result in each column is shown in \textbf{bold}, and the
second-best is \underline{underlined}. DUO achieves the highest average AUROC and ranks first or second in most settings.
}
\vspace{-3mm}
\label{tab:main-results}
\centering
\footnotesize
\resizebox{\textwidth}{!}{
\begin{tabular}{l ccc ccc ccc ccc c}
\toprule
InD & \multicolumn{3}{c}{\textbf{CelebA} vs.} & \multicolumn{3}{c}{\textbf{CIFAR-10} vs.} & \multicolumn{3}{c}{\textbf{CIFAR-100} vs.} & \multicolumn{3}{c}{\textbf{SVHN} vs.} & \textbf{Avg} \\
\cmidrule(lr){2-4}\cmidrule(lr){5-7}\cmidrule(lr){8-10}\cmidrule(lr){11-13}
OOD & CIFAR-10 & CIFAR-100 & SVHN & CIFAR-100 & SVHN & CelebA & CIFAR-10 & CelebA & SVHN & CIFAR-10 & CIFAR-100 & CelebA & \\
\midrule \\[-3.6ex] \midrule
DoS & 0.630 & 0.615 & 0.808 & 0.504 & 0.752 & 0.456 & 0.491 & 0.520 & 0.777 & 0.911 & 0.904 & 0.956 & 0.693 \\
TT & 0.676 & 0.655 & 0.773 & 0.558 & 0.714 & 0.469 & 0.538 & 0.464 & 0.648 & 0.957 & 0.961 & \underline{0.994} & 0.701 \\
WAIC & 0.589 & 0.569 & 0.793 & 0.476 & 0.760 & 0.469 & 0.502 & 0.530 & 0.782 & 0.978 & 0.974 & 0.955 & 0.698 \\
\midrule \\[-3.6ex] \midrule
\multicolumn{14}{c}{\emph{\textbf{Diffusion-based}}} \\
\midrule \\[-3.6ex] \midrule
NLL & 0.507 & 0.671 & 0.753 & 0.558 & 0.545 & 0.599 & 0.480 & 0.484 & 0.481 & 0.635 & 0.660 & 0.636 & 0.584 \\
IC & 0.510 & 0.673 & 0.755 & 0.552 & 0.540 & 0.583 & 0.460 & 0.466 & 0.469 & 0.625 & 0.653 & 0.625 & 0.576 \\
DDPM-OOD & 0.922 & 0.928 & \underline{0.992} & 0.618 & \underline{0.944} & 0.642 & 0.462 & 0.496 & \textbf{0.870} & 0.963 & 0.972 & \textbf{0.996} & 0.817 \\
LMD & 0.886 & 0.848 & 0.950 & 0.601 & 0.821 & 0.834 & 0.569 & 0.595 & 0.748 & 0.780 & 0.749 & 0.872 & 0.771 \\
SCOPED & 0.892 & 0.881 & 0.877 & 0.553 & 0.658 & 0.850 & 0.446 & \underline{0.821} & 0.582 & 0.800 & 0.780 & 0.932 & 0.756 \\
GEPC & 0.949 & 0.933 & 0.953 & 0.582 & 0.750 & \textbf{0.935} & 0.427 & 0.563 & 0.717 & 0.866 & 0.872 & 0.865 & 0.784 \\
DiffPath & \textbf{1.000} & \textbf{1.000} & 0.964 & 0.554 & 0.729 & \underline{0.885} & 0.475 & \textbf{0.887} & 0.724 & 0.919 & 0.941 & 0.328 & 0.784 \\
DiffPathV2 & \textbf{1.000} & 0.995 & 0.969 & 0.535 & 0.812 & 0.862 & 0.483 & 0.513 & 0.724 & 0.969 & 0.975 & 0.883 & 0.810 \\

EigenScore & \underline{0.965} & 0.944 & 0.888 & \textbf{0.880} & 0.810 & 0.873 & \underline{0.642} & 0.427 & 0.661 & \underline{0.992} & \underline{0.982} & \underline{0.994} & \underline{0.838} \\
\midrule
\textbf{\method\ (ours)} & \textbf{1.000} & \textbf{1.000} & \textbf{1.000} & \underline{0.876} & \textbf{0.986} & 0.872 & \textbf{0.657} & 0.544 & \underline{0.839} & \textbf{0.995} & \textbf{0.994} & \textbf{0.996} & \textbf{0.897} \\
\bottomrule
\end{tabular}
}
\end{table}

{\bf Main results.} Table~\ref{tab:main-results} shows that \method\ achieves the highest average AUROC among fully unsupervised methods, improving over the strongest baseline, EigenScore, from $0.838$ to $0.897$. The largest gains occur on challenging ID--OOD pairs, supporting the use of reconstruction and feature sensitivity as complementary signals. 
\method\ also achieves an average AUROC of 0.999 on the high-resolution FFHQ-256 benchmark, compared with 0.826 for GEPC (Appendix~\ref{app:highres}), and 0.835 on the near-OOD benchmark, compared with 0.794 for EigenScore
(Appendix~\ref{app:near-ood}). On ImageNet-64, it reaches 0.882, compared with 0.732 for EigenScore. Additional results confirm its performance across further
OOD datasets (Appendix~\ref{app:additional-ood}) and RL planning tasks (Appendix~\ref{app:rl-planning}).

{\bf Time complexity.} Table~\ref{tab:runtime} reports batch-1 inference costs. \method\ requires 197 neural function evaluations (NFEs) and 0.28 seconds per sample, substantially faster than EigenScore, DDPM-OOD, and LMD. GEPC and DiffPath have comparable or lower latency but substantially lower average AUROC.
DiffPath performs sequential integration along a DDIM-inverted diffusion path, whereas \method's probe evaluations are independent and can be parallelized. Consequently, \method\ achieves latency comparable to DiffPath despite using more NFEs. Overall, \method\ offers a strong accuracy--efficiency trade-off without EigenScore's costly iterative eigenspectrum estimation.

\begin{table}[t]
\caption{Runtime comparison of diffusion-based OOD detectors. We report NFE and single-image (batch-1) inference latency per sample, measured on one NVIDIA A100-SXM4-40GB.}
\vspace{-3mm}
\label{tab:runtime}
\centering
\small
\begin{tabular}{l cccccc}
\toprule
Method & LMD & DDPM-OOD & DiffPath & GEPC & EigenScore & \method\ (ours) \\
\midrule
NFE & 10{,}000 & 350 & 10 & 16 & 7{,}000 & 197 \\
Latency (s) & 190.85 & 3.62 & 0.29 & 0.15 & 9.48 & 0.28 \\
\bottomrule
\end{tabular}
\end{table}

{\bf Ablation study.} Appendices~\labelcref{app:estimator-ablation,app:layer-ablation,app:ablation-alpha-kens,app:variance} evaluate the main design choices of \method, including the probe count and scale, spectral rank, number of noise draws, feature tap, fusion weight, and ensemble size. The sensitivity estimator is robust to these choices: varying $M$, $\delta$, $K_{\mathrm{spec}}$, and $N_{\epsilon}$ keeps the average AUROC between 0.892 and 0.901, with similarly small variation across random seeds. Layer sweeps further show that intermediate decoder features outperform early and output-level features, and that the ID-only selection procedure reliably identifies taps in this informative region.

\section{Related work}

{\bf Likelihood-based methods.}
Likelihood-based OOD detectors score inputs under an ID-trained generative model. However, likelihood can be misaligned with typicality, allowing OOD inputs to receive higher likelihood than ID data~\citep{nalisnick2019deep,choi2019waic,ren2019likelihood}. This behavior has been attributed to input complexity~\citep{serra2020input} and the geometry of
high-dimensional densities~\citep{kamkari2024geometric}. \method\ avoids likelihood estimation and instead uses calibrated reconstruction and feature sensitivity from a frozen diffusion model.

{\bf Reconstruction-based methods.}
Reconstruction-based detectors measure how accurately an ID-trained model reconstructs a test input. Prior work strengthens this signal using latent distances, normalized or semantic residuals, and typicality corrections~\citep{denouden2018improving,zhou2022rethinking,osada2023out}. Diffusion-based methods use denoising errors across noise levels, manifold mappings, or layer-wise reconstructions~\citep{graham2023denoising,liu2023unsupervised,yang2024diffusion}. These approaches remain primarily determined by reconstruction magnitude. \method\ keeps reconstruction but adds feature sensitivity, which can detect shifts without large residuals.

{\bf Feature-based methods.}
Feature-based detectors evaluate how typical an input's representation is under the ID distribution. Common approaches use confidence or logit scores~\citep{hendrycks2016baseline,liang2017enhancing,liu2020energy}, explicit feature models~\citep{lee2018simple,sun2022out,wang2022vim}, or activation shaping and pruning~\citep{sun2021react,sun2022dice}. Intermediate features can also provide complementary OOD information without OOD validation data~\citep{de2025mysteries}. Unlike methods that score static feature values or classifier outputs, \method\ measures how intermediate diffusion features respond to local input perturbations.

{\bf Diffusion-geometry and trajectory methods.}
Recent diffusion-based detectors extract OOD signals from the geometry or dynamics of a frozen denoiser. These include DiffPath and DiffPathV2~\citep{heng2024diffpath,abdi2025zero}, EigenScore~\citep{shoushtari2026eigenscore}, GEPC~\citep{rouzoumka2026gepc}, and SCOPED~\citep{barkley2025scoped}. \method\ differs by measuring the sensitivity spectrum of intermediate denoising features and combining it with a calibrated reconstruction score. \citet{jeon2025sharpness} instead use the output-level Hessian spectrum to detect memorized training samples, whereas \method\ uses intermediate-feature Jacobians for OOD detection.

\section{Limitations and future work}
We presented \method, a training-free and label-free OOD detector built from a frozen diffusion model. The method reads the model through two complementary channels: a reconstruction channel, which measures the denoising residual, and a feature-sensitivity channel, which measures the local response of intermediate denoising features. Both channels are calibrated using ID data only and fused into a single one-sided OOD score. However, the method requires a diffusion model trained on the ID distribution. This can limit practical use when ID data are scarce or when training a diffusion model is expensive. Extending the same readout to foundation diffusion models, with only lightweight ID calibration or adaptation, is an important direction for future work.

A second limitation is that the sensitivity channel requires additional forward passes to estimate feature-Jacobian spectra. Although substantially cheaper than power-iteration-based methods, it is still more expensive than a single reconstruction score. Future work could reduce this cost with adaptive probe allocation, structured perturbations, or analytic Jacobian-vector products.

\section*{Reproducibility Statement }
We have made several efforts to support the reproducibility of our experiments. The datasets used in this work, including CIFAR-10, CIFAR-100, SVHN, CelebA, TinyImageNet, and others, are publicly available. We describe the main experimental settings, such as the model architecture, noise schedule, hyperparameters, and aggregation procedure. Results are reported over multiple random seeds, and the sensitivity to important parameters is examined in Appendices~\labelcref{app:estimator-ablation,app:layer-ablation,app:ablation-alpha-kens,app:variance}. We also include the source code and experiment scripts in the supplementary material to make our results easier to reproduce.
\section*{AI use statement}
Generative AI tools were used to assist with drafting and polishing parts of the manuscript, including improving wording, grammar, clarity, and readability. All AI-assisted content was reviewed and edited by the authors. The authors take full responsibility for the final content of the paper.

\bibliography{refs}

\begin{thebibliography}{46}
\providecommand{\natexlab}[1]{#1}
\providecommand{\url}[1]{\texttt{#1}}
\expandafter\ifx\csname urlstyle\endcsname\relax
  \providecommand{\doi}[1]{doi: #1}\else
  \providecommand{\doi}{doi: \begingroup \urlstyle{rm}\Url}\fi

\bibitem[Abdi et~al.(2025)Abdi, Valiuddin, Caetano, Viviers, and van~der
  Sommen]{abdi2025zero}
Lemar Abdi, Amaan Valiuddin, Francisco Caetano, Christiaan Viviers, and Fons
  van~der Sommen.
\newblock Zero-shot image anomaly detection using generative foundation models.
\newblock In \emph{Proceedings of the IEEE/CVF International Conference on
  Computer Vision (ICCV) Workshops}, pp.\  3604--3613, 2025.

\bibitem[Barkley et~al.(2026)Barkley, Culbertson, and
  Fridovich-Keil]{barkley2025scoped}
Brett Barkley, Preston Culbertson, and David Fridovich-Keil.
\newblock {SCOPED}: Score-curvature out-of-distribution proximity evaluator for
  diffusion.
\newblock In \emph{International Conference on Learning Representations
  (ICLR)}, 2026.

\bibitem[Chen et~al.(2023)Chen, Pham, Wang, Doron, Moshkov, Plummer, and
  Caicedo]{chen2023chammi}
Zitong~Sam Chen, Chau Pham, Siqi Wang, Michael Doron, Nikita Moshkov, Bryan
  Plummer, and Juan~C Caicedo.
\newblock Chammi: A benchmark for channel-adaptive models in microscopy
  imaging.
\newblock \emph{Advances in Neural Information Processing Systems},
  36:\penalty0 19700--19713, 2023.

\bibitem[Choi et~al.(2018)Choi, Jang, and Alemi]{choi2019waic}
Hyunsun Choi, Eric Jang, and Alexander~A. Alemi.
\newblock {WAIC}, but why? generative ensembles for robust anomaly detection.
\newblock \emph{arXiv preprint arXiv:1810.01392}, 2018.

\bibitem[Choi et~al.(2020)Choi, Uh, Yoo, and Ha]{choi2020stargan}
Yunjey Choi, Youngjung Uh, Jaejun Yoo, and Jung-Woo Ha.
\newblock Stargan v2: Diverse image synthesis for multiple domains.
\newblock In \emph{Proceedings of the IEEE/CVF conference on computer vision
  and pattern recognition}, pp.\  8188--8197, 2020.

\bibitem[De~la Jara et~al.(2025)De~la Jara, Rodriguez-Opazo, Teney, Ranasinghe,
  and Abbasnejad]{de2025mysteries}
Ignacio~Meza De~la Jara, Cristian Rodriguez-Opazo, Damien Teney, Damith
  Ranasinghe, and Ehsan Abbasnejad.
\newblock Mysteries of the deep: Role of intermediate representations in out of
  distribution detection.
\newblock \emph{arXiv preprint arXiv:2510.05782}, 2025.

\bibitem[Denouden et~al.(2018)Denouden, Salay, Czarnecki, Abdelzad, Phan, and
  Vernekar]{denouden2018improving}
Taylor Denouden, Rick Salay, Krzysztof Czarnecki, Vahdat Abdelzad, Buu Phan,
  and Sachin Vernekar.
\newblock Improving reconstruction autoencoder out-of-distribution detection
  with mahalanobis distance.
\newblock \emph{arXiv preprint arXiv:1812.02765}, 2018.

\bibitem[Dhariwal \& Nichol(2021)Dhariwal and Nichol]{dhariwal2021diffusion}
Prafulla Dhariwal and Alexander~Quinn Nichol.
\newblock Diffusion models beat {GANs} on image synthesis.
\newblock In \emph{Advances in Neural Information Processing Systems
  (NeurIPS)}, 2021.

\bibitem[Fu et~al.(2020)Fu, Kumar, Nachum, Tucker, and Levine]{fu2020d4rl}
Justin Fu, Aviral Kumar, Ofir Nachum, George Tucker, and Sergey Levine.
\newblock {D4RL}: Datasets for deep data-driven reinforcement learning.
\newblock \emph{arXiv preprint arXiv:2004.07219}, 2020.

\bibitem[Graham et~al.(2023)Graham, Pinaya, Tudosiu, Nachev, Ourselin, and
  Cardoso]{graham2023denoising}
Mark~S Graham, Walter~HL Pinaya, Petru-Daniel Tudosiu, Parashkev Nachev,
  Sebastien Ourselin, and M.~Jorge Cardoso.
\newblock Denoising diffusion models for out-of-distribution detection.
\newblock In \emph{Proceedings of the IEEE/CVF Conference on Computer Vision
  and Pattern Recognition (CVPR) Workshops}, pp.\  2948--2957, 2023.

\bibitem[Hendrycks \& Gimpel(2016)Hendrycks and Gimpel]{hendrycks2016baseline}
Dan Hendrycks and Kevin Gimpel.
\newblock A baseline for detecting misclassified and out-of-distribution
  examples in neural networks.
\newblock \emph{arXiv preprint arXiv:1610.02136}, 2016.

\bibitem[Heng et~al.(2024)Heng, Thiery, and Soh]{heng2024diffpath}
Alvin Heng, Alexandre~H. Thiery, and Harold Soh.
\newblock Out-of-distribution detection with a single unconditional diffusion
  model.
\newblock In \emph{Advances in Neural Information Processing Systems
  (NeurIPS)}, 2024.

\bibitem[Janner et~al.(2022)Janner, Du, Tenenbaum, and
  Levine]{janner2022diffuser}
Michael Janner, Yilun Du, Joshua~B. Tenenbaum, and Sergey Levine.
\newblock Planning with diffusion for flexible behavior synthesis.
\newblock In \emph{International Conference on Machine Learning (ICML)}, 2022.

\bibitem[Jeon et~al.(2025)Jeon, Kim, and No]{jeon2025sharpness}
Dongjae Jeon, Dueun Kim, and Albert No.
\newblock Understanding and mitigating memorization in generative models via
  sharpness of probability landscapes.
\newblock In \emph{International Conference on Machine Learning (ICML)}, 2025.

\bibitem[Kadkhodaie et~al.(2024)Kadkhodaie, Guth, Simoncelli, and
  Mallat]{kadkhodaie2024generalization}
Zahra Kadkhodaie, Florentin Guth, Eero~P. Simoncelli, and St{\'e}phane Mallat.
\newblock Generalization in diffusion models arises from geometry-adaptive
  harmonic representations.
\newblock In \emph{International Conference on Learning Representations
  (ICLR)}, 2024.

\bibitem[Kamb \& Ganguli(2025)Kamb and Ganguli]{kamb2024creativity}
Mason Kamb and Surya Ganguli.
\newblock An analytic theory of creativity in convolutional diffusion models.
\newblock In \emph{International Conference on Machine Learning (ICML)},
  Proceedings of Machine Learning Research, 2025.

\bibitem[Kamkari et~al.(2024)Kamkari, Ross, Cresswell, Caterini, Krishnan, and
  Loaiza-Ganem]{kamkari2024geometric}
Hamidreza Kamkari, Brendan~Leigh Ross, Jesse~C. Cresswell, Anthony~L. Caterini,
  Rahul~G. Krishnan, and Gabriel Loaiza-Ganem.
\newblock A geometric explanation of the likelihood {OOD} detection paradox.
\newblock In \emph{International Conference on Machine Learning (ICML)},
  Proceedings of Machine Learning Research, 2024.

\bibitem[Karras et~al.(2019)Karras, Laine, and Aila]{karras2019stylegan}
Tero Karras, Samuli Laine, and Timo Aila.
\newblock A style-based generator architecture for generative adversarial
  networks.
\newblock In \emph{Proceedings of the IEEE/CVF Conference on Computer Vision
  and Pattern Recognition}, pp.\  4401--4410, 2019.

\bibitem[Karras et~al.(2022)Karras, Aittala, Aila, and
  Laine]{karras2022elucidating}
Tero Karras, Miika Aittala, Timo Aila, and Samuli Laine.
\newblock Elucidating the design space of diffusion-based generative models.
\newblock \emph{Advances in neural information processing systems},
  35:\penalty0 26565--26577, 2022.

\bibitem[Krizhevsky(2009)]{krizhevsky2009learning}
Alex Krizhevsky.
\newblock Learning multiple layers of features from tiny images.
\newblock Technical report, University of Toronto, 2009.
\newblock URL
  \url{https://www.cs.toronto.edu/~kriz/learning-features-2009-TR.pdf}.

\bibitem[Lee et~al.(2018)Lee, Lee, Lee, and Shin]{lee2018simple}
Kimin Lee, Kibok Lee, Honglak Lee, and Jinwoo Shin.
\newblock A simple unified framework for detecting out-of-distribution samples
  and adversarial attacks.
\newblock \emph{Advances in neural information processing systems}, 31, 2018.

\bibitem[Liang et~al.(2017)Liang, Li, and Srikant]{liang2017enhancing}
Shiyu Liang, Yixuan Li, and Rayadurgam Srikant.
\newblock Enhancing the reliability of out-of-distribution image detection in
  neural networks.
\newblock \emph{arXiv preprint arXiv:1706.02690}, 2017.

\bibitem[Liu et~al.(2020)Liu, Wang, Owens, and Li]{liu2020energy}
Weitang Liu, Xiaoyun Wang, John Owens, and Yixuan Li.
\newblock Energy-based out-of-distribution detection.
\newblock \emph{Advances in neural information processing systems},
  33:\penalty0 21464--21475, 2020.

\bibitem[Liu et~al.(2023)Liu, Zhou, Wang, and Weinberger]{liu2023unsupervised}
Zhenzhen Liu, Jin~Peng Zhou, Yufan Wang, and Kilian~Q Weinberger.
\newblock Unsupervised out-of-distribution detection with diffusion inpainting.
\newblock In \emph{International Conference on Machine Learning}, pp.\
  22528--22538. PMLR, 2023.

\bibitem[Liu et~al.(2015)Liu, Luo, Wang, and Tang]{liu2015deep}
Ziwei Liu, Ping Luo, Xiaogang Wang, and Xiaoou Tang.
\newblock Deep learning face attributes in the wild.
\newblock In \emph{Proceedings of the IEEE International Conference on Computer
  Vision}, pp.\  3730--3738, 2015.
\newblock \doi{10.1109/ICCV.2015.425}.

\bibitem[Manor \& Michaeli(2024)Manor and Michaeli]{manor2024posterior}
Hila Manor and Tomer Michaeli.
\newblock On the posterior distribution in denoising: Application to
  uncertainty quantification.
\newblock In \emph{International Conference on Learning Representations
  (ICLR)}, 2024.

\bibitem[Miyasawa(1961)]{miyasawa1961empirical}
Koichi Miyasawa.
\newblock An empirical {B}ayes estimator of the mean of a normal population.
\newblock \emph{Bulletin of the International Statistical Institute},
  38\penalty0 (4):\penalty0 181--188, 1961.

\bibitem[Morningstar et~al.(2021)Morningstar, Ham, Gallagher, Lakshminarayanan,
  Alemi, and Dillon]{morningstar2021density}
Warren Morningstar, Cusuh Ham, Andrew Gallagher, Balaji Lakshminarayanan, Alex
  Alemi, and Joshua Dillon.
\newblock Density of states estimation for out of distribution detection.
\newblock In \emph{International Conference on Artificial Intelligence and
  Statistics}, pp.\  3232--3240. PMLR, 2021.

\bibitem[Nalisnick et~al.(2019{\natexlab{a}})Nalisnick, Matsukawa, Teh, Gorur,
  and Lakshminarayanan]{nalisnick2019deep}
Eric Nalisnick, Akihiro Matsukawa, Yee~Whye Teh, Dilan Gorur, and Balaji
  Lakshminarayanan.
\newblock Do deep generative models know what they don't know?
\newblock In \emph{International Conference on Learning Representations
  (ICLR)}, 2019{\natexlab{a}}.

\bibitem[Nalisnick et~al.(2019{\natexlab{b}})Nalisnick, Matsukawa, Teh, and
  Lakshminarayanan]{nalisnick2019detecting}
Eric Nalisnick, Akihiro Matsukawa, Yee~Whye Teh, and Balaji Lakshminarayanan.
\newblock Detecting out-of-distribution inputs to deep generative models using
  typicality.
\newblock \emph{arXiv preprint arXiv:1906.02994}, 2019{\natexlab{b}}.

\bibitem[Netzer et~al.(2011)Netzer, Wang, Coates, Bissacco, Wu, and
  Ng]{netzer2011reading}
Yuval Netzer, Tao Wang, Adam Coates, Alessandro Bissacco, Bo~Wu, and Andrew~Y.
  Ng.
\newblock Reading digits in natural images with unsupervised feature learning.
\newblock In \emph{NIPS Workshop on Deep Learning and Unsupervised Feature
  Learning}, 2011.

\bibitem[Nichol \& Dhariwal(2021)Nichol and Dhariwal]{nichol2021improved}
Alexander~Quinn Nichol and Prafulla Dhariwal.
\newblock Improved denoising diffusion probabilistic models.
\newblock In \emph{International Conference on Machine Learning (ICML)}, 2021.

\bibitem[Osada et~al.(2023)Osada, Takahashi, Ahsan, and Nishide]{osada2023out}
Genki Osada, Tsubasa Takahashi, Budrul Ahsan, and Takashi Nishide.
\newblock Out-of-distribution detection with reconstruction error and
  typicality-based penalty.
\newblock In \emph{Proceedings of the IEEE/CVF Winter Conference on
  Applications of Computer Vision}, pp.\  5551--5563, 2023.

\bibitem[Ren et~al.(2019)Ren, Liu, Fertig, Snoek, Poplin, Depristo, Dillon, and
  Lakshminarayanan]{ren2019likelihood}
Jie Ren, Peter~J Liu, Emily Fertig, Jasper Snoek, Ryan Poplin, Mark Depristo,
  Joshua Dillon, and Balaji Lakshminarayanan.
\newblock Likelihood ratios for out-of-distribution detection.
\newblock \emph{Advances in neural information processing systems}, 32, 2019.

\bibitem[Rouzoumka et~al.(2026)Rouzoumka, Pinsolle, Terreaux, Morisseau,
  Ovarlez, and Ren]{rouzoumka2026gepc}
Yadang~Alexis Rouzoumka, Jean Pinsolle, Eug{\'e}nie Terreaux, Christ{\`e}le
  Morisseau, Jean-Philippe Ovarlez, and Chengfang Ren.
\newblock Gepc: Group-equivariant posterior consistency for out-of-distribution
  detection in diffusion models.
\newblock \emph{arXiv preprint arXiv:2602.00191}, 2026.

\bibitem[Serr{\`a} et~al.(2020)Serr{\`a}, {\'A}lvarez, G{\'o}mez, Slizovskaia,
  N{\'u}{\~n}ez, and Luque]{serra2020input}
Joan Serr{\`a}, David {\'A}lvarez, Vicen{\c{c}} G{\'o}mez, Olga Slizovskaia,
  Jos{\'e}~F. N{\'u}{\~n}ez, and Jordi Luque.
\newblock Input complexity and out-of-distribution detection with
  likelihood-based generative models.
\newblock In \emph{International Conference on Learning Representations
  (ICLR)}, 2020.

\bibitem[Shoushtari et~al.(2026)Shoushtari, Wang, Shi, Asif, and
  Kamilov]{shoushtari2026eigenscore}
Shirin Shoushtari, Yi~Wang, Xiao Shi, M.~Salman Asif, and Ulugbek~S. Kamilov.
\newblock {EigenScore}: {OOD} detection using posterior covariance in diffusion
  models.
\newblock In \emph{International Conference on Learning Representations
  (ICLR)}, 2026.

\bibitem[Sun \& Li(2022)Sun and Li]{sun2022dice}
Yiyou Sun and Yixuan Li.
\newblock Dice: Leveraging sparsification for out-of-distribution detection.
\newblock In \emph{European conference on computer vision}, pp.\  691--708.
  Springer, 2022.

\bibitem[Sun et~al.(2021)Sun, Guo, and Li]{sun2021react}
Yiyou Sun, Chuan Guo, and Yixuan Li.
\newblock React: Out-of-distribution detection with rectified activations.
\newblock \emph{Advances in neural information processing systems},
  34:\penalty0 144--157, 2021.

\bibitem[Sun et~al.(2022)Sun, Ming, Zhu, and Li]{sun2022out}
Yiyou Sun, Yifei Ming, Xiaojin Zhu, and Yixuan Li.
\newblock Out-of-distribution detection with deep nearest neighbors.
\newblock In \emph{International conference on machine learning}, pp.\
  20827--20840. PMLR, 2022.

\bibitem[Vincent(2011)]{vincent2011connection}
Pascal Vincent.
\newblock A connection between score matching and denoising autoencoders.
\newblock \emph{Neural Computation}, 23\penalty0 (7):\penalty0 1661--1674,
  2011.

\bibitem[Wang et~al.(2022)Wang, Li, Feng, and Zhang]{wang2022vim}
Haoqi Wang, Zhizhong Li, Litong Feng, and Wayne Zhang.
\newblock Vim: Out-of-distribution with virtual-logit matching.
\newblock In \emph{Proceedings of the IEEE/CVF conference on computer vision
  and pattern recognition}, pp.\  4921--4930, 2022.

\bibitem[Wang et~al.(2026)Wang, Zhang, Zhang, Chen, Ma, and
  Qu]{wang2025lowrank}
Peng Wang, Huijie Zhang, Zekai Zhang, Siyi Chen, Yi~Ma, and Qing Qu.
\newblock Breaking the curse of dimensionality: Diffusion models efficiently
  learn low-dimensional distributions.
\newblock \emph{Journal of Machine Learning Research (JMLR)}, 2026.
\newblock To appear. arXiv:2409.02426.

\bibitem[Yang et~al.(2024{\natexlab{a}})Yang, Zhou, Li, and
  Liu]{yang2024generalized}
Jingkang Yang, Kaiyang Zhou, Yixuan Li, and Ziwei Liu.
\newblock Generalized out-of-distribution detection: A survey.
\newblock \emph{International Journal of Computer Vision}, 132\penalty0
  (12):\penalty0 5635--5662, 2024{\natexlab{a}}.

\bibitem[Yang et~al.(2024{\natexlab{b}})Yang, Cheng, Fang, Wang, Jiao, Cheng,
  Wang, and Gao]{yang2024diffusion}
Ying Yang, De~Cheng, Chaowei Fang, Yubiao Wang, Changzhe Jiao, Lechao Cheng,
  Nannan Wang, and Xinbo Gao.
\newblock Diffusion-based layer-wise semantic reconstruction for unsupervised
  out-of-distribution detection.
\newblock \emph{Advances in Neural Information Processing Systems},
  37:\penalty0 26846--26871, 2024{\natexlab{b}}.

\bibitem[Zhou(2022)]{zhou2022rethinking}
Yibo Zhou.
\newblock Rethinking reconstruction autoencoder-based out-of-distribution
  detection.
\newblock In \emph{Proceedings of the IEEE/CVF Conference on Computer Vision
  and Pattern Recognition (CVPR)}, pp.\  7379--7387, 2022.

\end{thebibliography}
\bibliographystyle{iclr2027_conference}

\newpage
\appendix
\section*{Appendix overview}
The appendix has five parts.
\begin{itemize}
\item \textbf{Appendix~\ref{app:proofs}} contains the derivations and
proofs behind Section~\ref{sec:method} and
Section~\ref{subsec:theory-fusion}: the Tweedie and Miyasawa identities
(\ref{app:tweedie-miyasawa}), the proof of
Proposition~\ref{prop:mse} with its inversion condition (\ref{app:proof-mse}), and the fusion theorem
with its optimal convex weight (\ref{app:proof-two-channel-fusion}).
\item \textbf{Appendix~\ref{app:theory-details}} tests the analysis
with two controlled examples of reconstruction inversion and
value--sensitivity non-redundancy (\ref{app:toy}), an estimate of the
MSE decomposition on trained denoisers (\ref{app:prop2-validation}),
a tap-versus-output sensitivity comparison (\ref{app:tap-output-check}),
and a check of the fusion model of Section~\ref{subsec:theory-fusion}
against measured fusion performance (\ref{app:fusion-diagnostic}).
\item \textbf{Appendix~\ref{app:molrg-diagnostics}} develops the
solvable-model analysis: the two-sheet model, the proof of
Theorem~\ref{thm:branch-dichotomy}, and a per-sample separation remark
(\ref{app:branch-dichotomy}); the posterior-covariance decomposition
for Gaussian mixtures, with a synthetic sweep (\ref{app:molrg-mixtures});
the high-noise limit (\ref{app:molrg-tap-highnoise}); and the
feature-versus-readout construction (\ref{app:tap-readout-two-sheet}).
\item \textbf{Appendix~\ref{app:implementation}} gives implementation
details: the datasets, baselines, denoisers with their conditioning
conventions, and calibration protocol
(\ref{subsec:baselines-datasets}); the label-free selection of the tap
and noise levels, with its empirical validation
(\ref{app:feature-selection}); and the ID-only evaluation of the
fusion weight (\ref{app:id-gamma}).
\item \textbf{Appendix~\ref{app:additional-experiments}} reports
further experiments: near-OOD, high-resolution, ImageNet-64, and
additional OOD benchmarks (\ref{app:near-ood}--\ref{app:additional-ood});
an OOD-assisted selection diagnostic (\ref{app:oodval}); the
reinforcement-learning planning study (\ref{app:rl-planning});
ablations of the estimator hyperparameters, the decoder tap, and the
fusion weight together with the noise-ensemble size
(\ref{app:estimator-ablation}--\ref{app:ablation-alpha-kens}); and
seed variability (\ref{app:variance}).
\end{itemize}

\section{Proofs and mathematical details}
\label{app:proofs}

We use the following notation throughout the proofs. The clean variable is $X_1$ and the noisy variable is
\[
    X_t=\alpha_tX_1+\beta_tX_0,
    \qquad
    X_0\sim\mathcal{N}(0,I_d),
    \qquad
    X_0\perp X_1 .
\]
A realized noisy state is denoted by $\vx_t$. When $X_1\sim p$, the noisy variable
$X_t$ has density $p_t$; when $X_1\sim q$, it has density $q_t$. Throughout, $\alpha_t,\beta_t>0$. Expectations and
covariances with subscript $p$ or $q$ are taken under the corresponding joint
distribution induced by the forward process. Thus,
\[
    D_p(\vx_t,t)
    =
    \mathbb{E}_p[X_1\mid X_t=\vx_t],
    \qquad
    D_q(\vx_t,t)
    =
    \mathbb{E}_q[X_1\mid X_t=\vx_t].
\]
We write $\|\cdot\|_2$ for the Euclidean norm of a vector and the spectral norm of a
matrix, depending on context. For a matrix $A$, $s_j(A)$ denotes its $j$-th largest
singular value. For a symmetric matrix $M$, $\lambda_j(M)$ denotes its $j$-th largest
eigenvalue.

\subsection{Derivation of Tweedie's and Miyasawa's identities}
\label{app:tweedie-miyasawa}

We first derive the two standard identities used in the main text. Let $X_1\sim p$,
let $X_0\sim\mathcal{N}(0,I_d)$, and define
\[
    X_t=\alpha_tX_1+\beta_tX_0.
\]
For a fixed $t$, the noisy density is
\[
    p_t(\vx_t)
    =
    \int p(\vx)\,
    \varphi_{\beta_t^2 I_d}(\vx_t-\alpha_t\vx)\,d\vx,
\]
where $\varphi_{\beta_t^2 I_d}$ is the Gaussian density with covariance
$\beta_t^2I_d$. Differentiating with respect to $\vx_t$ gives
\begin{align}
    \nabla_{\vx_t}p_t(\vx_t)
    &=
    \int p(\vx)\,
    \nabla_{\vx_t}\varphi_{\beta_t^2 I_d}(\vx_t-\alpha_t\vx)\,d\vx
    \nonumber\\
    &=
    \int p(\vx)\,
    \left(
        -\frac{\vx_t-\alpha_t\vx}{\beta_t^2}
    \right)
    \varphi_{\beta_t^2 I_d}(\vx_t-\alpha_t\vx)\,d\vx .
\end{align}
Dividing by $p_t(\vx_t)$ yields
\begin{align}
    \nabla_{\vx_t}\log p_t(\vx_t)
    &=
    -\frac{1}{\beta_t^2}
    \left(
        \vx_t-\alpha_t
        \mathbb{E}_p[X_1\mid X_t=\vx_t]
    \right).
\end{align}
Rearranging gives Tweedie's formula:
\begin{equation}
    D_p(\vx_t,t)
    =
    \mathbb{E}_p[X_1\mid X_t=\vx_t]
    =
    \frac{1}{\alpha_t}
    \left(
        \vx_t+\beta_t^2\nabla_{\vx_t}\log p_t(\vx_t)
    \right).
    \label{eq:app-tweedie}
\end{equation}

Next, we derive the posterior-covariance identity. The posterior density of $X_1$
given $X_t=\vx_t$ is proportional to
\[
    p(\vx)\,
    \exp\left(
        -\frac{\|\vx_t-\alpha_t\vx\|_2^2}{2\beta_t^2}
    \right).
\]
Differentiating the posterior mean under the integral gives
\[
    \frac{\partial}{\partial \vx_t}
    \mathbb{E}_p[X_1\mid X_t=\vx_t]
    =
    \frac{\alpha_t}{\beta_t^2}
    \operatorname{Cov}_p[X_1\mid X_t=\vx_t].
\]
Therefore,
\begin{equation}
    \operatorname{Cov}_p[X_1\mid X_t=\vx_t]
    =
    \frac{\beta_t^2}{\alpha_t}
    \frac{\partial D_p(\vx_t,t)}{\partial \vx_t}.
    \label{eq:app-miyasawa-1}
\end{equation}
Differentiating~\eqref{eq:app-tweedie} gives
\[
    \frac{\partial D_p(\vx_t,t)}{\partial \vx_t}
    =
    \frac{1}{\alpha_t}
    \left(
        I_d+\beta_t^2\nabla_{\vx_t}^2\log p_t(\vx_t)
    \right).
\]
Substituting into~\eqref{eq:app-miyasawa-1} yields
\begin{equation}
    \operatorname{Cov}_p[X_1\mid X_t=\vx_t]
    =
    \frac{\beta_t^2}{\alpha_t^2}
    \left(
        I_d+\beta_t^2\nabla_{\vx_t}^2\log p_t(\vx_t)
    \right).
    \label{eq:app-miyasawa-2}
\end{equation}
The same derivation applies to $q$ by replacing $p$ with $q$.

\subsection{Proof of Proposition~\ref{prop:mse}}
\label{app:proof-mse}

\begin{proof}
Fix a noise level $t$. For ID data, let $X_1\sim p$ and
$X_t=\alpha_tX_1+\beta_tX_0$. Since
\[
    D_p(X_t,t)=\mathbb{E}_p[X_1\mid X_t],
\]
the expected squared error given $X_t$ is the trace of the posterior covariance. Therefore,
\begin{align}
    \mathbb{E}_{X_1\sim p,X_0}
    \left[
        \|X_1-D_p(X_t,t)\|_2^2
    \right]
    &=
    \mathbb{E}_{\vx_t\sim p_t}
    \left[
        \operatorname{tr}
        \operatorname{Cov}_p[X_1\mid X_t=\vx_t]
    \right].
    \label{eq:app-id-risk}
\end{align}

For OOD data, let $X_1\sim q$ and again let
$X_t=\alpha_tX_1+\beta_tX_0$. The score is still computed with the ID denoiser
$D_p$, but we introduce the OOD Bayes denoiser
\[
    D_q(X_t,t)=\mathbb{E}_q[X_1\mid X_t]
\]
as an analytical reference. Decompose the residual as
\[
    X_1-D_p(X_t,t)
    =
    \bigl(X_1-D_q(X_t,t)\bigr)
    +
    \bigl(D_q(X_t,t)-D_p(X_t,t)\bigr).
\]
Conditioning on $X_t=\vx_t$ gives
\begin{align}
    &\mathbb{E}_q
    \left[
        \|X_1-D_p(X_t,t)\|_2^2
        \mid X_t=\vx_t
    \right]
    \nonumber\\
    &\quad =
    \mathbb{E}_q
    \left[
        \|X_1-D_q(\vx_t,t)\|_2^2
        \mid X_t=\vx_t
    \right]
    +
    \|D_q(\vx_t,t)-D_p(\vx_t,t)\|_2^2
    \nonumber\\
    &\qquad+
    2
    \left\langle
        \mathbb{E}_q[X_1-D_q(\vx_t,t)\mid X_t=\vx_t],
        D_q(\vx_t,t)-D_p(\vx_t,t)
    \right\rangle .
\end{align}
The cross term is zero because
\[
    \mathbb{E}_q[X_1-D_q(\vx_t,t)\mid X_t=\vx_t]=0.
\]
Hence,
\begin{align}
    \mathbb{E}_{X_1\sim q,X_0}
    \left[
        \|X_1-D_p(X_t,t)\|_2^2
    \right]
    &=
    \mathbb{E}_{\vx_t\sim q_t}
    \left[
        \operatorname{tr}
        \operatorname{Cov}_q[X_1\mid X_t=\vx_t]
    \right]
    \nonumber\\
    &\quad+
    \mathbb{E}_{\vx_t\sim q_t}
    \left[
        \|D_q(\vx_t,t)-D_p(\vx_t,t)\|_2^2
    \right].
    \label{eq:app-ood-risk}
\end{align}

Using Tweedie's formula for both $p$ and $q$,
\[
    D_p(\vx_t,t)
    =
    \frac{1}{\alpha_t}
    \left(
        \vx_t+\beta_t^2\nabla\log p_t(\vx_t)
    \right),
\]
and
\[
    D_q(\vx_t,t)
    =
    \frac{1}{\alpha_t}
    \left(
        \vx_t+\beta_t^2\nabla\log q_t(\vx_t)
    \right).
\]
Therefore,
\[
    D_q(\vx_t,t)-D_p(\vx_t,t)
    =
    \frac{\beta_t^2}{\alpha_t}
    \left(
        \nabla\log q_t(\vx_t)-\nabla\log p_t(\vx_t)
    \right),
\]
and
\[
    \|D_q(\vx_t,t)-D_p(\vx_t,t)\|_2^2
    =
    \frac{\beta_t^4}{\alpha_t^2}
    \left\|
        \nabla\log q_t(\vx_t)-\nabla\log p_t(\vx_t)
    \right\|_2^2 .
\]
Substituting this identity into~\eqref{eq:app-ood-risk} and subtracting the ID risk
in~\eqref{eq:app-id-risk} gives
\begin{align}
    \Delta_{\mathrm{MSE}}(t)
    &=
    \underbrace{
    \mathbb{E}_{\vx_t\sim q_t}
    \left[
        \operatorname{tr}
        \operatorname{Cov}_q[X_1\mid X_t=\vx_t]
    \right]
    -
    \mathbb{E}_{\vx_t\sim p_t}
    \left[
        \operatorname{tr}
        \operatorname{Cov}_p[X_1\mid X_t=\vx_t]
    \right]
    }_{B(t)}
    \nonumber\\
    &\quad+
    \frac{\beta_t^4}{\alpha_t^2}
    \mathbb{E}_{\vx_t\sim q_t}
    \left[
        \left\|
            \nabla\log p_t(\vx_t)-\nabla\log q_t(\vx_t)
        \right\|_2^2
    \right].
\end{align}
This proves~\eqref{eq:mse-decomp}.
\end{proof}

\begin{remark}[Interpreting the MSE gap]
\label{rem:mse-inv}
At an interior noise level, the score-mismatch energy is positive
whenever $p\neq q$. However, OOD data may also be easier to denoise
than ID data, $B(t)<0$. This can result in the two distributions having the same expected reconstruction MSE, or even OOD data having a lower expected MSE: by~\eqref{eq:mse-decomp}, $\Delta_{\mathrm{MSE}}(t)<0$ exactly when $-B(t)$ exceeds the score-mismatch energy.
\end{remark}
Appendix~\ref{app:prop2-validation} examines this explanation
on the two most strongly inverted benchmark pairs using
trained denoisers.

\subsection{Optimal two-channel fusion: statement and proof}
\label{app:proof-two-channel-fusion}

This subsection states the general linear-score analysis behind
Eq.~\eqref{eq:convex-deflection}. Let $Y=(S_R,S_J)^\top$ have finite second
moments, with $\Delta:=\mathbb{E}_q[Y]-\mathbb{E}_p[Y]$ and
$C:=\operatorname{Cov}_p(Y)$. For a linear score $w^\top Y$ with
$w^\top Cw>0$, its standardized mean gap is
\begin{equation}
    d(w^\top Y)
    =
    \frac{w^\top\Delta}{\sqrt{w^\top Cw}} ,
    \label{eq:deflection-definition}
\end{equation}
where $d(S):=\bigl(\mathbb{E}_q[S]-\mathbb{E}_p[S]\bigr)/\sqrt{\operatorname{Var}_p(S)}$ denotes the standardized mean gap of a scalar score $S$. Equation~\eqref{eq:convex-deflection} of the main text is the specialization
to the ID-standardized channels and $w=(\gamma,1-\gamma)^\top$.

\begin{theorem}[Optimal two-channel linear fusion]
\label{thm:two-channel-fusion}
Let $Z=(Z_R,Z_J)^\top$ have ID mean $0$, ID covariance
$C=\begin{psmallmatrix}1&\rho\\ \rho&1\end{psmallmatrix}$ with $|\rho|<1$,
and OOD mean $\Delta=(\Delta_R,\Delta_J)^\top$. Among linear scores
$S_w=w^\top Z$, the best squared standardized mean gap is attained at
$w\propto C^{-1}\Delta$ and equals
\begin{equation}
\max_{w\neq0} d(S_w)^2
=\Delta^\top C^{-1}\Delta
=\Delta_J^2+\frac{(\Delta_R-\rho\Delta_J)^2}{1-\rho^2}.
\label{eq:optimal-linear-deflection}
\end{equation}
The same holds with $R$ and $J$ swapped. Hence the best fusion is never
worse than either channel alone, and it is strictly better than $Z_J$
alone unless $\Delta_R=\rho\Delta_J$ (and than $Z_R$ alone unless
$\Delta_J=\rho\Delta_R$).
\end{theorem}

\begin{proof}
For $S_w=w^\top Z$ we have $\mathbb{E}_q[S_w]-\mathbb{E}_p[S_w]=w^\top\Delta$
and $\operatorname{Var}_p(S_w)=w^\top Cw$, so
$d(S_w)^2=(w^\top\Delta)^2/(w^\top Cw)$.
Since $|\rho|<1$, $C$ is positive definite, and
$\langle u,v\rangle_C:=u^\top Cv$ is an inner product. Writing
$w^\top\Delta=\langle w,C^{-1}\Delta\rangle_C$, the Cauchy--Schwarz
inequality gives
\[
d(S_w)^2\le\frac{\|w\|_C^2\,\|C^{-1}\Delta\|_C^2}{\|w\|_C^2}
=\Delta^\top C^{-1}\Delta,
\]
with equality if and only if $w\propto C^{-1}\Delta$.
Using $C^{-1}=\frac{1}{1-\rho^2}\begin{psmallmatrix}1&-\rho\\-\rho&1\end{psmallmatrix}$
and completing the square,
\[
\Delta^\top C^{-1}\Delta
=\frac{\Delta_R^2-2\rho\Delta_R\Delta_J+\Delta_J^2}{1-\rho^2}
=\Delta_J^2+\frac{(\Delta_R-\rho\Delta_J)^2}{1-\rho^2}.
\]
The sensitivity channel alone has $d(Z_J)^2=\Delta_J^2$, so the second term
is the gain from fusion; it vanishes only when $\Delta_R=\rho\Delta_J$.
Exchanging $R$ and $J$ gives the other case.
\end{proof}

\begin{remark}[Optimal convex weight]
\label{rem:convex-weight}
The deployed score uses $w=(\gamma,1-\gamma)^\top$ with $\gamma\in[0,1]$.
Since $d$ is invariant to positive rescaling of $w$, $S_\gamma$ attains the
optimum of Theorem~\ref{thm:two-channel-fusion} if and only if
$(\gamma,1-\gamma)^\top$ is a positive multiple of
\[
C^{-1}\Delta=\frac{1}{1-\rho^2}
\begin{pmatrix}\Delta_R-\rho\Delta_J\\ \Delta_J-\rho\Delta_R\end{pmatrix}.
\]
When both coordinates are positive, normalizing them to sum to one gives
\[
\gamma^\star=\frac{\Delta_R-\rho\Delta_J}{(1-\rho)(\Delta_R+\Delta_J)}.
\]
Otherwise, the only interior stationary point of $\gamma\mapsto d(S_\gamma)$,
if any, is a minimum (where $w\propto-C^{-1}\Delta$), so the maximum lies at an
endpoint: $\gamma^\star=1$ (reconstruction only) if $\Delta_R\ge\Delta_J$,
and $\gamma^\star=0$ (sensitivity only) otherwise.
\end{remark}

\begin{remark}[Attainable separation]
\label{rem:dstar-bound}
Let $d^\star:=(\Delta^\top C^{-1}\Delta)^{1/2}$. Every convex weight gives a
nonzero $w$, so $d(S_\gamma)\le d^\star$ for all $\gamma\in[0,1]$.
In particular, if $d^\star<1$, no choice of $\gamma$ separates the ID and OOD
means by even one ID standard deviation.
\end{remark}

\section{Additional experiments for the posterior-geometry analysis}
\label{app:theory-details}
This appendix examines the posterior-geometry analysis with controlled and numerical experiments. We begin with examples of reconstruction inversion and the
distinct information carried by feature values and sensitivity (\ref{app:toy}). We then estimate the MSE decomposition
of Proposition~\ref{prop:mse} using trained denoisers (\ref{app:prop2-validation}), and compare OOD sensitivity at intermediate decoder taps with that at the denoiser output, together with diagnostics of the mechanism in Proposition~\ref{prop:tap-readout-two-sheet} (\ref{app:tap-output-check}). Finally, Appendix~\ref{app:fusion-diagnostic} compares the fusion
predictions of Section~\ref{subsec:theory-fusion} with the
observed fusion behavior.
\subsection{Controlled examples}
\label{app:toy}

\begin{figure}[t]
    \centering
    \includegraphics[width=.7\linewidth]{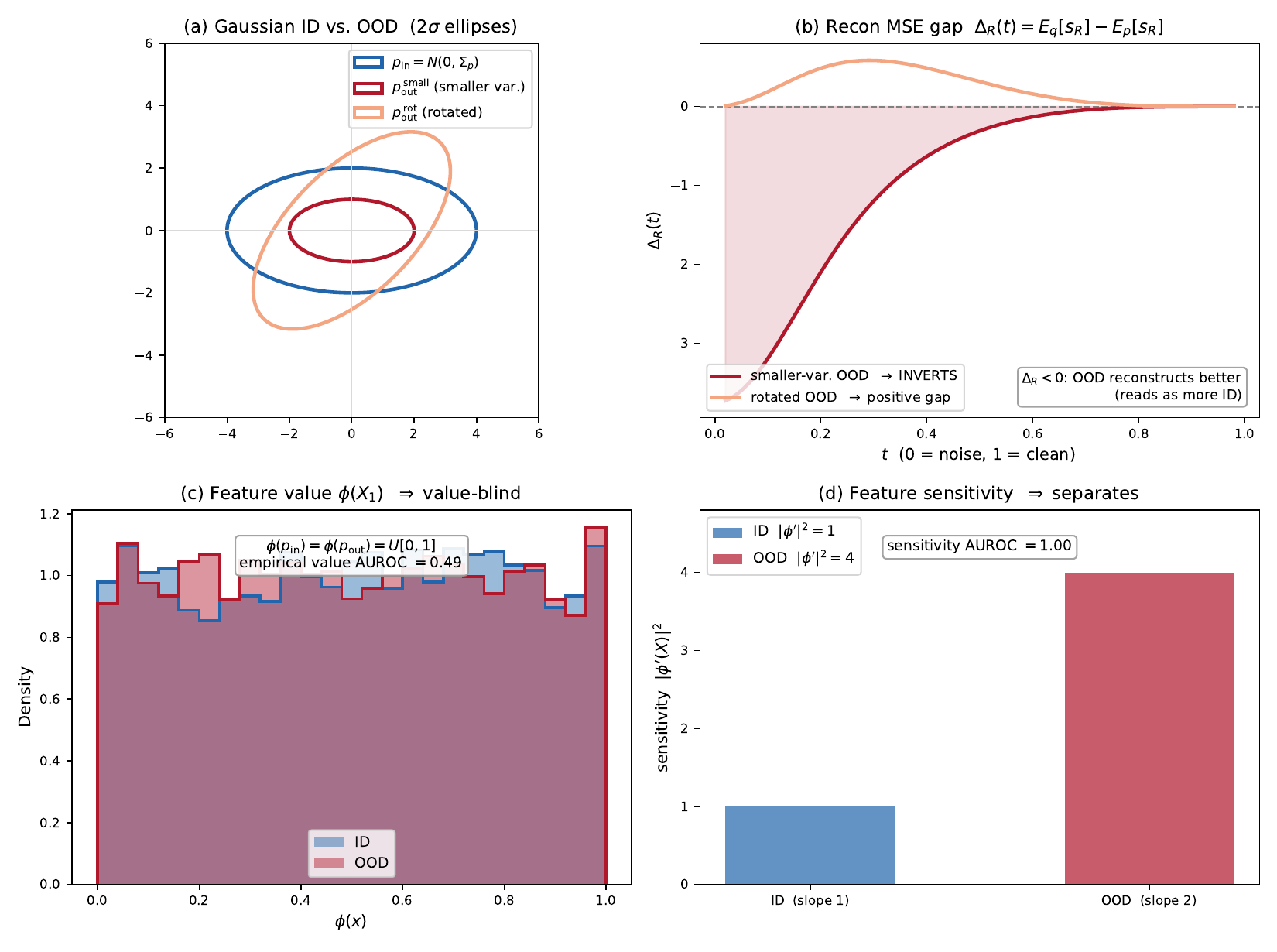}
    \caption{
    \textbf{Controlled toy examples.}
    (a) Gaussian ID and OOD distributions in Toy A, shown by
    $2\sigma$ ellipses.
    (b) OOD-minus-ID expected reconstruction gap under the
    Bayes ID denoiser. The smaller-variance OOD law yields a
    negative gap, demonstrating reconstruction inversion;
    the rotated covariance yields a positive gap.
    (c) In Toy B, feature values follow $U[0,1]$ under both
    distributions, so any detector using only these values
    performs at chance in population.
    (d) The squared feature derivative $|\phi'(X_1)|^2$ equals
    $1$ for ID and $4$ for OOD, giving perfect separation
    in this construction.
    }
    \label{fig:controlled-toy}
\end{figure}

We give two controlled examples. The first shows that a
Bayes-optimal denoiser can assign a lower expected reconstruction
error to the OOD samples. The second shows that local feature
sensitivity can distinguish distributions whose feature values
have identical laws. Figure~\ref{fig:controlled-toy} illustrates
both constructions.

\paragraph{Toy A: reconstruction can invert.}
Let
\[
    p=\mathcal{N}(0,\Sigma_p),
    \qquad
    q=\mathcal{N}(0,\Sigma_q),
\]
and consider the forward process
\[
    X_t=\alpha_tX_1+\beta_tX_0,
    \qquad
    X_0\sim\mathcal{N}(0,I).
\]
For the ID distribution $p$, the Bayes denoiser is linear:
\begin{equation}
    D_p(\vx_t,t)
    =
    J_p(t)\vx_t,
    \qquad
    J_p(t)
    =
    \alpha_t\Sigma_p
    \left(
        \alpha_t^2\Sigma_p+\beta_t^2I
    \right)^{-1}.
    \label{eq:toy-gaussian-denoiser}
\end{equation}
Therefore the fixed-time reconstruction score satisfies
\begin{equation}
    s_{\mathrm{MSE}}(\vx;t)
    =
    \mathbb{E}_{\epsilon}
    \left[
        \left\|
        \vx-D_p(\alpha_t\vx+\beta_t\epsilon,t)
        \right\|_2^2
    \right]
    =
    \|A_t\vx\|_2^2
    +
    \beta_t^2\|J_p(t)\|_F^2,
    \label{eq:toy-recon-score}
\end{equation}
where
\[
    A_t=\alpha_tJ_p(t)-I.
\]
The constant second term cancels when comparing ID and OOD expectations, giving
\begin{equation}
    \Delta_{\mathrm{MSE}}(t)
    =
    \mathbb{E}_{q}[s_{\mathrm{MSE}}(X_1;t)]
    -
    \mathbb{E}_{p}[s_{\mathrm{MSE}}(X_1;t)]
    =
    \operatorname{tr}
    \left[
        A_t^\top A_t(\Sigma_q-\Sigma_p)
    \right].
    \label{eq:toy-recon-gap}
\end{equation}
The expected reconstruction gap is determined by the covariance
difference weighted by $A_t^\top A_t$. A negative gap means that
OOD samples receive lower reconstruction error on average.
Since $A_t^\top A_t\succ0$, the gap is negative whenever
$\Sigma_q\preceq\Sigma_p$ and $\Sigma_q\neq\Sigma_p$, and nonnegative for a
rotated $\Sigma_q=R\Sigma_pR^\top$ (Figure~\ref{fig:controlled-toy}(b)). Thus, an OOD distribution can appear easier to reconstruct than the ID distribution, resulting in unreliable OOD detection.

\paragraph{Toy B: value and sensitivity are non-redundant.}
The Gaussian example above has an input-independent denoiser Jacobian, so it cannot
illustrate sample-dependent sensitivity. We therefore use a separate feature-map
construction to isolate value--sensitivity non-redundancy. Let
\[
    p=U[0,1],
    \qquad
    q=U[2,2.5],
\]
and define
\begin{equation}
    \phi(x)
    =
    \begin{cases}
        x, & x\in[0,1],\\
        2(x-2), & x\in[2,2.5],
    \end{cases}
    \label{eq:toy-feature-map}
\end{equation}
with a smooth bridge on $(1,2)$, where neither distribution places mass. Then
under both $X_1\sim p$ and $X_1\sim q$, the feature value $\phi(X_1)$ is uniformly distributed on $[0,1]$,
so every detector that depends only on the feature value $\phi(X_1)$ is blind in population. In contrast,
\[
    |\phi'(X_1)|^2=1
    \quad p\text{-a.s.},
    \qquad
    |\phi'(X_1)|^2=4
    \quad q\text{-a.s.},
\]
so the sensitivity statistic separates perfectly (Figure~\ref{fig:controlled-toy}(c,d)). This toy example isolates the basic mathematical fact that the value of a representation and its local sensitivity are different observables.

\subsection{Empirical MSE decomposition}
\label{app:prop2-validation}

We examine CIFAR-10 and CIFAR-100 as ID and SVHN as OOD, the two
most strongly inverted MSE benchmark pairs, using the frozen models.
We estimate the components of Proposition~\ref{prop:mse}
using trained denoisers as approximations to Bayes denoisers.
The score-mismatch energy is measured by the squared difference between their
outputs on the same noisy OOD inputs, which equals its score form by Tweedie's identity; the OOD denoiser is used only for this diagnostic.

\begin{figure}[t]
    \centering
    \includegraphics[width=.7\linewidth]
    {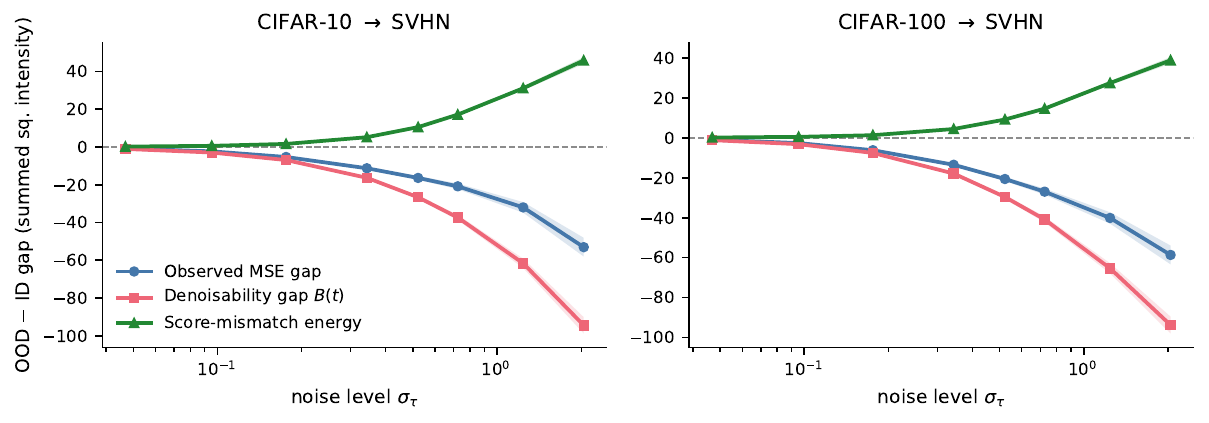}
    \caption{
    Empirical components of the reconstruction gap.
On both pairs, the negative denoisability gap $B(t)$
outweighs the positive score-mismatch energy, leaving OOD data
with lower observed reconstruction MSE.
Shading shows $\pm1$ standard error.
    }
    \label{fig:prop1-decomposition}
\end{figure}

Figure~\ref{fig:prop1-decomposition} shows that SVHN has
lower reconstruction error under its own denoiser than
ID data under theirs, so the denoisability gap $B(t)$ is negative. The score-mismatch energy offsets only
$18\%$--$51\%$ of it, leaving the observed
reconstruction gap negative.
The sum of the estimated components differs from the
observed gap by at most $1.3\%$ for $\sigma_\tau\le0.34$
($\sigma_\tau=\beta_\tau/\alpha_\tau$)
and $8.3\%$ overall. These results support denoisability as an explanation for MSE
inversion: on both pairs, the negative denoisability gap
outweighs the positive score-mismatch energy. Reconstruction MSE
can therefore assign lower scores to easier-to-denoise OOD data,
even when evaluated with the ID denoiser.

\subsection{Sensitivity across decoder layers}
\label{app:tap-output-check}

\paragraph{Measuring OOD sensitivity shift across layers.} We compare sensitivity at intermediate decoder taps and the
denoiser output using $32$ probes with $\delta=0.1$.
For each tap and for the output, the vertical axis reports 
$\frac{\mathbb E_{\mathrm{OOD}}[s]-\mu_{\mathrm{ID}}}
     {\sigma_{\mathrm{ID}}}$,
where $s$ is the sensitivity score $e_{\ell,\tau}$ and $\mu_{\mathrm{ID}}$,
$\sigma_{\mathrm{ID}}$ are its ID calibration mean and standard deviation.
Thus, a value of $1$ means that the OOD mean is one ID standard
deviation above the ID mean. Positive values indicate the
desired mean ordering, zero indicates equal means, and negative
values indicate reversed mean ordering.

\begin{figure}[t]
    \centering
    \includegraphics[width=.7\linewidth]
    {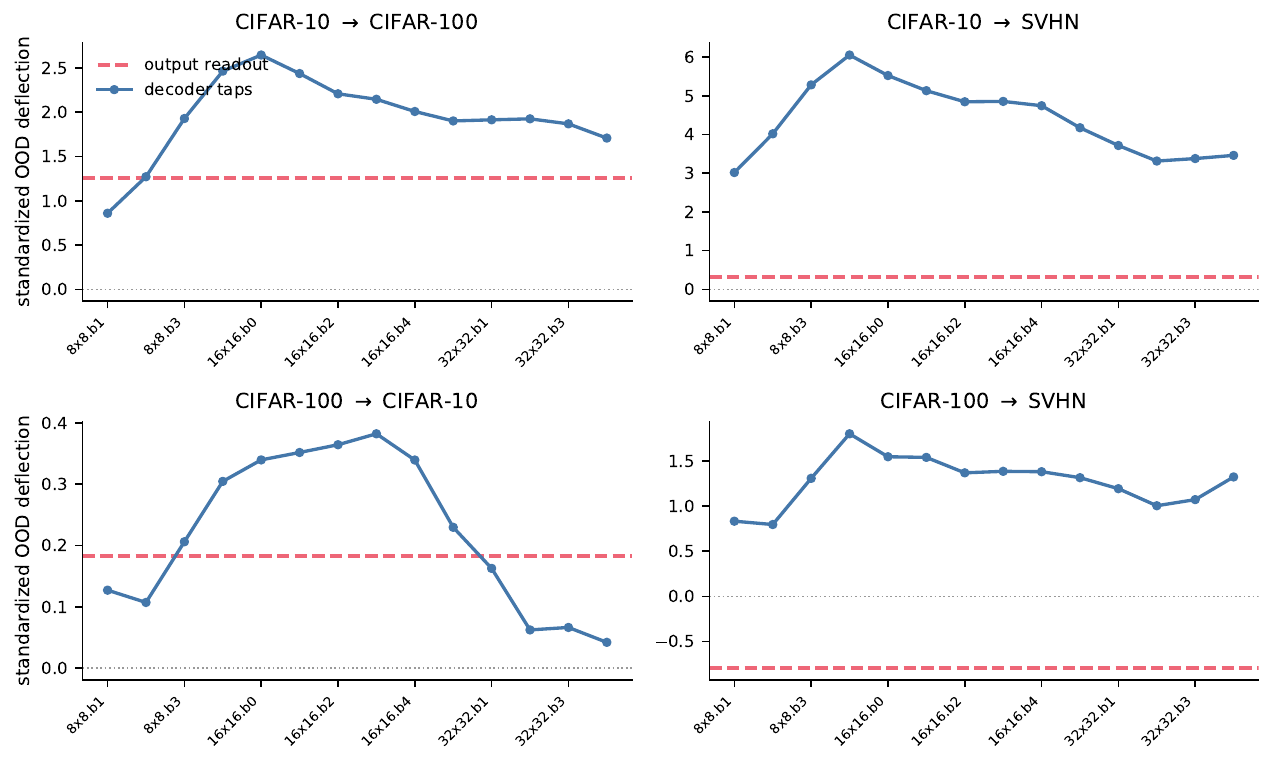}
    \caption{
    OOD sensitivity shifts across intermediate decoder taps
    (solid) and at the denoiser output (dashed) at $\tau=100$,
    measured in ID calibration standard deviations.
    }
    \label{fig:tap-output-sweep}
\end{figure}

Figure~\ref{fig:tap-output-sweep} shows four representative
pairs at $\tau=100$. In all four, the strongest intermediate
taps show larger OOD shifts than the output. For the SVHN
pairs, tap shifts remain positive while the output shift
is near zero or negative. These results support using
intermediate taps on the studied pairs and complement the
layer sweep in Appendix~\ref{app:layer-ablation}.

\paragraph{Feature sensitivity in the trained U-Net.}
We study the frozen CIFAR-10 and CIFAR-100 denoisers, using
each model's training distribution as ID and SVHN and the other
CIFAR dataset as OOD. The alignment and tap--output comparisons
use $32$ input probes with $\delta=0.1$ at the tap
\texttt{dec.16x16\_block1}.

The alignment diagnostic uses the CIFAR-10 denoiser at
$\tau=100$. We fit $10$ K-means clusters to spatially pooled
tap activations from ID calibration images and obtain soft
assignment probabilities by a softmax over squared distances to
these cluster centers.
For each test image, we measure whether perturbations along
the dominant sensitivity direction change its nearest-cluster
assignment probability. Specifically, alignment is the absolute
cosine similarity across probes between the response along that
direction and the change in assignment probability, with the
nearest center and the softmax temperature held fixed.

We also measure how concentrated the sensitivity responses are
using the effective dimension $(\operatorname{tr}G)^2/\operatorname{tr}(G^2)$ of
the tap-Jacobian Gram $G$: lower values indicate that fewer
directions dominate. This diagnostic uses the CIFAR-10 model,
$512$ probes per image at $\tau\in\{100,200\}$.

\begin{figure}[t]
\centering
\includegraphics[width=\linewidth]{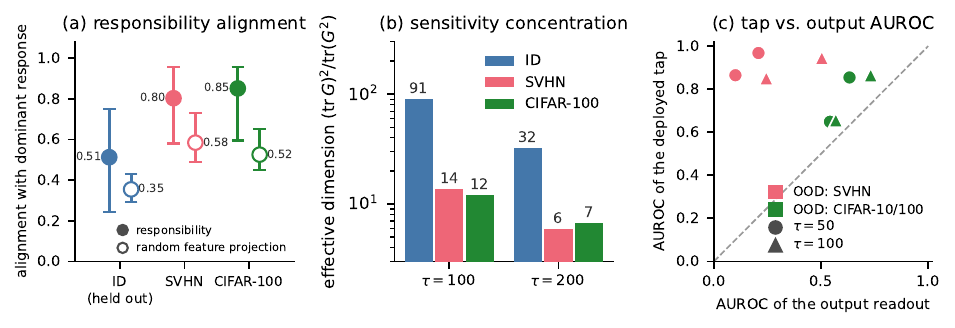}
\caption{Sensitivity diagnostics in trained U-Nets.
\textbf{(a)} Alignment with ID-cluster assignment changes (filled)
and random weighted sum of the feature values (open).
\textbf{(b)} Effective dimension of tap responses; lower values
indicate greater concentration. Panels (a,b) use the CIFAR-10 denoiser.
\textbf{(c)} Tap versus output AUROC for SVHN and near-OOD pairs;
points above the diagonal favor the tap.}
\label{fig:mechanism-diagnostics}
\end{figure}

Figure~\ref{fig:mechanism-diagnostics} shows higher median
alignment with cluster-assignment changes than with random
feature projections, together with more concentrated OOD sensitivity.
For the tap--output comparison at
$\tau\in\{50,100\}$, the tap achieves higher AUROC on the
SVHN and CIFAR pairs shown. These observations are consistent
with the feature--readout contrast in
Appendix~\ref{app:tap-readout-two-sheet}.

\subsection{Fusion-model diagnostic}
\label{app:fusion-diagnostic}

We test whether the two-channel deflection model of Section~\ref{subsec:theory-fusion} explains the fusion behavior of the detector. For each ID--OOD pair, we use the cached calibrated per-image channels $z_R$ and $z_J$ and estimate
\[
\Delta_R=\mathbb{E}_q[Z_R]-\mathbb{E}_p[Z_R],
\qquad
\Delta_J=\mathbb{E}_q[Z_J]-\mathbb{E}_p[Z_J],
\qquad
\rho=\operatorname{Corr}_p(Z_R,Z_J).
\]
We then compare the predicted curve
\[
d(S_\gamma)
=
\frac{
\gamma\Delta_R+(1-\gamma)\Delta_J
}{
\sqrt{
\gamma^2+(1-\gamma)^2+2\gamma(1-\gamma)\rho
}
},
\qquad
\operatorname{AUROC}_{\mathrm{pred}}(\gamma)
=
\Phi\!\left(\frac{d(S_\gamma)}{\sqrt{2}}\right),
\]
where $\Phi$ is the standard normal distribution function, with the
measured AUROC of
$S_\gamma=\gamma z_R+(1-\gamma)z_J$ over $\gamma\in[0,1]$.

\begin{figure}[H]
\centering
\includegraphics[width=0.75\linewidth]{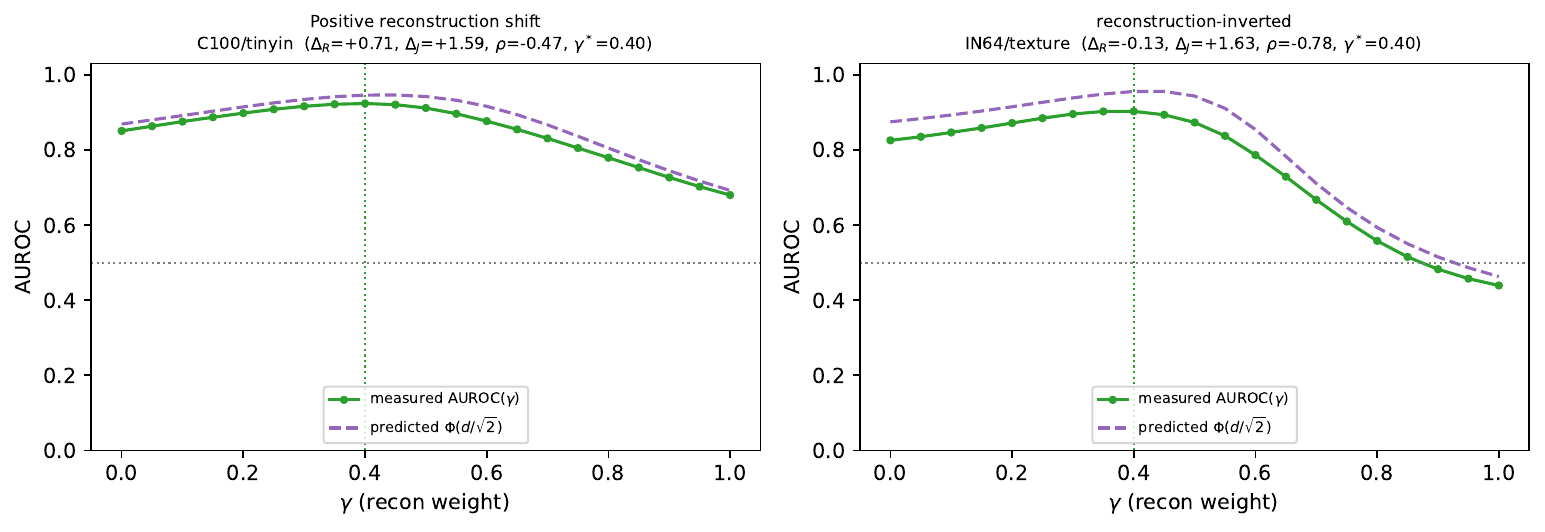}
\caption{
Predicted (dashed) and measured (solid) fusion AUROC for two
representative ID--OOD pairs. Both favor an intermediate
weight, including the case where reconstruction alone is inverted.
}
\label{fig:fusion_deflection}
\end{figure}

Figure~\ref{fig:fusion_deflection} shows two representative ID--OOD pairs.
Both examples favor an intermediate fusion weight.
In the reconstruction-inverted example, a moderate reconstruction
weight still improves AUROC, consistent with the variance
reduction from negative channel correlation. Performance falls
as the weight approaches reconstruction-only scoring.
Across the 14 evaluated pairs, the model approximates the
measured curves with a mean absolute AUROC error of about $0.037$.
This diagnostic is not used to determine the deployed weight as it requires OOD data.  We instead select the deployed fusion weight using the ID-only procedure in Appendix~\ref{app:id-gamma}.

\section{Branch ambiguity in Gaussian mixtures}
\label{app:molrg-diagnostics}

This appendix develops the solvable model behind the
analysis of Section~\ref{sec:theory}. We first prove
Theorem~\ref{thm:branch-dichotomy} by reducing the two-sheet model
to the single coordinate along which the sheets separate, where
reconstruction error and branch uncertainty can be compared in
closed form (\ref{app:branch-dichotomy}). We then show that the
same uncertainty between plausible components appears in the
posterior covariance of general Gaussian mixtures and illustrate its
effect on OOD scores with a synthetic sweep (\ref{app:molrg-mixtures}). Next, we characterize
the high-noise limit, where this signal collapses
(\ref{app:molrg-tap-highnoise}). Finally, we write the Bayes denoiser in
the same model as a readout of an intermediate feature whose sensitivity
measures the same branch uncertainty, and show that a readout with
near-optimal denoising risk can reverse the output's sensitivity ordering
while the feature's ordering is unchanged; this motivates measuring
sensitivity before the output (\ref{app:tap-readout-two-sheet}).

\subsection{Two sheets: reconstruction and branch uncertainty}
\label{app:branch-dichotomy}

\paragraph{The model.}
ID samples lie on two parallel sheets, while OOD samples lie on
their midplane. Their variation along the sheets is identical (Figure~\ref{fig:branch-intuition}).
Formally, let
\[
X_1=W+mSe_b\ \ (\text{ID}),\qquad X_1=W\ \ (\text{OOD}),\qquad
W\sim\mathcal N(0,\Sigma),\qquad
\mathbb P(S=\pm1)=\tfrac12.
\]
Here $W$ and $S$ are independent, $\Sigma$ has rank $r<d$,
and $e_b$ is a unit vector orthogonal to its range.
The distance between the ID sheets is $\omega=2m>0$.

Fix an interior noise level and observe
$X_t=\alpha_tX_1+\beta_tX_0$ as in Eq.~\eqref{eq:vp-forward}, with
$\alpha_t,\beta_t>0$ and independent $X_0\sim\mathcal N(0,I_d)$.
Only the coordinate along $e_b$ distinguishes the two laws.
Dividing this noisy coordinate by its noise standard deviation gives
\[
U:=\frac{e_b^\top X_t}{\beta_t}
=
\begin{cases}
\xi,&\text{under OOD},\\
\zeta_t S/2+\xi,&\text{under ID},
\end{cases}
\qquad
\xi:=e_b^\top X_0\sim\mathcal N(0,1),\qquad
\zeta_t:=\frac{\alpha_t\omega}{\beta_t}.
\]
Thus $\zeta_t$ is the distance between the noisy ID branches,
measured in noise standard deviations.

\paragraph{What the ID denoiser predicts.}
Given $U=u$, the posterior odds of the two ID branches are
$\exp(\zeta_tu)$. The predicted branch coordinate and its
posterior variance are therefore
\[
f_t(u):=\mathbb E_p[mS\mid U=u]
       =m\tanh(\zeta_tu/2),
\qquad
v_t(u):=\operatorname{Var}_p(mS\mid U=u)
       =m^2-f_t(u)^2.
\]
Near the midplane, the two branches are equally plausible:
the prediction is close to zero and the variance is close to
$m^2$. Farther from the midplane, one branch becomes more likely
and the variance decreases. More precisely, $v_t$ is even and
strictly decreasing in $|u|$.

The within-sheet posterior is Gaussian, with covariance
\[
C_t=\beta_t^2\Sigma(\alpha_t^2\Sigma+\beta_t^2I_d)^{-1}.
\]
This matrix has rank $r$, satisfies $C_te_b=0$, and does not
depend on the observed input. Given $X_t$, the within-sheet coordinate $W$
and the branch $S$ are conditionally independent, because they are independent
a priori and the likelihood factorizes into a term in $(I_d-e_be_b^\top)\vx_t$ that depends only on $W$
and a term in $e_b^\top\vx_t$ that depends only on $S$. Hence
\begin{equation}
\operatorname{Cov}_p[X_1\mid X_t=\vx_t]
=C_t+v_t\!\left(\frac{e_b^\top\vx_t}{\beta_t}\right)e_be_b^\top.
\label{eq:branch-covariance-short}
\end{equation}
The first term describes uncertainty along the sheets; the
second describes uncertainty about which sheet generated the
input. OOD inputs are scored with this same ID posterior.

\paragraph{The two scores.}
Write $\mathcal S_K(\cdot)$ for the sum of the $K$ largest
eigenvalues of a positive semidefinite matrix.
Below, $K=K_{\mathrm{spec}}$, and $(z)_+:=\max\{z,0\}$.
For a clean input $\vx$ and noise draw $\epsilon\sim\mathcal N(0,I_d)$ as in Eq.~\eqref{eq:mse-fixed-t}, we analyze
\[
\begin{aligned}
s_{\mathrm{MSE}}(\vx;t)
&=\mathbb E_{\epsilon}\|\vx-D_p(\alpha_t\vx+\beta_t\epsilon,t)\|_2^2,\\
s_{\mathrm{spec}}(\vx;t)
&=\mathbb E_{\epsilon}\mathcal S_{K_{\mathrm{spec}}}
  \bigl(\operatorname{Cov}_p[X_1\mid
         X_t=\alpha_t\vx+\beta_t\epsilon]\bigr).
\end{aligned}
\]
These expectations average exactly over the forward noise, and the
gaps $\Delta_{\mathrm{MSE}}(t)$ and $\Delta_{\mathrm{spec}}(t)$ are the
OOD-minus-ID differences of the expected scores.

The proof below reduces both gaps to the direction separating
the sheets. With strong noise, the denoiser averages the two
ID branches and predicts near the OOD midplane, giving OOD
lower reconstruction error. With sufficiently weak noise,
the ID branch becomes identifiable and the reconstruction
ordering is restored. The covariance score instead captures the
uncertainty behind the averaging. Branch uncertainty is greatest near the midplane,
where noisy OOD inputs fall more often than noisy ID inputs.
This gives OOD a higher expected covariance trace at every
interior noise level. The same conclusion holds for the
top-$K$ score when the branch contribution is large enough
to enter the retained spectrum.

\begin{proof}[Proof of Theorem~\ref{thm:branch-dichotomy}]
The argument depends on just two averages: the branch uncertainty
seen at noisy OOD and ID inputs,
\[
\bar v_q(t):=\mathbb E\,v_t(\xi),
\qquad
\bar v_p(t):=\mathbb E\,v_t(\xi+\zeta_t/2).
\]
The second expression also covers the negative ID branch because
$v_t$ and the Gaussian noise are symmetric.

\emph{Reconstruction.}
An OOD sample has true branch coordinate zero, so its squared
branch error is $f_t(U)^2=m^2-v_t(U)$.
Its expected squared branch error is therefore $m^2-\bar v_q(t)$.
For ID samples, the posterior mean is the Bayes predictor, so its
expected squared branch error equals the expected posterior variance,
$\bar v_p(t)$.
The within-sheet reconstruction errors have the same expectation
under both laws and cancel in their difference.

\emph{Covariance spectrum.}
The branch contributes one direction orthogonal to the range of
$C_t$. For any $1\le K\le d$, its eigenvalue either enters the
top $K$ or falls below the smallest retained within-sheet eigenvalue:
\begin{equation}
\mathcal S_K(C_t+ve_be_b^\top)
=\mathcal S_K(C_t)+(v-\lambda_K(C_t))_+,
\qquad v\ge0.
\label{eq:branch-topk-cutoff}
\end{equation}
Here the eigenvalues of $C_t$ are ordered decreasingly, including
zeros. The constant $\mathcal S_K(C_t)$ cancels between ID and OOD.
Together with the reconstruction calculation, this gives
\begin{equation}
\begin{aligned}
\Delta_{\mathrm{MSE}}(t)&=m^2-\bar v_q(t)-\bar v_p(t),\\
\Delta_{\mathrm{spec}}(t)
&=\mathbb E\,\bigl(v_t(\xi)-\lambda_K(C_t)\bigr)_+
 -\mathbb E\,\bigl(v_t(\xi+\zeta_t/2)-\lambda_K(C_t)\bigr)_+.
\end{aligned}
\label{eq:branch-gap-short}
\end{equation}

\emph{Why OOD has greater branch uncertainty.}
A centered Gaussian is more likely to fall near zero than a
shifted Gaussian with the same variance:
\[
\mathbb P(|\xi|<b)>\mathbb P(|\xi+a|<b),
\qquad a,b>0.
\]
To check this, write the shifted probability as
$\Phi(b-a)+\Phi(b+a)-1$.
Its derivative in $a$ is
$\varphi(b+a)-\varphi(b-a)<0$, where $\Phi$ and $\varphi$
are the standard normal distribution function and density.
Every set $\{u:v_t(u)>c\}$, for $0<c<m^2$, is an interval
centered at zero. Hence $v_t(U)$ exceeds each such level more often
under OOD than under ID. Since
$\mathbb E\,v_t(U)=\int_0^{m^2}\mathbb P\bigl(v_t(U)>c\bigr)\,dc$ under
either law, integrating over $c$ gives
$\bar v_q(t)>\bar v_p(t)$.
The same argument applies to
$(v_t-\lambda_K(C_t))_+$ whenever $\lambda_K(C_t)<m^2$:
its nonempty positive level sets are also centered intervals.
Thus the top-$K$ gap is strictly positive under this condition.
If $\lambda_K(C_t)\ge m^2$, the extra term is identically zero
and so is the gap. For the covariance trace, take $K=d$;
then $\lambda_d(C_t)=0$ and the gap is
$\bar v_q(t)-\bar v_p(t)>0$ at every interior noise level.

\emph{Why reconstruction changes sign.}
When $\zeta_t\to0$, the branches cannot be distinguished and
both average variances tend to $m^2$.
When $\zeta_t\to\infty$, both tend to zero.
These limits follow from the formula for $v_t$: its
$\tanh$ argument tends to zero in the first case and has
magnitude tending to infinity almost surely in the second.
Since $0\le v_t\le m^2$, dominated convergence also gives
continuity in $\zeta_t$.
Equation~\eqref{eq:branch-gap-short} now yields
\[
\Delta_{\mathrm{MSE}}(t)\longrightarrow
\begin{cases}
-m^2,&\zeta_t\to0,\\
\phantom{-}m^2,&\zeta_t\to\infty.
\end{cases}
\]
Thus reconstruction is inverted for sufficiently small
$\zeta_t$ and correctly ordered for sufficiently large
$\zeta_t$, giving the stated thresholds
$0<\zeta_-<\zeta_+<\infty$.
\end{proof}
The proof shows that branch uncertainty never favors ID in the spectral score, and strictly favors OOD whenever the branch variance enters the retained spectrum, whereas the reconstruction score depends on whether the two ID branches are sufficiently resolved at the chosen noise level. With exact noise averaging, the spectral ordering also holds sample by sample under the same condition $\lambda_{K_{\mathrm{spec}}}(C_t)<\omega^2/4$.

\begin{remark}[Separation after exact noise averaging]
\label{rem:branch-persample}
Fix an interior noise level and $1\le K_{\mathrm{spec}}\le d$.
If $\lambda_{K_{\mathrm{spec}}}(C_t)<\omega^2/4$, every OOD sample has a
higher noise-averaged spectral score than every ID sample.
The population AUROC is therefore $1$ at that noise level.
The within-sheet contribution is constant, and the branch
coordinate of a clean sample is either $0$ or $\pm m$.
Consequently,
\[
s_{\mathrm{spec}}(\vx;t)=\mathcal S_K(C_t)+
\begin{cases}
\mathbb E\,\bigl(v_t(\xi)-\lambda_K(C_t)\bigr)_+,
&\vx\text{ on the OOD midplane},\\
\mathbb E\,\bigl(v_t(\xi+\zeta_t/2)-\lambda_K(C_t)\bigr)_+,
&\vx\text{ on either ID sheet}.
\end{cases}
\]
The strict ordering follows from the positive gap proved above.
\end{remark}

In the model, draw-to-draw fluctuation is the only variation within a class.
On real data, additional noise draws leave the AUROC almost unchanged
(Table~\ref{tab:ablation-sensitivity}, $N_\epsilon$), so the variation comes from the
clean images instead.

\paragraph{Connection to the Jacobian score.}
Miyasawa's identity~\eqref{eq:miyasawa} relates the Bayes
denoiser Jacobian to the same posterior covariance:
\[
J_{D_p}(\vx_t,t):=\frac{\partial D_p(\vx_t,t)}{\partial\vx_t}
=\frac{\alpha_t}{\beta_t^2}\operatorname{Cov}_p[X_1\mid X_t=\vx_t].
\]
Thus the eigenvalues of the Jacobian Gram $J_{D_p}^\top J_{D_p}$ are the
squared covariance eigenvalues scaled by $\alpha_t^2/\beta_t^4$. For any
$1\le K\le d$, the noise-averaged OOD--ID gap of its top-$K$ score is
\[
\frac{\alpha_t^2}{\beta_t^4}
\left[
\mathbb E\,\bigl(v_t(\xi)^2-\lambda_K(C_t)^2\bigr)_+
-\mathbb E\,\bigl(v_t(\xi+\zeta_t/2)^2-\lambda_K(C_t)^2\bigr)_+
\right].
\]
The same Gaussian comparison makes this gap strictly positive
exactly when $\lambda_K(C_t)<m^2$; otherwise it is zero.
This establishes the ordering for the Bayes output Jacobian.
Proposition~\ref{prop:tap-readout-two-sheet} in
Appendix~\ref{app:tap-readout-two-sheet} treats intermediate features
and their readouts.

\subsection{Extension to Gaussian mixtures}
\label{app:molrg-mixtures}

With more than two components, the denoiser again averages
over several possible explanations of the input.
For the mixture of low-rank Gaussians (MoLRG)
model~\citep{wang2025lowrank}, write
\begin{equation}
p=\sum_{k=1}^{L}\pi_k\mathcal N(\mu_k,\Sigma_k),
\qquad
\Sigma_k=A_kA_k^\top+\nu^2I_d,
\qquad
\pi_k>0,\quad\sum_k\pi_k=1.
\label{eq:molrg-model}
\end{equation}
Here $L$ is the number of components, $\pi_k$ is the probability
of component $k$, and $\mu_k\in\mathbb R^d$ and $\Sigma_k$ are
its mean and covariance. The matrix
$A_k\in\mathbb R^{d\times r}$, with $r<d$, describes variation
within the component's subspace. The parameter $\nu\ge0$
controls the standard deviation of isotropic variation around
that subspace. Let $Y\in\{1,\dots,L\}$ denote the component that
generates $X_1$. At a noisy input $\vx_t$, let $\rho_k$ be the posterior
probability that $Y = k$, and let $m_k,V_k$ be the corresponding
conditional mean and covariance, defined as
\[
\rho_k:=\mathbb P_p(Y=k\mid X_t=\vx_t), \:\:m_k :=\mathbb E_p[X_1\mid X_t=\vx_t,Y=k], \:\:V_k :=\operatorname{Cov}_p[X_1\mid X_t=\vx_t,Y=k].
\]
The Bayes denoiser averages the component predictions:
$D_p= \E[X_1\mid X_t=\vx_t] = \sum_k\rho_km_k$.
The posterior covariance separates uncertainty within a component from
disagreement between components.

\begin{proposition}[Uncertainty within and between components]
\label{prop:posterior-covariance-split}
At each noisy input, the posterior covariance is
\begin{equation}
\operatorname{Cov}_p[X_1\mid X_t=\vx_t]
=\underbrace{\sum_k\rho_kV_k}_{\text{within components}}
+\underbrace{\sum_k\rho_k(m_k-D_p)(m_k-D_p)^\top}
             _{\text{between components}}.
\label{eq:posterior-covariance-split}
\end{equation}
\end{proposition}

\begin{proof}
Within component $k$, write
$X_1-D_p=(X_1-m_k)+(m_k-D_p)$.
The cross terms have conditional mean zero because $\E[X_1 - m_k\mid X_t=\vx_t,Y=k] = 0$.
Averaging the remaining terms over the component probabilities
gives the identity.
\end{proof}

The first term measures uncertainty after the component is
known. The second measures disagreement between the component
predictions. It is positive semidefinite, so this disagreement
remains in the covariance even when the denoiser averages it
out of its prediction. An OOD--ID ordering still requires
assumptions about the two distributions.

For the synthetic calculations below, all these quantities are
available by Gaussian conditioning. With
$\Sigma_{k,t}=\alpha_t^2\Sigma_k+\beta_t^2I_d$ and $\sum_k\rho_k(\vx_t,t)=1$,
\[
\rho_k(\vx_t,t)
\propto\pi_k\mathcal N(\vx_t;\alpha_t\mu_k,\Sigma_{k,t}),\:\:
m_k(\vx_t,t)=\mu_k+\alpha_t\Sigma_k\Sigma_{k,t}^{-1}(\vx_t-\alpha_t\mu_k),\:\:
V_k(t)=\Sigma_k-\alpha_t^2\Sigma_k\Sigma_{k,t}^{-1}\Sigma_k.
\]
\paragraph{Synthetic setup.}
We compute the Bayes reconstruction and spectral scores in mixtures
beyond the matched two-sheet model.
We use $L\in\{2,4\}$ equally weighted components in $d=16$.
For $L=2$ the two means are $\pm a$ times a unit vector; for $L=4$ they lie
at distance $a$ from the origin along four orthonormal directions.
Each component has an independently sampled rank-$r$ subspace, spanned by
the orthonormal columns of $A_k$, and isotropic standard deviation
$\nu_{\mathrm{id}}=0.5$.
The OOD law is an isotropic Gaussian, centered at the mean
of the component centers, with standard deviation
$\nu_{\mathrm{ood}}$. Unless stated otherwise, $a=4$, $r=0$, and
$\nu_{\mathrm{ood}}=\nu_{\mathrm{id}}$.
Noise is added as $X_1+\sigma_tX_0$, where
$\sigma_t=\beta_t/\alpha_t$; this equals $X_t/\alpha_t$ in the notation
of Eq.~\eqref{eq:vp-forward}.

\begin{figure}[t]
\centering
\includegraphics[width=0.5\linewidth]{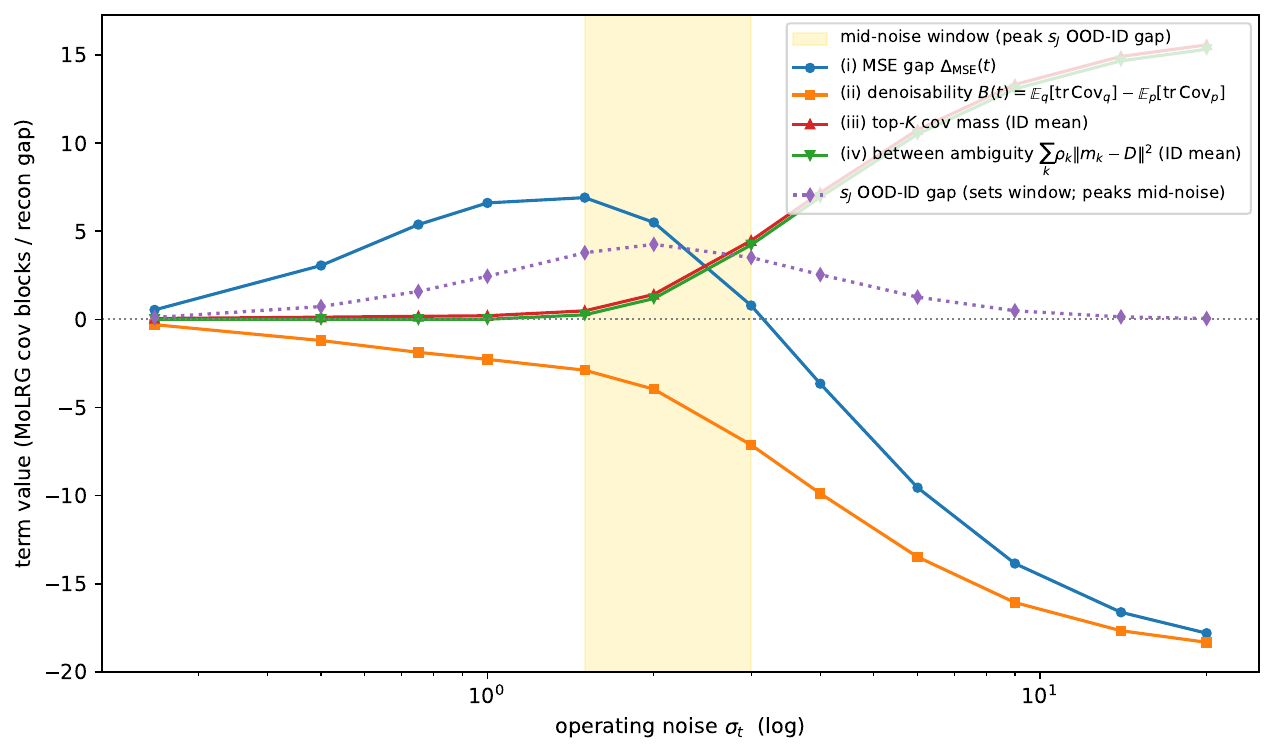}
\caption{
Noise dependence of the Bayes scores in the two-component mixture
($L=2$, $r=0$).
The reconstruction gap (blue) eventually becomes negative,
while the covariance spectral gap (purple, dotted) peaks at
intermediate noise and approaches zero at high noise.
}
\label{fig:molrg-noisetime}
\end{figure}

\paragraph{Effect of noise.}
Figure~\ref{fig:molrg-noisetime} varies the noise level in the
two-component configuration ($L=2$, $r=0$), with
$\nu_{\mathrm{ood}}=\nu_{\mathrm{id}}/2$.
The reconstruction gap (blue) eventually becomes negative.
The spectral gap (purple) peaks at intermediate noise and
remains positive after reconstruction inverts.
The negative denoisability gap (orange) indicates that OOD data are easier to denoise here.

At high noise, the top-$K$ covariance mass of ID inputs (red) closely
follows their between-component uncertainty (green), while the
spectral gap approaches zero. The ID posterior becomes
nearly independent of the observed input, so its uncertainty
remains large but differs little between ID and OOD inputs.
This agrees with Proposition~\ref{prop:high-noise-collapse}
(Appendix~\ref{app:molrg-tap-highnoise}).

\begin{figure}[t]
\centering
\includegraphics[width=0.98\linewidth]{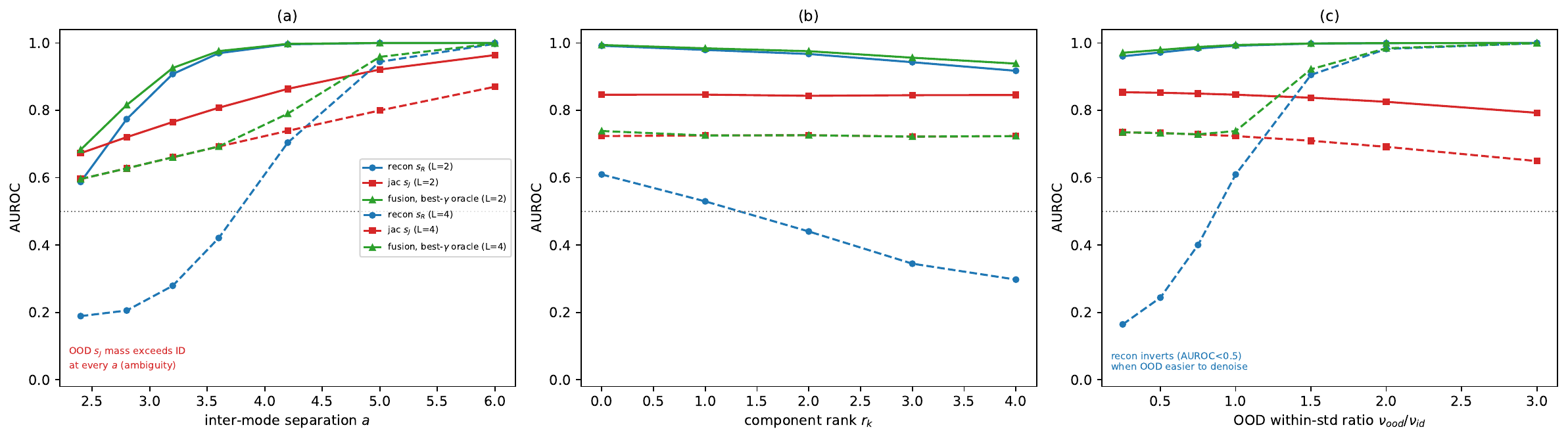}
\caption{
AUROC of Bayes reconstruction (blue), spectral scoring (red),
and oracle fusion (green) at $\sigma_t=2$ as we vary
\textbf{(a)} component separation,
\textbf{(b)} component rank, and
\textbf{(c)} OOD standard deviation relative to the ID isotropic
standard deviation.
Solid and dashed curves use two and four components, respectively.
The dotted line marks chance performance.
Spectral scoring remains above chance in settings where
reconstruction is inverted.
}
\label{fig:morlg_sweep}
\end{figure}

\paragraph{Parameter sweeps.}
We study two- and four-component mixtures at fixed $\sigma_t=2$,
where the spectral gap of the noise sweep is largest among the tested
noise levels (Figure~\ref{fig:molrg-noisetime}).
We vary three factors separately: component separation,
component rank, and OOD standard deviation
(Figure~\ref{fig:morlg_sweep}).

Each setting uses $1500$ ID test samples, $1500$ OOD test
samples, and $1500$ independent ID calibration samples.
Reconstruction MSE uses the Bayes denoiser. The spectral score uses the exact ID
posterior covariance at one noise draw, retaining
$K_{\mathrm{spec}}=\min\{d,(L-1)+Lr\}$ eigenvalues, the number of
between-component and within-subspace directions. This gives
$K_{\mathrm{spec}}=1$ in the noise sweep of
Figure~\ref{fig:molrg-noisetime} ($L=2$, $r=0$).
Both scores are standardized using ID calibration data.
For fusion, we report the best AUROC over a weight grid
that includes both individual channels. This oracle uses
OOD data and measures the potential benefit of fusion;
it does not evaluate the deployed weight.

\paragraph{Results.}
The four-component mixtures show three distinct effects.
In Figure~\ref{fig:morlg_sweep}(a), increasing component
separation restores the reconstruction ordering, while
spectral AUROC remains above chance throughout.
In panel (b), increasing component rank lowers reconstruction
AUROC from approximately $0.61$ to $0.30$, while spectral
AUROC stays near $0.72$ with the stated choice of $K_{\mathrm{spec}}$.
In panel (c), compact OOD data receive an inverted reconstruction
ordering; increasing their spread restores the ordering.
The spectral score remains above chance across this sweep.

Together, these results show that reconstruction ordering
depends strongly on mixture geometry and OOD variability,
while spectral information can remain useful when reconstruction
inverts. This behavior is consistent with the branch-ambiguity
mechanism in mixtures beyond the theorem's matched-sheet
setting. At $\sigma_t=2$, reconstruction does not invert in any tested
two-component configuration; in the two-component noise sweep it
inverts only at higher noise (Figure~\ref{fig:molrg-noisetime}).

\subsection{The high-noise limit}
\label{app:molrg-tap-highnoise}

At the high-noise endpoint, the observation carries no
information about the clean sample. The Bayes denoiser
therefore returns the ID mean, and its posterior uncertainty
returns to the ID prior covariance.

\begin{proposition}[Bayes scores at high noise]
\label{prop:high-noise-collapse}
Let $p$ be the finite Gaussian mixture above, and let $q$
have finite second moment. Write $\mu_p,\mu_q$ for their
means and $\Sigma_p,\Sigma_q$ for their covariances.
As $\alpha_t\to0$ and $\beta_t\to1$, for every fixed $\vx_t$,
\[
D_p(\vx_t,t)\to\mu_p,\qquad
\operatorname{Cov}_p[X_1\mid X_t=\vx_t]\to\Sigma_p,\qquad
\|J_{D_p}(\vx_t,t)\|_2=O(\alpha_t).
\]
For any fixed $1\le K_{\mathrm{spec}}\le d$, the expected
OOD--ID posterior-covariance spectral gap tends to zero,
whereas
\[
\Delta_{\mathrm{MSE}}(t)\to
\operatorname{tr}\Sigma_q-\operatorname{tr}\Sigma_p
+\|\mu_q-\mu_p\|_2^2.
\]
\end{proposition}

\begin{proof}
The Gaussian conditioning formulas of Appendix~\ref{app:molrg-mixtures} give
$\rho_k(\vx_t,t)\to\pi_k$, $m_k(\vx_t,t)\to\mu_k$, and
$V_k(t)\to\Sigma_k$, uniformly on bounded sets of inputs.
Substituting these limits into $D_p=\sum_k\rho_km_k$ and
Eq.~\eqref{eq:posterior-covariance-split} gives $\mu_p$ and $\Sigma_p$.
Miyasawa's identity then gives the Jacobian rate.
To pass to expected scores, the same formulas give, near
this endpoint,
\[
\|D_p(\vx_t,t)\|_2^2+
\operatorname{tr}\operatorname{Cov}_p[X_1\mid X_t=\vx_t]
\le \kappa(1+\|\vx_t\|_2^2),
\]
where $\kappa$ is independent of $\vx_t$ and $t$.
Under $X_t=\alpha_tX_1+\beta_tX_0$, this bounds the spectral
score and reconstruction loss by an integrable multiple of
$1+\|X_1\|_2^2+\|X_0\|_2^2$.
Dominated convergence therefore gives the same limiting
spectral score $\mathcal S_{K_{\mathrm{spec}}}(\Sigma_p)$
under ID and OOD.
The limiting reconstruction loss is $\|X_1-\mu_p\|_2^2$;
its OOD-minus-ID expectation gives the stated expression.
\end{proof}
The vanishing spectral gap at high noise motivates excluding this regime
when selecting sensitivity noise levels, consistent with the empirical
noise sweep in Figure~\ref{fig:noise-sweep}
(Appendix~\ref{app:feature-selection}).

\subsection{Branch uncertainty in feature sensitivity}
\label{app:tap-readout-two-sheet}

The preceding analysis identifies branch uncertainty as a signal
that reconstruction can miss. We now express the Bayes denoiser
through a feature containing the branch probability and the
predicted position along the sheets. We show that this feature's
sensitivity measures the same uncertainty. We then show that the Bayes
output can lose this signal from its top-$K$ spectrum, and that another
readout of the same feature can reverse the output's ordering at
near-optimal denoising risk, while the feature's ordering is unchanged.

\paragraph{A feature representation of the Bayes denoiser.}
We use the two-sheet model of Appendix~\ref{app:branch-dichotomy}
at a fixed interior noise level. Recall that $m=\omega/2$,
$U=e_b^\top\vx_t/\beta_t$, and
$\zeta_t=\alpha_t\omega/\beta_t$.
Let $\rho_1$ be the posterior probability of the positive branch and
$\widehat{\mathbf{w}}_t$ the predicted position along the sheets:
\[
\begin{aligned}
\rho_1(\vx_t)
&:=\mathbb P_p(S=+1\mid X_t=\vx_t)
  =\frac{1}{1+e^{-\zeta_tU}},\\
\widehat{\mathbf{w}}_t(\vx_t)
&:=\mathbb E_p[W\mid X_t=\vx_t]
  =\alpha_t\Sigma
   (\alpha_t^2\Sigma+\beta_t^2I_d)^{-1}\vx_t,\\
\phi(\vx_t)
&:=\bigl(\rho_1(\vx_t),\widehat{\mathbf{w}}_t(\vx_t)\bigr).
\end{aligned}
\]
The Bayes denoiser combines these two quantities to predict the
clean input:
\[
D_p(\vx_t,t)
=\widehat{\mathbf{w}}_t(\vx_t)
 +\omega\bigl(\rho_1(\vx_t)-\tfrac12\bigr)e_b.
\]

For a feature or output map $h$, write
$J_h:=\partial h/\partial\vx_t$. With $\mathcal S_K$ and $(z)_+$ as in
Appendix~\ref{app:branch-dichotomy}, we use the same spectral statistic
at the feature and at the output:
\[
\begin{aligned}
s_{h,K}(\vx_t)
&:=\mathcal S_K\bigl(J_h(\vx_t)^\top J_h(\vx_t)\bigr),\\
\Delta_{h,K}
&:=\mathbb E_{X_t\sim q_t}s_{h,K}(X_t)
  -\mathbb E_{X_t\sim p_t}s_{h,K}(X_t).
\end{aligned}
\]
A positive gap means that OOD inputs have a higher mean score.
These expectations include the forward noise.

The within-sheet prediction is linear, so its Jacobian is constant.
Define
\[
J_w:=\frac{\partial\widehat{\mathbf{w}}_t}{\partial\vx_t},
\qquad
H_t:=J_w^\top J_w,
\qquad
c_K:=\lambda_K(H_t),
\]
where eigenvalues are ordered decreasingly, including zeros.
The branch contributes a direction orthogonal to these
within-sheet responses. To affect the top-$K$ score, its
sensitivity must exceed the cutoff $c_K$.

\begin{proposition}[Branch uncertainty appears in feature sensitivity]
\label{prop:tap-readout-two-sheet}
Fix an interior noise level and $1\le K=K_{\mathrm{spec}}\le d$, and let
$v_t(u)=m^2\operatorname{sech}^2(\zeta_tu/2)$ be the branch variance of
Appendix~\ref{app:branch-dichotomy}.
The sensitivity of the branch probability is proportional to
the squared Bayes branch variance:
\begin{equation}
\bigl\|\nabla_{\vx_t}\rho_1(\vx_t)\bigr\|_2^2
=\frac{\alpha_t^2}{\omega^2\beta_t^4}\,v_t(U)^2
=:a_t(U).
\label{eq:feature-uncertainty-link}
\end{equation}
The feature score is therefore
\begin{equation}
s_{\phi,K}(\vx_t)
=\mathcal S_K(H_t)+(a_t(U)-c_K)_+.
\label{eq:tap-feature-score}
\end{equation}
Its mean gap satisfies $\Delta_{\phi,K}>0$ whenever
\begin{equation}
c_K<a_t(0)=\frac{\zeta_t^2}{16\beta_t^2}.
\label{eq:tap-branch-cutoff}
\end{equation}
\end{proposition}

The proof below uses the observation that the branch posterior is most sensitive near the midplane, where both ID branches have the same posterior probability. Because noisy OOD inputs visit this region more frequently than noisy ID inputs, their average branch sensitivity is larger. Their mean feature sensitivity is
therefore higher whenever the branch response can enter the
retained spectrum.

\begin{proof}
Differentiating the branch probability gives
\[
\nabla_{\vx_t}\rho_1
=\frac{\zeta_t}{4\beta_t}
 \operatorname{sech}^2(\zeta_tU/2)e_b,
\qquad
a_t(u)=\frac{\zeta_t^2}{16\beta_t^2}
        \operatorname{sech}^4(\zeta_tu/2).
\]
Using $m=\omega/2$ and $\zeta_t=\alpha_t\omega/\beta_t$
gives Eq.~\eqref{eq:feature-uncertainty-link}.
Since $J_we_b=0$, stacking the branch probability and the
within-sheet prediction gives
\[
J_\phi^\top J_\phi=H_t+a_t(U)e_be_b^\top.
\]
The branch direction is orthogonal to the range of $H_t$.
Its eigenvalue increases the top-$K$ sum precisely when it
exceeds $c_K$, giving
Eq.~\eqref{eq:tap-feature-score}.

Under OOD, $U=\xi$ with $\xi\sim\mathcal N(0,1)$; under ID,
$U=\xi\pm\zeta_t/2$ with equal probabilities.
As shown in the proof of Theorem~\ref{thm:branch-dichotomy},
\[
\mathbb P(|\xi|<b)
>\mathbb P(|\xi+\zeta_t/2|<b),\qquad b>0.
\]
The input-dependent part of the score,
$(a_t(U)-c_K)_+$, is even and decreases with $|U|$ wherever
it is positive. Under Eq.~\eqref{eq:tap-branch-cutoff}, its
nonempty positive level sets are intervals centered at zero.
OOD inputs reach each such level more often than ID inputs;
integrating these probabilities gives $\Delta_{\phi,K}>0$.
\end{proof}

The proposition connects the two analyses directly: the feature
responds to the same branch uncertainty studied in the Bayes
covariance. The cutoff condition~\eqref{eq:tap-branch-cutoff} concerns
whether that response is retained by the spectral score. For $K>r$, we
have $c_K=0$, so the condition holds automatically. For $K\le r$, it
specifies how strong the branch response must be.

\begin{remark}[A feature signal can be absent from the output spectrum]
\label{rem:feature-output-cutoff}
The Bayes output multiplies the branch probability by $\omega$
while leaving the within-sheet prediction unchanged. Its branch contribution to the Gram is therefore multiplied
by $\omega^2$. Since $J_w$ is symmetric and $J_we_b=0$,
the cross terms vanish, giving
$J_{D_p}^\top J_{D_p}=H_t+\omega^2a_t(U)e_be_b^\top$.
The output score is then
\[
s_{D_p,K}
=\mathcal S_K(H_t)+(\omega^2a_t(U)-c_K)_+.
\] Consequently, if
$\omega^2a_t(0)\le c_K<a_t(0)$, which requires $\omega<1$,
then
\begin{equation}
\Delta_{\phi,K}>0,
\qquad
s_{D_p,K}\equiv\mathcal S_K(H_t),
\qquad
\Delta_{D_p,K}=0.
\label{eq:tap-readout-orderings}
\end{equation}
The feature retains a varying branch response among its top
directions, while the output score contains only the constant
within-sheet contribution. This comparison depends on the relative
weight of those directions in the chosen representation.
\end{remark}
To illustrate the zero-gap window $\omega^2a_t(0)\le c_K<a_t(0)$ of Remark~\ref{rem:feature-output-cutoff} with $K<r$, take $d=64$, $r=8$, $K=3$,
$\alpha_t=\beta_t=1/\sqrt2$, and $\omega=1/2$, with all eight
nonzero eigenvalues of $\Sigma$ equal to $1/9$. Then
\[
\omega^2a_t(0)=\frac{1}{128}
<c_3=\frac{1}{50}
<a_t(0)=\frac{1}{32}.
\]
This example also has an inverted reconstruction gap.
Indeed, $\zeta_t=1/2$, and
Eq.~\eqref{eq:branch-gap-short} together with
$\tanh^2 z\le z^2$ gives
\[
\frac{\Delta_{\mathrm{MSE}}(t)}{m^2}
\le -1+\frac{\zeta_t^2}{2}+\frac{\zeta_t^4}{16}<0.
\]
Thus the same Bayes denoiser can have an inverted mean MSE gap
and a constant output spectral score, while its feature has a
positive mean spectral gap.

For well-separated branches ($\zeta_t\ge4$), the Bayes output is correctly
ordered, but near-optimal denoising risk does not protect this ordering.
Flatten the Bayes branch prediction on the band around the midplane where
noisy OOD inputs are more frequent than ID inputs, and blend it back just
outside. This gives a smooth readout $\widehat D$ of the same feature whose
branch sensitivity vanishes where OOD inputs are more frequent and exceeds
the cutoff where ID inputs are more frequent, so $\Delta_{\widehat D,K}<0$,
while $\Delta_{\phi,K}>0$ is unchanged. Since ID inputs rarely visit the band,
the ID denoising risk of $\widehat D$ exceeds that of $D_p$ by at most
$2\omega^2e^{-\zeta_t^2/128}/(\sqrt{2\pi}\,\zeta_t)$.

\begin{figure}[t]
\centering
\includegraphics[width=.5\linewidth]{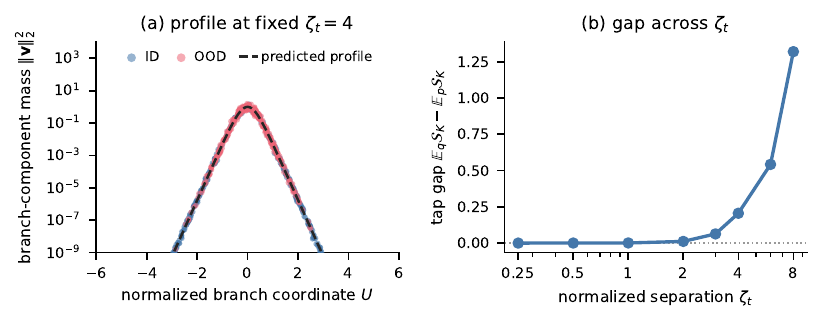}
\caption{Feature sensitivity in the two-sheet model.
\textbf{(a)} Measured branch sensitivity at $\zeta_t=4$ and
the predicted profile (dashed).
\textbf{(b)} Estimated mean OOD-minus-ID feature score as the
normalized separation $\zeta_t$ varies.}
\label{fig:molrg-numerical-check}
\end{figure}

We also check
Proposition~\ref{prop:tap-readout-two-sheet} numerically with a
finite-difference probe estimate of the feature Gram ($M=32$,
$\delta=0.1$, $K=3$), using $\omega=4$, $d=64$, $r=8$, and $400$ inputs
per distribution; the within-sheet part of the feature is passed through a
fixed random linear map, to which the proof applies unchanged.
At $\zeta_t=4$, measured branch sensitivity peaks near the
midplane and follows the predicted profile
(Figure~\ref{fig:molrg-numerical-check}(a)); the theoretical curve
includes the $M/d$ normalization factor of the probe Gram.
ID and OOD inputs follow the same profile, but noisy OOD inputs
visit the region of high sensitivity more often.
The estimated mean feature score is higher for OOD at every
tested separation $\zeta_t\ge0.5$, and the gap grows over the
tested range (Figure~\ref{fig:molrg-numerical-check}(b)).
Appendix~\ref{app:tap-output-check} examines the corresponding
feature-versus-output behavior in trained U-Nets, using the tap
selected by Algorithm~\ref{alg:selection} for CIFAR-10.

\section{Implementation details} \label{app:implementation}
This appendix describes the experimental setup and how the
detector is configured. We first list the datasets, baselines,
denoisers, and calibration protocol (\ref{subsec:baselines-datasets}).
We then give the sensitivity estimator and the ID-only selection of
the decoder tap and sensitivity noise levels, with its empirical
validation (\ref{app:feature-selection}). Finally, we describe how
the fusion weight is evaluated using ID data alone (\ref{app:id-gamma}).
\subsection{Baselines and datasets}
\label{subsec:baselines-datasets}

We follow the evaluation protocol of EigenScore~\citep{shoushtari2026eigenscore} for datasets, baselines, and diffusion models whenever possible, so that the comparison aligns directly with prior diffusion-based OOD detection results.

\paragraph{Datasets.}
For the main $32{\times}32$ benchmark, images are resized to $32{\times}32$ when necessary, and we use the official train/test splits. For each ID--OOD pair, a held-out ID calibration split is used for score normalization, feature selection, noise-level selection, and fusion-weight evaluation, while a disjoint test split is used for AUROC evaluation. \method\ uses no OOD samples during calibration or feature selection. Additional results on ImageNet-64, FFHQ-256, and extended OOD datasets are provided in Appendix~\ref{app:additional-experiments}. For calibration, we randomly sample 500 ID images; for testing, we sample a disjoint set of 500 ID and 500 OOD images.

\paragraph{Baselines.}
We compare against the representative generative OOD detectors listed in Section~\ref{subsec:exp-setup}, including likelihood-based, reconstruction-based, and diffusion-based methods. For Glow-based baselines such as DoS, TT, and WAIC, we use the sample-wise scoring protocol of \citet{morningstar2021density}. For diffusion-based baselines, we use the official implementations when available and otherwise follow the evaluation settings reported in the corresponding papers. For the main benchmark, DiffPath and DiffPathV2 use a single EDM diffusion model trained on CelebA. See \citet[Fig.~5]{shoushtari2026eigenscore} for an analysis of DiffPath with this model. 

\paragraph{Denoisers and calibration.}
For the $32{\times}32$ experiments, the denoisers are trained with the EDM framework~\citep{karras2022elucidating} on the same train splits as prior work. These denoisers are $\sigma$-conditioned: the denoiser is called as $D_\theta(\cdot,\sigma_\tau)$ with $\sigma_\tau=\sqrt{1/\bar\alpha_\tau-1}$ under the linear-$\beta$ DDPM schedule, where $\bar\alpha_\tau$ is the cumulative schedule product. Thus $\alpha_\tau=\sqrt{\bar\alpha_\tau}$ and $\beta_\tau=\sqrt{1-\bar\alpha_\tau}$ in Eq.~\eqref{eq:vp-forward}, and a discrete timestep $\tau$ indexes the noise level through $\sigma_\tau$. The FFHQ-256 denoiser uses the guided-diffusion architecture~\citep{dhariwal2021diffusion} and is conditioned on the raw integer timestep. The ImageNet-64 denoiser is the unconditional improved-diffusion model~\citep{nichol2021improved}; its network is conditioned on the integer timestep rescaled by $1000/4000$, whereas $\bar\alpha_\tau$ remains indexed by the raw integer. For \method, all detector-specific statistics are computed using the frozen ID-trained denoiser. Unless otherwise stated, calibration uses only ID samples and is used to estimate channel-normalization statistics, select the sensitivity tap, choose the sensitivity noise levels, and evaluate the fusion weight $\hat\gamma$ (Appendix~\ref{app:id-gamma}). The OOD-assisted variant \method-OODVal (Appendix~\ref{app:oodval}) uses a disjoint validation split from one auxiliary OOD dataset only for feature tap and sensitivity-noise selection; all other hyperparameters remain fixed.

\subsection{Sensitivity estimator and feature selection}
\label{app:feature-selection}

\paragraph{Sensitivity estimator.} The sensitivity score $s_J$~\eqref{eq:sJ} uses a central-difference estimate of the local feature response. Throughout, the noised input is formed in the $\sigma$-parameterization used by the $32{\times}32$ denoisers: $\vx_\tau^{(r)}=\vx+\sigma_\tau\epsilon^{(r)}$, where $\epsilon^{(r)}\sim\mathcal{N}(0,I_d)$ is drawn independently for each repetition $r=1,\dots,N_\epsilon$. This input equals $\vx_\tau/\alpha_\tau$ under Eq.~\eqref{eq:vp-forward} with $\sigma_\tau=\beta_\tau/\alpha_\tau$. For each repetition, we draw $M$ probe directions $\vv_1^{(r)},\dots,\vv_M^{(r)}$ i.i.d.\ from $\mathcal{N}(0,I_d)$, normalize them to unit Euclidean norm, and form the response matrix $R^{(r)}_{\ell,\tau}(\vx)\in\mathbb{R}^{M\times m_\ell}$ with rows
\begin{equation} \big[R^{(r)}_{\ell,\tau}(\vx)\big]_{j\cdot} = \frac{ \phi_\ell\big(\vx_\tau^{(r)}+\delta \vv_j^{(r)},\tau\big)-\phi_\ell\big(\vx_\tau^{(r)}-\delta \vv_j^{(r)},\tau\big) }{2\delta}, \qquad j=1,\dots,M, \label{eq:feature-response} \end{equation}
at probe scale $\delta=0.1$. The per-sample, per-timestep sensitivity statistic used by Algorithms~\ref{alg:detect} and~\ref{alg:selection} is the top-$K_{\mathrm{spec}}$ spectral mass of the probe Gram, averaged over the $N_\epsilon$ repetitions:
\begin{equation} e_{\ell,\tau}(\vx) = \frac{1}{N_\epsilon}\sum_{r=1}^{N_\epsilon} \sum_{k=1}^{K_{\mathrm{spec}}} \lambda_k\Big(R^{(r)}_{\ell,\tau}(\vx)\,R^{(r)}_{\ell,\tau}(\vx)^{\top}\Big), \label{eq:sensitivity-cov} \end{equation}
i.e.\ the sum of the $K_{\mathrm{spec}}$ largest squared singular values of the response matrix. Up to the $1/M$ normalization, which is absorbed by the per-timestep standardization of Algorithm~\ref{alg:detect}, $R^{(r)\top}R^{(r)}$ is the empirical sensitivity covariance $C_\ell$ of Section~\ref{subsec:sensitivity}. The method does not require forming the full Jacobian $J_\ell$; it only measures local feature responses.

 \paragraph{Reconstruction statistic.} The per-timestep reconstruction statistic is the per-pixel denoising MSE, evaluated on a fixed timestep grid (sensitivity noise pool $\mathcal{P}$).
 
\begin{algorithm}[t]
\caption{\textsc{ID-only selection of sensitivity features}.}
\label{alg:selection}
\begin{minipage}[t]{0.49\linewidth}
\textbf{A. Tap selection by proxy shifts}
\begin{algorithmic}[1]
\Require Frozen denoiser $D_\theta$; ID calibration set $\mathcal{X}_{\mathrm{cal}}$; candidate taps $\mathcal{L}$; proxy shifts $\mathcal{G}$; sensitivity noise pool $\mathcal{P}$; numerical stabilizer $\eta$ as in Eq.~\eqref{eq:zchannels}
\Ensure Selected tap $\ell^\star$
\For{$\ell\in\mathcal{L}$}
    \For{$\tau\in\mathcal{P}$}
        \State $E_{\ell,\tau}^{\mathrm{id}}
        \gets
        \{e_{\ell,\tau}(\vx):\vx\in\mathcal{X}_{\mathrm{cal}}\}$
        \For{$g\in\mathcal{G}$}
            \State $E_{\ell,\tau}^{g}
            \gets
            \{e_{\ell,\tau}(g\vx):\vx\in\mathcal{X}_{\mathrm{cal}}\}$
            \State $a_{\ell,\tau}^{g}
            \gets
            \mathrm{AUROC}(E_{\ell,\tau}^{\mathrm{id}},E_{\ell,\tau}^{g})$
        \EndFor
        \State $A_{\mathrm{geom}}(\ell,\tau)
        \gets
        |\mathcal{G}|^{-1}
        \sum_{g\in\mathcal{G}}a_{\ell,\tau}^{g}$
    \EndFor
    \State $S_{\mathrm{tap}}(\ell)
    \gets
    |\mathcal{P}|^{-1}
    \sum_{\tau\in\mathcal{P}}
    A_{\mathrm{geom}}(\ell,\tau)$
\EndFor
\State \Return $\ell^\star=\argmax_{\ell\in\mathcal{L}}S_{\mathrm{tap}}(\ell)$
\end{algorithmic}
\end{minipage}
\hfill
\begin{minipage}[t]{0.49\linewidth}
\textbf{B. Noise selection by ID stability}
\begin{algorithmic}[1]
\Require Frozen denoiser $D_\theta$; ID calibration set $\mathcal{X}_{\mathrm{cal}}$; selected tap $\ell^\star$; sensitivity noise pool $\mathcal{P}$; ensemble size $K_{\mathrm{ens}}$
\Ensure Selected noise levels $\mathcal{T}^\star$ and weights $\{w_\tau\}$
\For{$\tau\in\mathcal{P}$}
    \State $E_\tau
    \gets
    \{e_{\ell^\star,\tau}(\vx):\vx\in\mathcal{X}_{\mathrm{cal}}\}$
    \State $\mathrm{CV}(\tau)
    \gets
    \dfrac{\operatorname{std}(E_\tau)}
    {|\operatorname{mean}(E_\tau)|+\eta}$
\EndFor
\State $\mathcal{T}^\star
\gets$
the $K_{\mathrm{ens}}$ levels in $\mathcal{P}$ with smallest $\mathrm{CV}(\tau)$
\For{$\tau\in\mathcal{T}^\star$}
    \State $w_\tau
    \gets
    \dfrac{\mathrm{CV}(\tau)^{-1}}
    {\sum_{\tau'\in\mathcal{T}^\star}\mathrm{CV}(\tau')^{-1}}$
\EndFor
\State \Return $\mathcal{T}^\star,\{w_\tau\}_{\tau\in\mathcal{T}^\star}$
\end{algorithmic}
\end{minipage}
\end{algorithm}

\paragraph{How and why the ID-only selection works.}
Algorithm~\ref{alg:selection} selects both the feature tap and the sensitivity timesteps using only the ID calibration set. We focus on decoder taps because these blocks integrate decoder features with skip-connected encoder information, making them natural points to probe the combined representation before the final output mapping. Section~\ref{subsec:theory-feature-sensitivity} discusses why such intermediate responses may be attenuated downstream, and Appendix~\ref{app:layer-ablation} evaluates this choice empirically. In Stage~A, for each candidate decoder tap $\ell$ and timestep $\tau\in\mathcal{P}$, we evaluate the sensitivity statistic on the original calibration images and on geometrically transformed copies $g(\vx)$, with $g\in\mathcal{G}$. We compute the AUROC separating the two sets, average it over transformations and candidate timesteps, and select the tap with
the largest resulting proxy score. In Stage~B, we fix the selected tap $\ell^\star$, evaluate its sensitivity statistic on untransformed ID samples at each candidate timestep, rank the timesteps by coefficient of variation, retain the $K_{\mathrm{ens}}$ most stable levels, and combine them using normalized inverse-CV weights.

The standardized mean gap~\eqref{eq:deflection-definition} clarifies the role of these two stages. Detection quality depends on an OOD-induced shift in the score, measured relative to the score's variability under ID data. The shift is unavailable during selection, whereas the ID variability can be estimated from the calibration split. Stage~A therefore uses controlled structural perturbations as a surrogate for
the unknown OOD shift. The transformations in $\mathcal{G}$ preserve much of the image content and low-level appearance while disrupting the spatial characteristics of the ID distribution. A tap whose sensitivity spectrum changes little under these perturbations is less likely to expose local-response directions associated with ID structure, whereas a tap that clearly separates clean images from their transformed counterparts provides a structurally informative sensitivity statistic. This criterion does not assume that the proxy
shifts match the eventual OOD distribution; it uses them only to rank taps by their response to violations of ID structure.

Stage~B targets the ID variability of the statistic. Because each
$e_{\ell^\star,\tau}$ is calibrated relative to its ID distribution, an OOD-induced displacement is useful only if it is large compared with the fluctuation of the statistic at that timestep. The coefficient of variation measures this fluctuation relative to the statistic's scale: low-CV timesteps yield more stable and well-conditioned calibration statistics, whereas high-CV timesteps are more sensitive to sampling noise and atypical ID responses. The
inverse-CV weights therefore assign greater influence to the more stable timesteps. Thus, Stage~A selects \emph{what representation} to probe, while Stage~B selects \emph{where along the diffusion path} that representation can be measured most reliably.

\paragraph{Tap-selection results.} We analyze the ID-only tap-selection criterion used in Algorithm~\ref{alg:selection}. Recall that each candidate decoder tap $\ell$ is scored by $S_{\mathrm{tap}}(\ell)$, the average AUROC with which its sensitivity statistic separates ID calibration samples from their geometrically transformed counterparts. This score is not computed using target OOD data; it is only a proxy for whether a tap exposes sensitivity structure that responds to ID-derived distributional perturbations.

\begin{figure}[t]
    \centering
    \includegraphics[width=.75\linewidth]{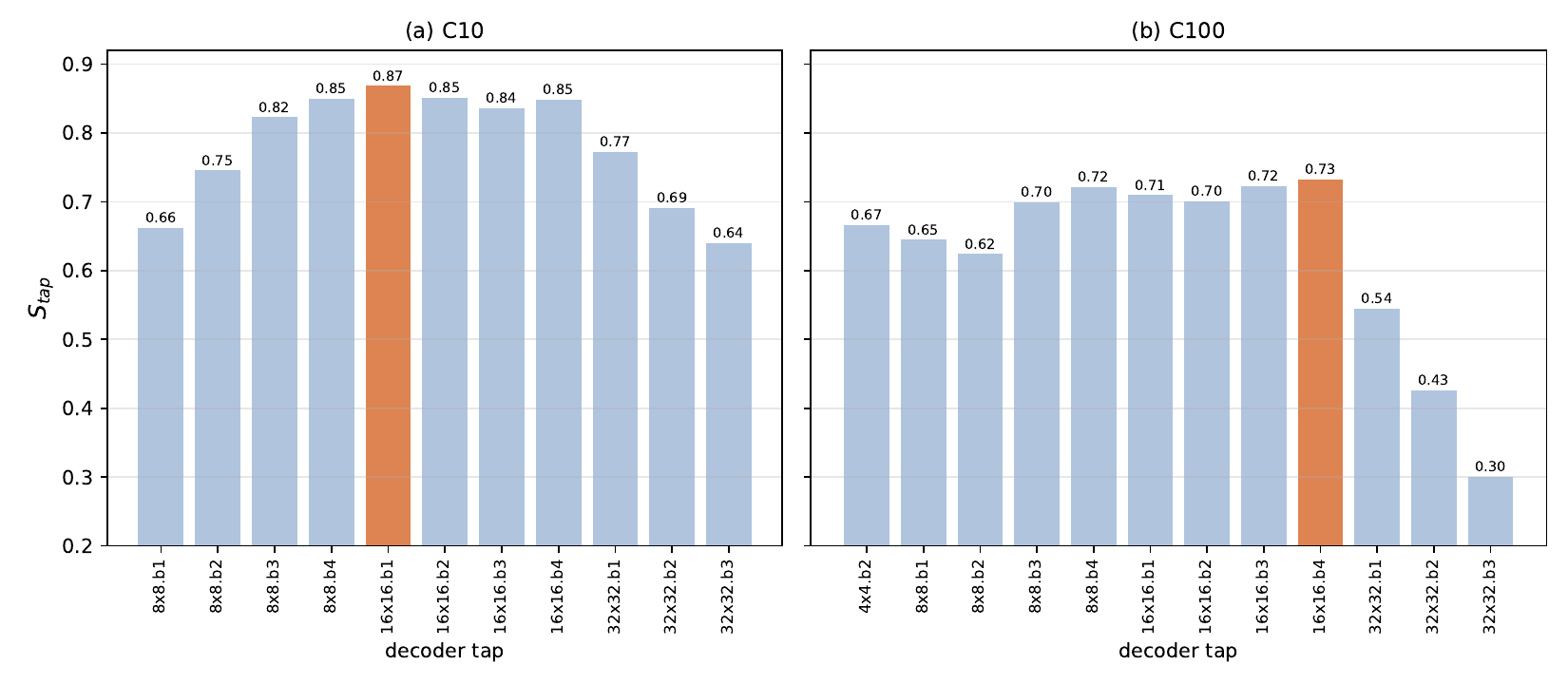}
    \caption{
    \textbf{ID-only feature tap selection on CIFAR datasets.}
    We report the proxy selection score $S_{\mathrm{tap}}(\ell)$ for each candidate decoder tap $\ell$ on
    (a) CIFAR-10 and (b) CIFAR-100. The selected tap is highlighted in orange. The score is computed only
    from ID calibration samples and their geometric proxy shifts, without using target OOD data. For both
    datasets, the most informative sensitivity taps occur in intermediate decoder blocks rather than at the
    earliest or latest layers.
    }
    \label{fig:tap-selection-cifar}
\end{figure}

\begin{figure}[t]
    \centering
    \includegraphics[width=0.5\linewidth]{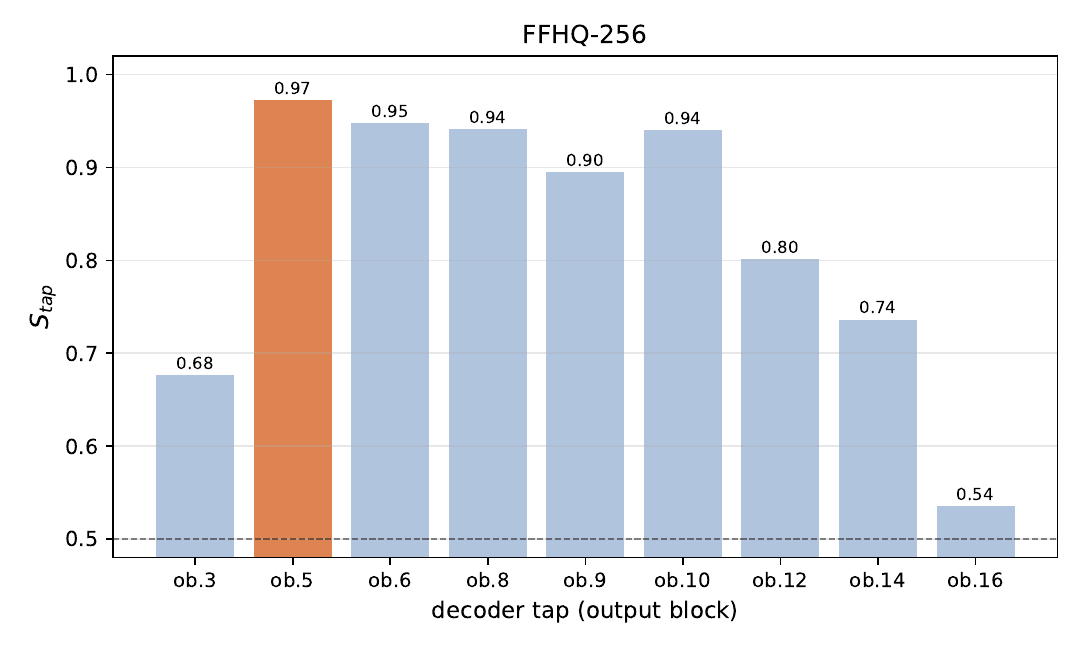}
    \caption{
    \textbf{ID-only feature tap selection on FFHQ-256.}
    We report $S_{\mathrm{tap}}(\ell)$ across candidate decoder blocks; the \texttt{ob} labels index
    the output (decoder) blocks of the guided-diffusion U-Net. The selected block is highlighted
    in orange, and the dashed line marks chance level. The ID-only proxy criterion selects an early-to-middle output block, indicating that useful
    sensitivity information is strongest before the final reconstruction layers fully compress the denoising
    representation.
    }
    \label{fig:tap-selection-ffhq}
\end{figure}

Figure~\ref{fig:tap-selection-cifar} shows $S_{\mathrm{tap}}(\ell)$ for every candidate tap of the CIFAR-10 and CIFAR-100 denoisers, and Figure~\ref{fig:tap-selection-ffhq} shows the same score for FFHQ-256. For CIFAR-10, the score increases from early $8{\times}8$ decoder blocks to a maximum at \texttt{16x16.b1} ($S_{\mathrm{tap}}=0.87$) and then decreases toward later $32{\times}32$ blocks. CIFAR-100 shows a similar pattern, with the best tap at \texttt{16x16.b4} ($S_{\mathrm{tap}}=0.73$). The same criterion selects \texttt{8x8.b1} for both SVHN ($S_{\mathrm{tap}}=0.69$) and CelebA ($S_{\mathrm{tap}}=0.82$). For FFHQ-256, the best output block is \texttt{ob.5} ($S_{\mathrm{tap}}=0.97$), while both earlier and substantially later blocks are less informative.

These results support two observations. First, sensitivity information is not uniformly distributed across the denoiser: some taps expose substantially stronger local-response signals than others. Second, the most useful taps tend to occur in intermediate or mid-to-late decoder layers, where the representation has recovered meaningful image structure but has not yet been fully compressed into the final reconstruction. This justifies selecting the tap from ID calibration data rather than fixing the layer in advance.

\paragraph{Noise-level selection.}
After selecting the feature tap, we choose the sensitivity noise levels using the
ID-only stability criterion in Algorithm~\ref{alg:selection}. For the $32{\times}32$ denoisers,
the candidate pool is
\[
    \mathcal{P}=\{10,25,50,100,150\}.
\]
For each candidate level, we compute the sensitivity scores on the ID calibration split,
rank levels by their coefficient of variation, and retain the $K_{\mathrm{ens}}$ most
stable levels. This step uses only
ID samples and does not use any target OOD data.

\begin{figure}[t]
    \centering
    \begin{subfigure}[t]{0.8\linewidth}
        \centering
        \includegraphics[width=\linewidth]{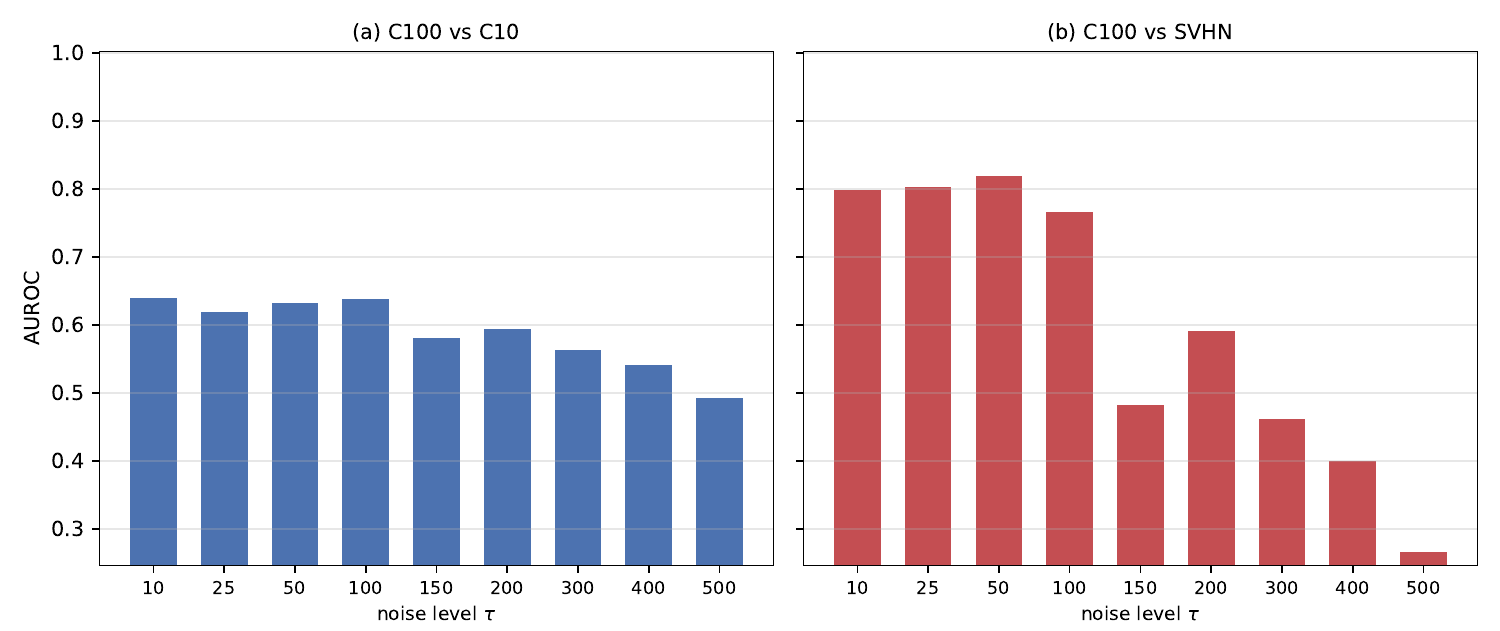}
        \caption{CIFAR-100 ID with the selected tap fixed.}
        \label{fig:noise-sweep-c100}
    \end{subfigure}

    \vspace{0.8em}

    \begin{subfigure}[t]{0.8\linewidth}
        \centering
        \includegraphics[width=\linewidth]{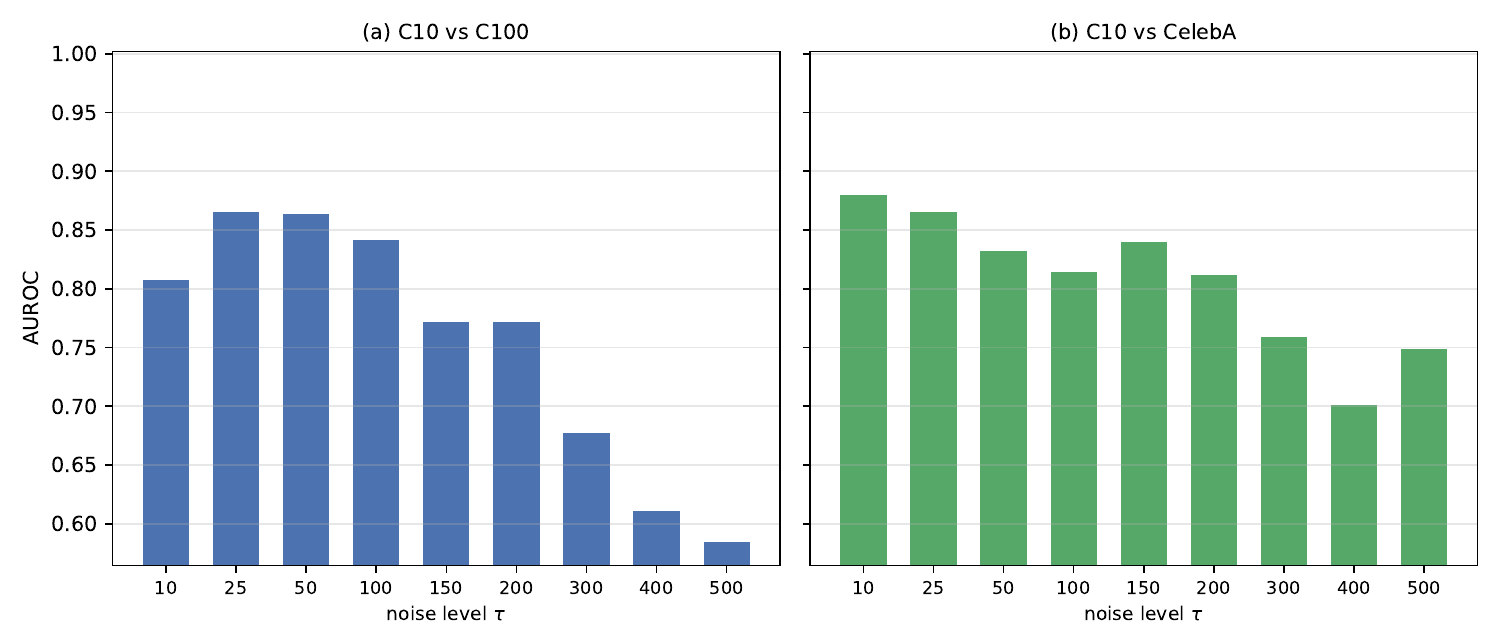}
        \caption{CIFAR-10 ID with the selected tap fixed.}
        \label{fig:noise-sweep-c10}
    \end{subfigure}

    \caption{
    \textbf{Retrospective sweep over sensitivity noise levels.}
    We fix the selected decoder tap and the reconstruction channel and sweep the single
    noise level $\tau$ used by the sensitivity channel. Bars show the AUROC of the fused score at the deployed weight $\hat\gamma$ (Appendix~\ref{app:id-gamma}) for each ID--OOD pair. Performance is strongest at low-to-intermediate noise levels and often degrades at high noise, supporting the use of an ID-selected ensemble over a small early/mid-noise candidate set.}
    \label{fig:noise-sweep}
\end{figure}

Figure~\ref{fig:noise-sweep} provides a retrospective diagnostic of this choice. The
sweep varies the single noise level of the sensitivity channel, with the selected tap
and the reconstruction channel fixed. The strongest performance is typically achieved at
low-to-intermediate noise levels. For example, with CIFAR-10 as ID
(Figure~\ref{fig:noise-sweep-c10}), the best single
levels for CIFAR-100 and CelebA occur around $\tau=25$--$50$, while performance decreases
at larger noise levels. A similar pattern appears for CIFAR-100 as ID
(Figure~\ref{fig:noise-sweep-c100}), where early
levels such as $\tau=10$--$50$ are most informative for SVHN and remain competitive for
CIFAR-10. Very large noise levels often degrade performance because the noisy input
contains less sample-specific structure, consistent with the high-noise collapse of
Proposition~\ref{prop:high-noise-collapse} (Appendix~\ref{app:molrg-tap-highnoise}). This supports the use of a small ID-selected
ensemble over early and intermediate noise levels, rather than relying on a single
arbitrary timestep.

\subsection{ID-only evaluation of the fusion weight}
\label{app:id-gamma}

The fusion analysis of Appendix~\ref{app:proof-two-channel-fusion}
gives the convex weight that maximizes the standardized mean gap
for specified channel shifts and ID correlation.
We estimate the correlation directly from ID calibration data
and replace the unknown OOD shifts with shifts induced by the
geometric transformations $\mathcal G$ used for feature selection.
This gives a proxy-based weight without using OOD data.

Fix an ID denoiser and calibration set $\mathcal X_{\mathrm{cal}}$.
Let $\widetilde s_R$ and $\widetilde s_J$ denote the aggregated
reconstruction and sensitivity channels of
Algorithm~\ref{alg:detect}. 
The correlation estimate
\[
\hat\rho
=
\operatorname{Corr}_{\vx\in\mathcal X_{\mathrm{cal}}}
\bigl(\widetilde s_R(\vx),\widetilde s_J(\vx)\bigr)
\]
is computed directly on the ID calibration set.
Each transformation $g\in\mathcal G$ yields a surrogate shifted set
$\{g\vx:\vx\in\mathcal X_{\mathrm{cal}}\}$ for estimating the channel
shifts. For the sensitivity channel, we define
\[
\widehat\Delta_J
=
\frac{1}{|\mathcal G|}
\sum_{g\in\mathcal G}
\frac{
\operatorname{mean}_{g}(\widetilde s_J)
-
\operatorname{mean}_{\mathrm{cal}}(\widetilde s_J)
}{
\operatorname{std}_{\mathrm{cal}}(\widetilde s_J)
},
\]
where $\operatorname{mean}_g$ is taken over the shifted set of $g$,
and $\widehat\Delta_R$ is defined analogously.
We obtain $\hat\gamma$ by substituting
$(\hat\rho,\widehat\Delta_R,\widehat\Delta_J)$ for
$(\rho,\Delta_R,\Delta_J)$ in the optimal convex weight of
Remark~\ref{rem:convex-weight}. We take its interior solution only when $\hat d^\star\ge1$: by Remark~\ref{rem:dstar-bound}, below this level no weight separates the proxy means by one ID standard deviation, so we use the endpoint rule of Remark~\ref{rem:convex-weight}.

\begin{table}[t]
\caption{ID-only estimates used to determine the fusion weight
$\hat\gamma$ for each deployed denoiser.}
\label{tab:id-gamma}
\centering
\small
\begin{tabular}{lcccc}
\toprule
ID & $\hat\rho$ & $\widehat\Delta_J$ & $\widehat\Delta_R$
&  $\hat\gamma$ \\
\midrule
CIFAR-10 & $-0.096$ & $2.629$ & $0.439$  & $0.206$ \\
CIFAR-100 & $-0.367$ & $1.265$ & $0.154$  & $0.318$ \\
SVHN & $-0.347$ & $1.138$ & $0.674$  & $0.438$ \\
CelebA & $-0.191$ & $1.966$ & $2.053$  & $0.507$ \\
\bottomrule
\end{tabular}
\end{table}

\section{Additional Experiments} \label{app:additional-experiments}
This appendix extends the main evaluation. We first report
additional benchmarks: near-OOD pairs (\ref{app:near-ood}),
high-resolution FFHQ (\ref{app:highres}), ImageNet-64
(\ref{app:imagenet64}), and further OOD datasets
(\ref{app:additional-ood}). We then present an OOD-assisted variant of
feature selection, used only as a diagnostic (\ref{app:oodval}),
and a small transfer study on reinforcement-learning trajectories
(\ref{app:rl-planning}). The remaining subsections ablate the
sensitivity-estimation hyperparameters
(\ref{app:estimator-ablation}), the decoder tap
(\ref{app:layer-ablation}), and the fusion weight and
noise-ensemble size (\ref{app:ablation-alpha-kens}), and report
variability across random seeds (\ref{app:variance}).
\subsection{Near-OOD performance}
\label{app:near-ood}

We also evaluate near-OOD detection, where the OOD distribution is visually or semantically close to the ID data. Table~\ref{tab:nearood} reports results for CIFAR-10 and CIFAR-100 as ID datasets, with CIFAR-100/CIFAR-10 and TinyImageNet as near-OOD datasets. This setting is challenging because OOD samples often share low-level statistics and object-like structure with the ID distribution.

\method\ performs well in this difficult regime, achieving the best average AUROC of $0.835$, ahead of EigenScore's $0.794$. On the CIFAR-10/CIFAR-100 cross-dataset pairs, where most other diffusion baselines are close to chance, \method\ reaches $0.876$ for CIFAR-10 vs.\ CIFAR-100, nearly matching EigenScore's $0.880$, and obtains the best result for CIFAR-100 vs.\ CIFAR-10, improving from $0.642$ to $0.657$. On the TinyImageNet pairs it is the second-best detector ($0.952$ and $0.855$), behind DiffPath, whose trajectory-based statistic is particularly effective for this specific shift. Overall, these results show that \method\ is robust on near-OOD benchmarks, especially on fine-grained CIFAR-10/CIFAR-100 shifts, while also highlighting that different diffusion-based signals remain complementary across near-OOD types.

\begin{table}[t]
\caption{
Near-OOD detection results measured by AUROC. We evaluate CIFAR-10 and CIFAR-100 as ID datasets against semantically close OOD distributions, including CIFAR-100/CIFAR-10 and TinyImageNet. All baseline results use their default configuration. Best result per column is shown in \textbf{bold}, and second-best is \underline{underlined}.
}
\label{tab:nearood}
\centering
\footnotesize
\begin{tabular}{l cc cc c}
\toprule
InD & \multicolumn{2}{c}{CIFAR-10 vs.} & \multicolumn{2}{c}{CIFAR-100 vs.} & Avg \\
\cmidrule(lr){2-3}\cmidrule(lr){4-5}
OOD & C100 & TinyImageNet & C10 & TinyImageNet & \\
\midrule
DDPM-OOD & 0.618 & 0.570 & 0.462 & 0.457 & 0.527 \\
LMD & 0.601 & 0.592 & 0.569 & 0.558 & 0.580 \\
GEPC & 0.582 & 0.634 & 0.427 & 0.629 & 0.568 \\
DiffPath & 0.554 & \textbf{0.993} & 0.475 & \textbf{0.995} & 0.754 \\
EigenScore & \textbf{0.880} & 0.899 & \underline{0.642} & 0.755 & \underline{0.794} \\
\midrule
\textbf{\method\ (ours)} & \underline{0.876} & \underline{0.952} & \textbf{0.657} & \underline{0.855} & \textbf{0.835} \\
\bottomrule
\end{tabular}

\end{table}

\subsection{High-resolution OOD detection}
\label{app:highres}

Table~\ref{tab:ffhq-highres} evaluates the high-resolution setting with FFHQ as the ID distribution and AFHQ and Microscopy as OOD datasets. \method\ achieves near-perfect detection on both OOD datasets. The gain is especially pronounced on AFHQ, where the strongest baseline, GEPC, reaches $0.818$. On Microscopy, where DiffPath is already strong, \method\ still performs near-perfectly. The GEPC row uses GEPC's method and shipped hyperparameters verbatim with the same frozen FFHQ denoiser and image sets as \method. Overall, these results show that the proposed reconstruction--sensitivity fusion transfers effectively beyond low-resolution benchmarks and remains reliable in a high-resolution face-model setting. 

The FFHQ-256 denoiser is the guided-diffusion model described in Appendix~\ref{app:implementation}, conditioned on the raw integer timestep. We keep $K_{\mathrm{spec}}=3$, $\delta=0.1$, and $N_\epsilon=1$, but use $M=8$ probe directions instead of $32$, because central-difference probing at $256{\times}256$ is memory-bound. The candidate pool is $\mathcal{P}=\{25,50,150,200,250\}$, and the reconstruction channel uses all of its levels with one noise draw each. Stage~A of Algorithm~\ref{alg:selection} selects the tap \texttt{ob.5} ($S_{\mathrm{tap}}=0.97$, Figure~\ref{fig:tap-selection-ffhq}), and Stage~B keeps the three most stable levels, $\{150,200,250\}$. The fusion weight follows the ID-only rule of Appendix~\ref{app:id-gamma}, with the same geometric proxy shifts.

\begin{table}[t]
\caption{High-resolution OOD detection (AUROC): FFHQ-$256$ vs.\ AFHQ and Microscopy
(CHAMMI). Best per column
\textbf{bold}, second-best \underline{underlined}.}
\label{tab:ffhq-highres}
\centering
\small
\begin{tabular}{l ccc}
\toprule
Method & \multicolumn{3}{c}{FFHQ vs.} \\
\cmidrule(lr){2-4}
 & AFHQ & Microscopy & Avg \\
\midrule
LMD & 0.485 & 0.552 & 0.519 \\
GEPC & \underline{0.818} & 0.834 & \underline{0.826} \\
DiffPath & 0.593 & \underline{0.998} & 0.796 \\
EigenScore & 0.532 & 0.899 & 0.716 \\
\midrule
\textbf{\method\ (ours)} & \textbf{0.999} & \textbf{0.999} & \textbf{0.999} \\
\bottomrule
\end{tabular}
\end{table}

\subsection{ImageNet-64 results}
\label{app:imagenet64}

We additionally evaluate OOD detection with ImageNet-64 as the ID distribution. Table~\ref{tab:imagenet64} reports results against SVHN and Textures. This setting is challenging because ImageNet contains diverse object categories and visual statistics, making OOD detection less separable than in narrower ID domains.

On ImageNet-64 \method\ achieves the best average AUROC in this setting, improving over the strongest baseline average from $0.732$ for EigenScore to $0.882$. On SVHN, \method\ reaches $0.952$, ahead of LMD ($0.901$), while on Textures it reaches $0.813$, second to DiffPathV2 ($0.860$).

The ImageNet-64 denoiser is the improved-diffusion model described in Appendix~\ref{app:implementation}, conditioned on the rescaled timestep. We keep $K_{\mathrm{spec}}=3$, $\delta=0.1$, and $N_\epsilon=1$, use $M=8$ probe directions as for FFHQ-256, and use the candidate pool $\mathcal{P}=\{25,50,150,200,250\}$, whose levels the reconstruction channel also uses with one noise draw each. Stage~A selects \texttt{ob.12}, and Stage~B keeps the three most stable levels, $\{150,200,250\}$. The fusion weight follows the ID-only rule of Appendix~\ref{app:id-gamma}, with the same geometric proxy shifts.

\begin{table}[H]
\caption{
OOD detection results on ImageNet-64 measured by AUROC. ImageNet-64 is used as the ID distribution, and SVHN and Textures are used as OOD datasets. Best result per column is shown in \textbf{bold}, and second-best is \underline{underlined}.
}
\label{tab:imagenet64}
\centering
\footnotesize
\begin{tabular}{l ccc}
\toprule
Method & \multicolumn{3}{c}{ImageNet vs.} \\
\cmidrule(lr){2-4}
 & SVHN & Texture & Avg \\
\midrule
DDPM-OOD & 0.208 & 0.570 & 0.389 \\
LMD & \underline{0.901} & 0.486 & 0.693 \\
DiffPath & 0.753 & 0.699 & 0.726 \\
DiffPathV2 & 0.547 & \textbf{0.860} & 0.703 \\
EigenScore & 0.755 & 0.709 & \underline{0.732} \\
\midrule
\textbf{\method\ (ours)} & \textbf{0.952} & \underline{0.813} & \textbf{0.882} \\
\bottomrule
\end{tabular}
\end{table}

\subsection{Additional OOD datasets}
\label{app:additional-ood}

We further evaluate \method\ on additional OOD datasets commonly used in image OOD detection: LSUN, iSUN, and Textures. Table~\ref{tab:additional_ood} reports results for CIFAR-10 and CIFAR-100 as ID datasets. For this part, we use the same protocol as in the main paper.

\method\ achieves the best average AUROC across these additional OOD datasets, improving over the strongest baseline, EigenScore, from $0.848$ to $0.893$ (DiffPath reaches $0.758$ and GEPC $0.677$). The improvement is especially clear on Textures, where \method\ reaches $0.947$ for CIFAR-10 ID (best baseline $0.831$, GEPC) and $0.832$ for CIFAR-100 ID (best baseline $0.761$, DiffPath). On scene-style OOD datasets such as LSUN and iSUN, EigenScore remains highly competitive, especially for CIFAR-100 ID; nevertheless, the fused reconstruction--sensitivity score gives the best overall performance across the six additional ID--OOD pairs.

\begin{table}[H]
\caption{
Additional OOD detection results measured by AUROC. We evaluate CIFAR-10 and CIFAR-100 as ID datasets against LSUN, iSUN, and Textures as OOD datasets. Best result per column is shown in \textbf{bold}, and second-best is \underline{underlined}.
}
\label{tab:additional_ood}
\centering
\footnotesize
  \begin{tabular}{l ccc ccc c}
  \toprule
  Method & \multicolumn{3}{c}{CIFAR-10 vs.} & \multicolumn{3}{c}{CIFAR-100 vs.} & Avg \\
  \cmidrule(lr){2-4}\cmidrule(lr){5-7}
   & LSUN & iSUN & Textures & LSUN & iSUN & Textures & \\
  \midrule
  GEPC & 0.672 & 0.694 & \underline{0.831} & 0.574 & 0.568 & 0.721 & 0.677 \\
  DiffPath & 0.618 & 0.911 & 0.780 & 0.609 & \underline{0.869} & \underline{0.761} & 0.758 \\
  EigenScore & \textbf{0.977} & \textbf{0.974} & 0.756 & \textbf{0.877} & \textbf{0.879} & 0.627 & \underline{0.848} \\
  \midrule
  \textbf{\method\ (ours)} & \underline{0.952} & \underline{0.955} & \textbf{0.947} & \underline{0.830} & 0.844 & \textbf{0.832} & \textbf{0.893} \\
  \bottomrule
  \end{tabular}
\end{table}

\subsection{OOD-assisted feature selection (diagnostic)}
\label{app:oodval}

\method-OODVal is an OOD-assisted diagnostic variant of \method\ that uses a disjoint validation split from a single auxiliary OOD dataset (CelebA for the CIFAR-10, CIFAR-100, and SVHN denoisers; CIFAR-10 for the CelebA denoiser) only for feature tap and sensitivity-noise selection; all other hyperparameters remain fixed, and the auxiliary split is never used for scoring or calibration. Because it consumes OOD examples, this variant is excluded from the fully unsupervised comparison of Table~\ref{tab:main-results} and is reported only as a diagnostic of the selection stage.

Table~\ref{tab:oodval} compares the deployed, fully unsupervised \method\ with \method-OODVal on the main benchmark. The variant reaches $0.920$ average AUROC versus $0.897$ for \method, indicating that feature and noise-level selection, rather than the readouts themselves, remain an important source of headroom for the fully unsupervised detector.

\begin{table}[H]
\caption{
Fully unsupervised \method\ versus the OOD-assisted diagnostic variant \method-OODVal, measured by AUROC on the main benchmark. The auxiliary OOD validation set is used only for feature tap and sensitivity-noise selection.
}
\label{tab:oodval}
\centering
\footnotesize
\resizebox{\textwidth}{!}{
\begin{tabular}{l ccc ccc ccc ccc c}
\toprule
InD & \multicolumn{3}{c}{CelebA vs.} & \multicolumn{3}{c}{CIFAR-10 vs.} & \multicolumn{3}{c}{CIFAR-100 vs.} & \multicolumn{3}{c}{SVHN vs.} & Avg \\
\cmidrule(lr){2-4}\cmidrule(lr){5-7}\cmidrule(lr){8-10}\cmidrule(lr){11-13}
OOD & CIFAR-10 & CIFAR-100 & SVHN & CIFAR-100 & SVHN & CelebA & CIFAR-10 & CelebA & SVHN & CIFAR-10 & CIFAR-100 & CelebA & \\
\midrule
\method\ & 1.000 & 1.000 & 1.000 & 0.876 & 0.986 & 0.872 & 0.657 & 0.544 & 0.839 & 0.995 & 0.994 & 0.996 & 0.897 \\
\method-OODVal & 1.000 & 1.000 & 1.000 & 0.898 & 0.982 & 0.941 & 0.645 & 0.769 & 0.817 & 0.996 & 0.995 & 1.000 & 0.920 \\
\bottomrule
\end{tabular}

}
\end{table}

\subsection{RL planning study}
\label{app:rl-planning}

As a small out-of-domain study, not intended as a full RL OOD benchmark, we
test whether the reconstruction--sensitivity score extends from images to
trajectories. We use offline RL trajectories from D4RL~\citep{fu2020d4rl} and
the released Diffuser planning models of \citet{janner2022diffuser} as frozen
denoisers. For each environment, the Diffuser model trained on one data quality
defines the ID, and windows from the other qualities are OOD; this
behavior-policy shift setting is also used as an RL case study by
\citet{barkley2025scoped}, with a diffusion model trained for the task. The
score is computed per trajectory window from the same two channels: the
reconstruction residual against the clean window and the sensitivity spectrum
of an intermediate temporal U-Net tap. The tap and noise levels are selected by
Algorithm~\ref{alg:selection}, with the geometric image transforms replaced by
trajectory proxy shifts of the ID calibration windows: Gaussian perturbations of
the state or action channels, a fixed random permutation of the timesteps, and
time reversal.

Table~\ref{tab:rl-planning} shows the resulting AUROC matrix. Large
behavior-policy shifts are detected reliably: random trajectories reach AUROC
between $0.997$ and $1.000$ for every ID model, and expert trajectories are also
detected well. The main failure mode is near-quality detection: medium
trajectories are ranked near chance as OOD against medium-replay ID data, which
they resemble as a sub-population. Overall, the same frozen-model construction,
ID-only selection, and one-sided fusion transfer to temporal diffusion models,
with only the proxy shifts adapted to trajectories.

\begin{table}[H]
\caption{
Trajectory-level OOD detection on D4RL using frozen Diffuser planning backbones.
Rows indicate the ID environment and data quality used for Diffuser;
columns indicate the OOD trajectory quality. We report one-sided AUROC per trajectory
window using the fused reconstruction--sensitivity score. ``--'' denotes the ID quality
itself; the last column averages the shown OOD qualities.
}
\label{tab:rl-planning}
\centering
\footnotesize
\begin{tabular}{llccccc}
\toprule
Environment & ID quality & Random & Medium & Med.-Replay & Expert & Avg. \\
\midrule
\multirow{2}{*}{HalfCheetah}
& Medium        & 0.999 & --    & 0.869 & 1.000 & 0.956 \\
& Medium-Replay & 0.997 & 0.513 & --    & 0.999 & 0.837 \\
\midrule
\multirow{2}{*}{Hopper}
& Medium        & 1.000 & --    & 0.991 & 1.000 & 0.997 \\
& Medium-Replay & 1.000 & 0.579 & --    & 0.998 & 0.859 \\
\midrule
\multirow{2}{*}{Walker2d}
& Medium        & 1.000 & --    & 0.940 & 0.993 & 0.978 \\
& Medium-Replay & 1.000 & 0.516 & --    & 0.882 & 0.799 \\
\bottomrule
\end{tabular}
\end{table}

\subsection{Ablation study on sensitivity-estimation hyperparameters}
\label{app:estimator-ablation}

Table~\ref{tab:ablation-sensitivity} studies the effect of the main sensitivity-estimation hyperparameters: the number of finite-difference probes $M$, the number of leading eigenvalues $K_{\mathrm{spec}}$, the number of noise repetitions $N_{\epsilon}$, and the finite-difference probe scale $\delta$. Overall, \method\ is stable across these choices. Increasing the number of probes from $24$ to $48$ changes the average AUROC only from $0.892$ to $0.901$, while the default $M=32$ already achieves $0.897$. Similarly, varying the spectral rank from $K_{\mathrm{spec}}=1$ to $3$ gives nearly identical average performance, between $0.892$ and $0.897$, indicating that the sensitivity signal is not driven by a fragile choice of eigenspectrum dimension. Increasing the number of repetitions from $N_{\epsilon}=1$ to $N_{\epsilon}=3$ yields only a marginal gain, from $0.897$ to $0.898$. Finally, varying the probe scale over $\delta\in\{0.05, 0.1, 0.15\}$ around the default $\delta=0.1$ leaves performance essentially unchanged (average AUROC $0.897$ at every scale, with no cell moving by more than $0.001$), indicating that the finite-difference responses are measured well inside the local-linear regime of the feature map. We also evaluated replacing the top-$K_{\mathrm{spec}}$ readout with the full Gram trace (the sum of all $M$ eigenvalues) at the deployed configuration: the 12-pair average AUROC drops from $0.897$ to $0.862$.

\begin{table}[t]
\caption{Ablation of sensitivity-estimation hyperparameters in AUROC. We vary the number of finite-difference probes $M$, leading eigenvalues $K_{\mathrm{spec}}$, noise repetitions $N_{\epsilon}$, and the probe scale $\delta$ while keeping all other settings fixed. Results show that \method\ is stable across estimator choices. All entries use the fused score at the deployed per-denoiser weights $\hat\gamma$ (Appendix~\ref{app:id-gamma}). Best result within each ablation block is shown in \textbf{bold}.}
\label{tab:ablation-sensitivity}
\centering
\footnotesize
\resizebox{\textwidth}{!}{
\begin{tabular}{l ccc ccc ccc ccc c}
\toprule
Probes $M$ & \multicolumn{3}{c}{CelebA vs.} & \multicolumn{3}{c}{CIFAR-10 vs.} & \multicolumn{3}{c}{CIFAR-100 vs.} & \multicolumn{3}{c}{SVHN vs.} & Avg \\
\cmidrule(lr){2-4}\cmidrule(lr){5-7}\cmidrule(lr){8-10}\cmidrule(lr){11-13}
 & CIFAR-10 & CIFAR-100 & SVHN & CIFAR-100 & SVHN & CelebA & CIFAR-10 & CelebA & SVHN & CIFAR-10 & CIFAR-100 & CelebA & \\
\midrule
24 & \textbf{1.000} & \textbf{0.999} & \textbf{1.000} & 0.871 & 0.985 & 0.859 & 0.650 & 0.516 & 0.838 & \textbf{0.995} & \textbf{0.994} & \textbf{0.996} & 0.892 \\
32 & \textbf{1.000} & \textbf{1.000} & \textbf{1.000} & 0.876 & \textbf{0.986} & 0.872 & 0.657 & 0.544 & \textbf{0.839} & \textbf{0.995} & \textbf{0.994} & \textbf{0.996} & 0.897 \\
48 & \textbf{1.000} & \textbf{1.000} & \textbf{1.000} & \textbf{0.881} & \textbf{0.986} & \textbf{0.883} & \textbf{0.662} & \textbf{0.573} & 0.838 & \textbf{0.995} & \textbf{0.994} & \textbf{0.996} & \textbf{0.901} \\
\midrule
Eigenvalues $K_{\mathrm{spec}}$ & \multicolumn{3}{c}{CelebA vs.} & \multicolumn{3}{c}{CIFAR-10 vs.} & \multicolumn{3}{c}{CIFAR-100 vs.} & \multicolumn{3}{c}{SVHN vs.} & Avg \\
\cmidrule(lr){2-4}\cmidrule(lr){5-7}\cmidrule(lr){8-10}\cmidrule(lr){11-13}
 & CIFAR-10 & CIFAR-100 & SVHN & CIFAR-100 & SVHN & CelebA & CIFAR-10 & CelebA & SVHN & CIFAR-10 & CIFAR-100 & CelebA & \\
\midrule
1 & \textbf{1.000} & \textbf{1.000} & \textbf{1.000} & 0.869 & 0.984 & \textbf{0.884} & 0.656 & 0.536 & 0.795 & 0.994 & 0.993 & \textbf{0.996} & 0.892 \\
2 & \textbf{1.000} & \textbf{1.000} & \textbf{1.000} & 0.873 & \textbf{0.986} & 0.873 & \textbf{0.659} & \textbf{0.549} & 0.827 & 0.994 & \textbf{0.994} & \textbf{0.996} & 0.896 \\
3 & \textbf{1.000} & \textbf{1.000} & \textbf{1.000} & \textbf{0.876} & \textbf{0.986} & 0.872 & 0.657 & 0.544 & \textbf{0.839} & \textbf{0.995} & \textbf{0.994} & \textbf{0.996} & \textbf{0.897} \\
\midrule
Repetitions $N_{\epsilon}$ & \multicolumn{3}{c}{CelebA vs.} & \multicolumn{3}{c}{CIFAR-10 vs.} & \multicolumn{3}{c}{CIFAR-100 vs.} & \multicolumn{3}{c}{SVHN vs.} & Avg \\
\cmidrule(lr){2-4}\cmidrule(lr){5-7}\cmidrule(lr){8-10}\cmidrule(lr){11-13}
 & CIFAR-10 & CIFAR-100 & SVHN & CIFAR-100 & SVHN & CelebA & CIFAR-10 & CelebA & SVHN & CIFAR-10 & CIFAR-100 & CelebA & \\
\midrule
1 & \textbf{1.000} & \textbf{1.000} & \textbf{1.000} & 0.876 & 0.986 & 0.872 & 0.657 & \textbf{0.544} & 0.839 & 0.995 & 0.994 & 0.996 & 0.897 \\
2 & \textbf{1.000} & \textbf{1.000} & \textbf{1.000} & \textbf{0.878} & \textbf{0.988} & \textbf{0.879} & 0.666 & 0.528 & 0.855 & 0.995 & 0.994 & 0.996 & \textbf{0.898} \\
3 & \textbf{1.000} & \textbf{1.000} & \textbf{1.000} & 0.877 & \textbf{0.988} & 0.877 & \textbf{0.667} & 0.528 & \textbf{0.856} & \textbf{0.996} & \textbf{0.995} & \textbf{0.998} & \textbf{0.898} \\
\midrule
Probe scale $\delta$ & \multicolumn{3}{c}{CelebA vs.} & \multicolumn{3}{c}{CIFAR-10 vs.} & \multicolumn{3}{c}{CIFAR-100 vs.} & \multicolumn{3}{c}{SVHN vs.} & Avg \\
\cmidrule(lr){2-4}\cmidrule(lr){5-7}\cmidrule(lr){8-10}\cmidrule(lr){11-13}
 & CIFAR-10 & CIFAR-100 & SVHN & CIFAR-100 & SVHN & CelebA & CIFAR-10 & CelebA & SVHN & CIFAR-10 & CIFAR-100 & CelebA & \\
\midrule
0.05 & \textbf{1.000} & \textbf{1.000} & \textbf{1.000} & \textbf{0.876} & \textbf{0.986} & \textbf{0.872} & \textbf{0.658} & \textbf{0.544} & 0.839 & \textbf{0.995} & \textbf{0.994} & \textbf{0.996} & \textbf{0.897} \\
0.1 & \textbf{1.000} & \textbf{1.000} & \textbf{1.000} & \textbf{0.876} & \textbf{0.986} & \textbf{0.872} & 0.657 & \textbf{0.544} & 0.839 & \textbf{0.995} & \textbf{0.994} & \textbf{0.996} & \textbf{0.897} \\
0.15 & \textbf{1.000} & 0.999 & \textbf{1.000} & \textbf{0.876} & \textbf{0.986} & \textbf{0.872} & 0.657 & \textbf{0.544} & \textbf{0.840} & \textbf{0.995} & \textbf{0.994} & \textbf{0.996} & \textbf{0.897} \\
\bottomrule
\end{tabular}

}
\end{table}

\subsection{Ablation study on layers}
\label{app:layer-ablation}

We further study the effect of the decoder tap used for the sensitivity channel. In this ablation, the sensitivity noise levels are set according to Algorithm~\ref{alg:selection}, and we sweep only over the feature tap $\ell$. 

\begin{figure}[t]
    \centering
    \includegraphics[width=.8\linewidth]{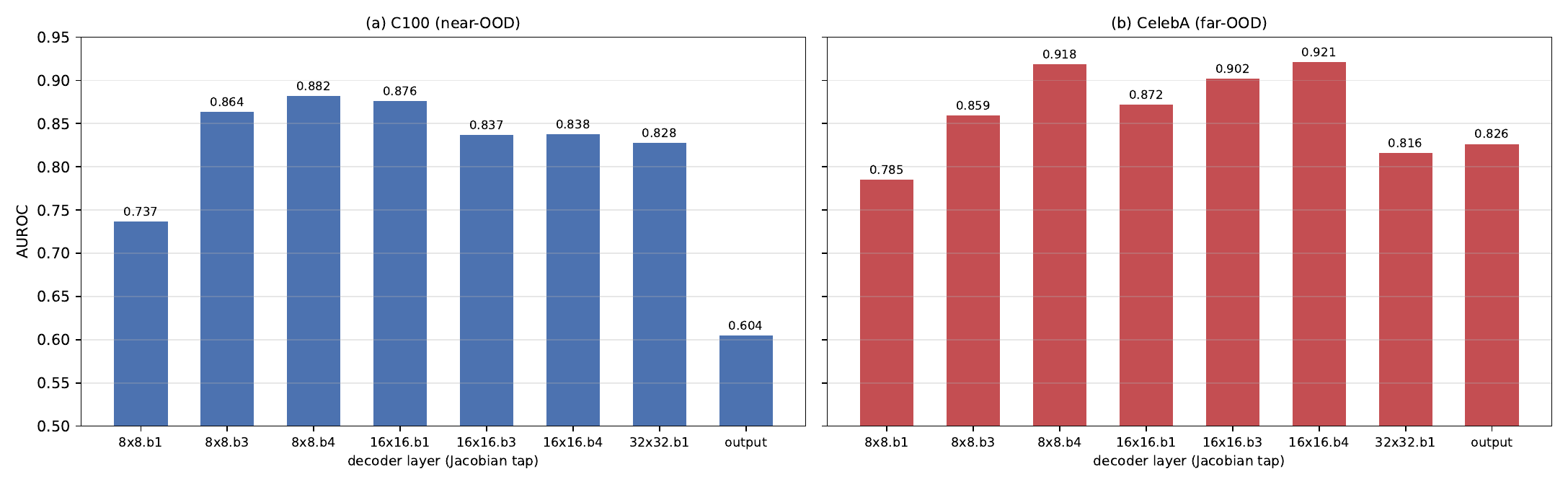}
    \caption{
    \textbf{Layer ablation with CIFAR-10 as ID.}
    We fix the sensitivity noise levels using the ID-only noise-selection procedure and sweep only the decoder tap used for sensitivity estimation. Intermediate decoder taps generally outperform both early layers and the final output layer. The bars show the fused score at the deployed weight $\hat\gamma$ (Appendix~\ref{app:id-gamma}) for each candidate tap.
    }
    \label{fig:layer-ablation-cifar10}
\end{figure}

\begin{figure}[t]
    \centering
    \includegraphics[width=.8\linewidth]{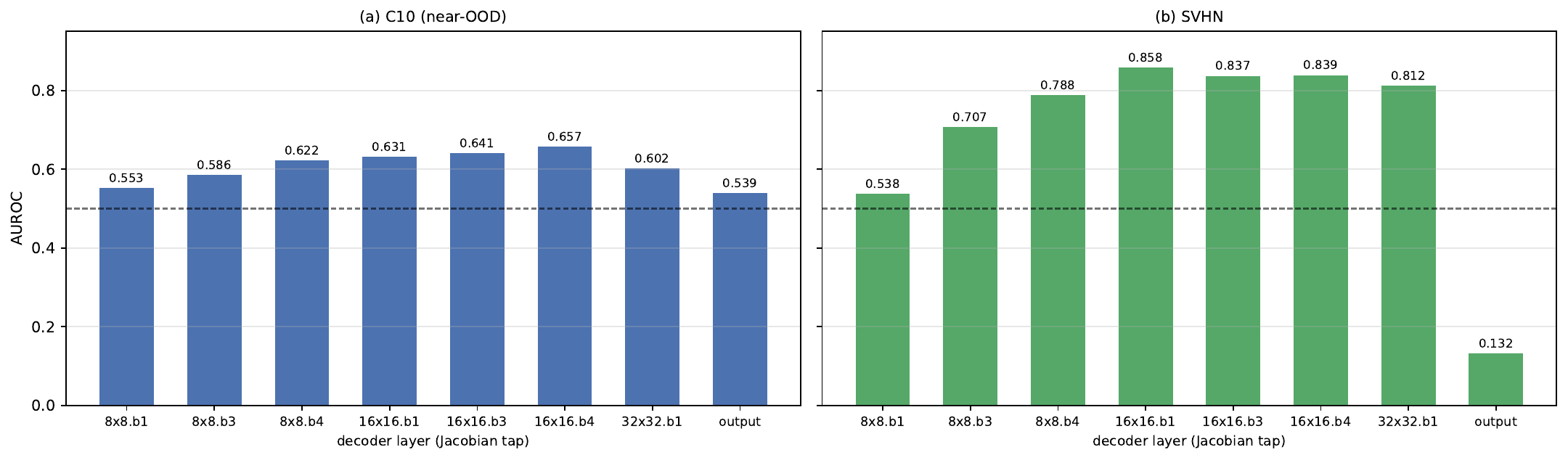}
    \caption{
    \textbf{Layer ablation with CIFAR-100 as ID.}
    With fixed selected noise levels, we vary the decoder tap used by the sensitivity channel. Mid-level decoder taps provide the strongest performance for both near-OOD and far-OOD settings, while the output layer is substantially weaker. The bars show the fused score at the deployed weight $\hat\gamma$ (Appendix~\ref{app:id-gamma}) for each candidate tap.
    }
    \label{fig:layer-ablation-cifar100}
\end{figure}

\begin{figure}[t]
    \centering
    \includegraphics[width=.7\linewidth]{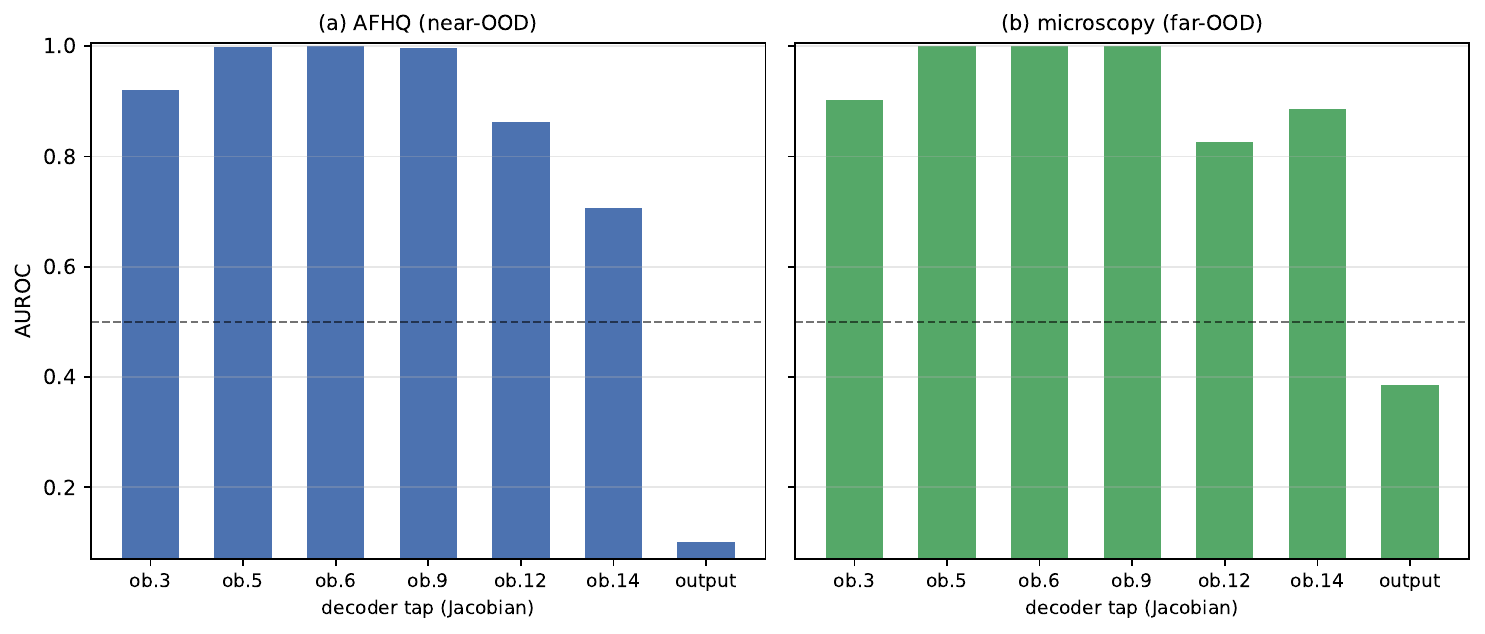}
    \caption{
    \textbf{Layer ablation with FFHQ-256 as ID.}
    We sweep output-block taps while keeping the sensitivity noise levels fixed. Early-to-middle output blocks give the strongest sensitivity signal for both AFHQ and Microscopy OOD datasets, whereas later blocks and the final output are less reliable. The bars show the fused score at the deployed weight $\hat\gamma$ (Appendix~\ref{app:id-gamma}) for each candidate tap.
    }
    \label{fig:layer-ablation-ffhq}
\end{figure}

Figures~\ref{fig:layer-ablation-cifar10}--\ref{fig:layer-ablation-ffhq} show that the sensitivity signal depends on the chosen feature tap. Across CIFAR-10, CIFAR-100, and FFHQ-256, intermediate or mid-to-late decoder layers generally perform better than very early layers or the final output layer. This supports using intermediate feature sensitivity rather than relying only on the output denoiser Jacobian. The ID-only tap-selection procedure follows the same pattern. Although it never sees target OOD data, it consistently selects taps in the informative intermediate region identified by the sweeps. Thus, the selection rule acts as a robust, label-free proxy for choosing useful sensitivity features. The gap to the OOD-assisted \method-OODVal variant (Appendix~\ref{app:oodval}) reflects the expected difference between a single ID-selected tap and an oracle choice tuned to a particular OOD distribution.

\subsection{Ablation study on $\gamma$ and $K_{\mathrm{ens}}$}
\label{app:ablation-alpha-kens}

We ablate the fusion weight $\gamma$ and the number of selected
sensitivity noise levels $K_{\mathrm{ens}}$ in
Table~\ref{tab:ablation-sensitivity_alpha_k}. The weight $\gamma$
controls the reconstruction contribution, with $\gamma=0$ corresponding
to sensitivity only and $\gamma=1$ to reconstruction only. On average,
neither channel alone performs best. Sensitivity alone reaches an AUROC
of $0.856$, while reconstruction alone drops to $0.704$, consistent with
the reconstruction-inversion failure discussed in
Section~\ref{subsec:mse-insufficient}. Intermediate weights perform
substantially better. Larger reconstruction weights help in a
few cases but degrade performance where reconstruction is unreliable.

We next vary $K_{\mathrm{ens}}$ while keeping the deployed
ID-evaluated weights $\hat\gamma$. Performance is stable for
$K_{\mathrm{ens}}\in\{1,2,3\}$, with average AUROC between $0.887$ and
$0.897$. Using three selected noise levels gives the best average result
and improves several difficult CIFAR-100 cases. Overall, the ablation
supports combining reconstruction with sensitivity and aggregating a
small set of ID-selected sensitivity levels.

\begin{table}[H]
\caption{
Ablation of the fusion weight $\gamma$ and sensitivity ensemble size
$K_{\mathrm{ens}}$, measured by AUROC. In the $\gamma$ block,
$K_{\mathrm{ens}}=3$, with $\gamma=0$ denoting sensitivity only and
$\gamma=1$ reconstruction only. The final row reports the deployed
per-denoiser ID-evaluated weights $\hat\gamma$
(Appendix~\ref{app:id-gamma}).
In the $K_{\mathrm{ens}}$ block, $\gamma$ is fixed to the deployed
$\hat\gamma$. Best results within each ablation block are shown in
\textbf{bold}.
}
\label{tab:ablation-sensitivity_alpha_k}
\centering
\footnotesize
\resizebox{\textwidth}{!}{
\begin{tabular}{l ccc ccc ccc ccc c}
\toprule
Weight $\gamma$ & \multicolumn{3}{c}{CelebA vs.} & \multicolumn{3}{c}{CIFAR-10 vs.} & \multicolumn{3}{c}{CIFAR-100 vs.} & \multicolumn{3}{c}{SVHN vs.} & Avg \\
\cmidrule(lr){2-4}\cmidrule(lr){5-7}\cmidrule(lr){8-10}\cmidrule(lr){11-13}
 & CIFAR-10 & CIFAR-100 & SVHN & CIFAR-100 & SVHN & CelebA & CIFAR-10 & CelebA & SVHN & CIFAR-10 & CIFAR-100 & CelebA & \\
\midrule
$0$ & 0.991 & 0.978 & 0.997 & 0.861 & \textbf{0.988} & 0.800 & 0.633 & 0.501 & \textbf{0.899} & 0.861 & 0.866 & 0.900 & 0.856 \\

$0.5$ & \textbf{1.000} & 0.999 & \textbf{1.000} & 0.854 & 0.956 & \textbf{0.918} & 0.630 & 0.614 & 0.679 & 0.997 & 0.996 & 0.998 & 0.887 \\

$1$ & 0.996 & 0.993 & 0.558 & 0.601 & 0.148 & 0.827 & 0.524 & \textbf{0.697} & 0.122 & 0.993 & 0.994 & \textbf{1.000} & 0.704 \\
\midrule
$\hat\gamma$ (deployed) & \textbf{1.000} & \textbf{1.000} & \textbf{1.000} & \textbf{0.876} & 0.986 & 0.872 & 0.657 & 0.544 & 0.839 & 0.995 & 0.994 & 0.996 & \textbf{0.897} \\
\midrule
Ensemble $K_{\mathrm{ens}}$ & \multicolumn{3}{c}{CelebA vs.} & \multicolumn{3}{c}{CIFAR-10 vs.} & \multicolumn{3}{c}{CIFAR-100 vs.} & \multicolumn{3}{c}{SVHN vs.} & Avg \\
\cmidrule(lr){2-4}\cmidrule(lr){5-7}\cmidrule(lr){8-10}\cmidrule(lr){11-13}
 & CIFAR-10 & CIFAR-100 & SVHN & CIFAR-100 & SVHN & CelebA & CIFAR-10 & CelebA & SVHN & CIFAR-10 & CIFAR-100 & CelebA & \\
\midrule
$1$ & \textbf{1.000} & \textbf{1.000} & \textbf{1.000} & \textbf{0.879} & 0.985 & 0.868 & 0.646 & 0.509 & 0.804 & \textbf{0.996} & \textbf{0.995} & \textbf{0.999} & 0.890 \\
$2$ & \textbf{1.000} & \textbf{1.000} & \textbf{1.000} & 0.861 & \textbf{0.987} & \textbf{0.879} & 0.650 & 0.457 & 0.817 & \textbf{0.996} & \textbf{0.995} & 0.998 & 0.887 \\
$3$ & \textbf{1.000} & \textbf{1.000} & \textbf{1.000} & 0.876 & 0.986 & 0.872 & \textbf{0.657} & \textbf{0.544} & \textbf{0.839} & 0.995 & 0.994 & 0.996 & \textbf{0.897} \\
\bottomrule
\end{tabular}

}
\end{table}

\subsection{Sensitivity to seed selection}
\label{app:variance}
To assess variability from randomness, we keep all method hyperparameters constant and change only the random seed.

\begin{table}[H]
\caption{Mean AUROC per ID dataset on the main benchmark (average over its
three OOD sets) with standard deviation over $5$ seeds.}
\label{tab:auroc-std}
\centering
\small
\begin{tabular}{lcccc|c}
\toprule
 & CelebA & CIFAR-10 & CIFAR-100 & SVHN & Avg \\
\midrule
AUROC & $0.9996 \pm 0.0004$ & $0.9113 \pm 0.0044$ & $0.6804 \pm 0.0172$ & $0.9946 \pm 0.0049$ & $0.8965 \pm 0.0051$ \\
\bottomrule
\end{tabular}
\end{table}
Table~\ref{tab:auroc-std} reports per-dataset means and standard deviations for \method\ on the main benchmark with the standard deviation computed over $5$ seeds.

\end{document}